\documentclass[10pt,twocolumn,letterpaper]{article}

\usepackage{wacv}
\usepackage{times}
\usepackage{epsfig}
\usepackage{graphicx}
\usepackage{amsmath}
\usepackage{amssymb}

\usepackage{xspace}
\usepackage[acronym]{glossaries}
\usepackage{float}
\usepackage{subfig}
\usepackage{mathtools}

\usepackage{xcolor}
\usepackage{booktabs}
\usepackage{colortbl}
\usepackage{threeparttable}
\usepackage{multirow}
\usepackage[ruled,vlined]{algorithm2e}
\usepackage{enumerate}

\newacronym{gan}{GAN}{Generative Adversarial Network}
\newacronym{asal}{ASAL}{Adversarial Sampling for Active Learning}
\newacronym{gaal}{GAAL}{Generative Adversarial Active Learning}
\newacronym{svm}{SVM}{Support Vector Machine}
\newacronym{pca}{PCA}{Principal Component Analysis}
\newacronym{al}{AL}{Active Learning}
\newacronym{adba}{ADBA}{Active Decision Boundary Annotation}
\DeclareMathOperator*{\argmin}{arg\,min}

\usepackage[pagebackref=true,breaklinks=true,letterpaper=true,colorlinks,bookmarks=false]{hyperref}

\wacvfinalcopy 

\def\wacvPaperID{500} 

\ifwacvfinal\fi
\begin{document}

\title{Adversarial Sampling for Active Learning}

\author{Christoph Mayer \hspace{1cm} Radu Timofte \\
{\tt\small \{chmayer,timofter\}@vision.ee.ethz.ch}\\
Computer Vision Lab, ETH Z{\"u}rich, Switzerland\\
}

\maketitle
\ifwacvfinal\thispagestyle{empty}\fi

\begin{abstract}
This paper proposes \acrshort{asal}, a new \acrshort{gan} based active learning method that generates high entropy samples. Instead of directly annotating the synthetic samples, \acrshort{asal} searches similar samples from the pool and includes them for training. Hence, the quality of new samples is high and annotations are reliable.
To the best of our knowledge, \acrshort{asal} is the first \acrshort{gan} based \acrshort{al} method applicable to multi-class problems that outperforms random sample selection. Another benefit of \acrshort{asal} is its small run-time complexity (sub-linear) compared to traditional uncertainty sampling (linear).
We present a comprehensive set of experiments on multiple traditional data sets and show that \acrshort{asal} outperforms similar methods and clearly exceeds the established baseline (random sampling). In the discussion section we analyze in which situations \acrshort{asal} performs best and why it is sometimes hard to outperform random sample selection.

\end{abstract}


\section{Introduction}
The goal of \gls{al} algorithms is to train a model most efficiently, i.e. achieving the best
performance with as few labelled samples as possible.
Typical \gls{al} algorithms operate in an iterative fashion, where in each \gls{al} cycle a query
strategy selects samples that the oracle should annotate. These samples are expected to improve
the model most effectively when added to the training set. This procedure continues until a predefined
stopping criteria is met.

In this paper we will mainly focus on pool based active learning, because a pool of unlabelled samples is
often available beforehand or can easily be built. 
Furthermore, annotating all pool samples serves as an ideal evaluation environment for active learning algorithms.
It enables to train a fully-supervised model that establishes a performance upper bound on this data set.
Similarly, randomly selecting instead of actively choosing samples establishes a lower bound. Then, the
goal of an active learning algorithm is to approximate the performance of the fully supervised model with as few
labelled samples as possible, while exceeding the performance of random sampling. 

\textit{Uncertainty sampling} is an effective query strategy that identifies samples that are more informative than random ones.
The heuristic is, that samples for which the model is most uncertain contain new information and improve the model.
Uncertainty sampling is the most commonly used \gls{al} strategy for \gls{gan} based \gls{al} methods~\cite{zhu2017,huijser2017}.
However, these related methods are designed for small and very simple datasets, cover only binary classification tasks and use \glspl{svm} for classification instead of CNNs. \gls{gaal}~\cite{zhu2017} even fails to outperform random sample selection.

Our contributions are as follows:

\begin{itemize}
    \item \gls{asal} is to the best of our knowledge the first \emph{pool} based \gls{al} method that uses \gls{gan} to tackle multi-class problems.
    \item \gls{asal} achieves sub-linear run-time complexity even though it searches the full pool in each \gls{al} cycle.
    \item We validate \gls{asal} on full image data sets (MNIST, CIFAR-10, CelebA, SVHN, LSUN) compared to related methods that use much simpler challenges such as: two class subsets of MNIST, CIFAR-10 or SVHN.
\end{itemize}
\section{Related Work}

We review related work on \gls{al} especially on pool based uncertainty sampling, \gls{gan} based \gls{al} strategies and methods attempting to improve the run-time complexity.

Pool-based active learning methods select new training samples from a predefined unlabelled data set~\cite{hospedales2013,nguyen2004,yang2015,sener2018,yoo2019learningloss}. A common query strategy to identify new samples is uncertainty sampling~\cite{joshi2009,yang2016}. 
A well known uncertainty sampling strategy that is used to train \glspl{svm}, is minimum distance sampling~\cite{tong2001,campbell2000}.
Minimum distance sampling requires a linear classifier in some feature space and assumes that the classifier is uncertain about samples in the vicinity of the separating hyper-plane. This strategy is mainly used for two class but can be extended to multi-class problems~\cite{jain2010}. Joshi~\etal~\cite{joshi2009} use information entropy to measure the uncertainty of the classifier for a particular sample. Computing uncertainty with information entropy is suitable for two or multiple classes.

Jain~\etal~\cite{jain2010} propose two hashing based method to accelerate minimum distance sampling by selecting new samples in sub-linear time.
These methods are designed to select the closest point (approximately) to a hyper-plane in a $k$-dimensional feature space, where the positions of the data points are fixed but the hyper-plane is allowed to move. Thus, these methods are limited to \glspl{svm} with fixed feature maps, because, if the feature map changes, the position of the samples become obsolete and need to be recomputed. Hence, the run time complexity is sub-linear for constant feature maps and linear otherwise.
Unfortunately, CNN based methods update their feature maps during training. Thus, their methods are as efficient as exhaustive uncertainty sampling if CNNs are involved.

Zhu and Bento~\cite{zhu2017} propose \gls{gaal}, that uses a \gls{gan} to generate uncertain synthetic samples in each \gls{al} cycle.
Generating instead of selecting uncertain samples leads to a constant run-time complexity
because producing a new sample is independent of the pool size but requires training a \gls{gan} beforehand.
Zhu and Bento~\cite{zhu2017} use the traditional minimal distance optimization problem but replace the variable $x$ (denoting a pool sample) with the trained generator. Then, they use gradient descent to minimize the objective. The latent variable minimizing the objective results in a synthetic image close to the separating hyper-plane. They annotate the synthetic sample and use it for training. Zhu and Bento~\cite{zhu2017} demonstrate \gls{gaal} on subsets of MNIST and CIFAR-10 (two classes) using linear \glspl{svm} and DCGANs~\cite{radford2015,goodfellow2014}. However, \gls{gaal} performs worse than random sampling on both data sets, because it suffers from sampling bias and annotating is arbitrarily hard caused by sometimes poor quality of the synthetic uncertain samples. Note, that \gls{gaal} requires visually distinct classes (\textit{horse} \& \textit{automobile}) to allow reliable annotations by humans.

\gls{adba}~\cite{huijser2017} is another \gls{gan} based \gls{al} strategy. The main contributions of \gls{adba} are, training the classifier in the latent space and a new annotation scheme. Hence, it requires computing the latent state representation of each data sample using the pretrained \gls{gan}. Then, \gls{adba} searches the most uncertain sample in the latent space and generates a
line that is perpendicular to the current decision boundary and crosses the most uncertain sample. Then, the \gls{gan} generates images along this line such that the annotator can specify for which image in the sequence along this line the class label changes.
\gls{adba} shows that it outperforms uncertainty sampling in the latent space but misses to compare to uncertainty sampling in image space. Computing the latent space representation for each sample using a \gls{gan} is very costly and requires high quality \glspl{gan}. Sampling lines in the latent space of multi-class problems might lead to many crossings. Such lines might be arbitrarily hard to annotate especially if many crossings are close and annotating a line is more costly than annotating a single image.

Thus, we propose \gls{asal} that reuses the sample generation idea of Zhu and Bento~\cite{zhu2017} but we use information entropy as uncertainty score and directly extend it to multiple classes.
Our main contribution is avoiding to annotate synthetic images by selecting the most similar samples from the pool with a newly developed sample matching method. We propose three different feature maps that we compute for each pool sample to fit a fast nearest neighbour model beforehand. During active learning, we compute the feature map of the synthetic sample and retrieve the most similar one from the pool in sub-linear time. Additionally, \gls{asal} uses CNN based classifiers instead of linear \glspl{svm}. For the generator we train Wasserstein \glspl{gan} beforehand~\cite{arjovsky2017}.

\begin{figure*}[t]
\centering
\includegraphics[width=0.8\textwidth, keepaspectratio]{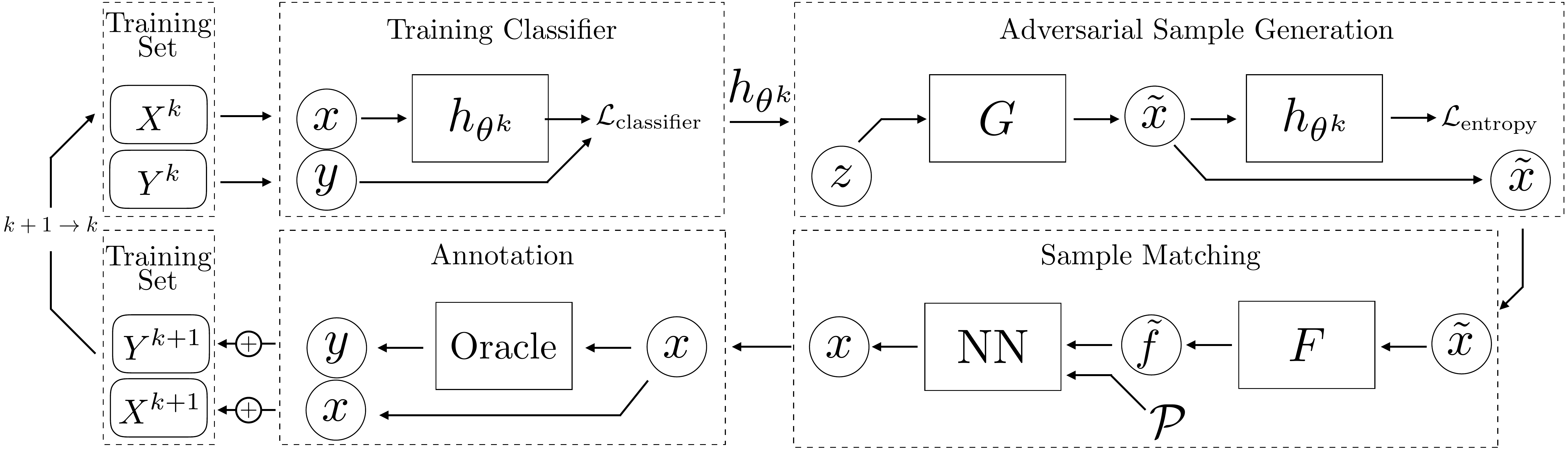}

\caption{\acrshort{asal}: $(X^k, Y^k)$ with $(x,y)$ is the training set at cycle $k$, $h_\theta$ is the classifier, $z$ the latent variable, $G$ the generator, $\tilde{x}$ the synthetic samples, $F$ the feature extractor, $\tilde f$ the features, $\mathcal{P}$ the pool and NN the nearest neighbour method.}
\label{fig:blockdiagram}
\end{figure*}
\section{Proposed \acrlong{asal}}\label{sec:asal}

\gls{asal} allows using \glspl{gan} for fast pool based active learning by generating samples and retrieving similar real samples from the pool, that will be labelled and added to the training set.
Fig.~\ref{fig:blockdiagram} shows the main components of the proposed \gls{asal}. $(X^k, Y^k)$ denote the data set, used to train the classifier $h$ with weights $\theta^k$ at active learning cycle $k$.
Then, the trained classifier $h_{\theta^k}$ and the generator $G$ enable producing uncertain samples $\tilde{x}$. The feature extractor $F$ computes features that the nearest neighbour model uses to retrieve the most similar real samples from the pool. Finally, an oracle annotates the new real samples from the pool and adds them to the training set. Then, the new \gls{al} cycle $k+1$ starts. 

In the remainder of this section, we introduce the adversarial sample generation and the sample matching strategy.

\subsection{Adversarial Sample Generation using GANs}
Instead of selecting uncertain samples from the pool, we follow Zhu~\etal~\cite{zhu2017} and generate such samples using a trained \gls{gan}. Such a \gls{gan} enables to approximate the underlying data distribution of the pool. The discriminator $D$ ensures that the samples drawn from the generator $G$ are indistinguishable from real samples. At convergence, the generator produces the function $G: \mathbb{R}^n \rightarrow \mathcal{X}$ that maps the latent space variable $z \sim \mathcal{N}(\mathbf{0}_n,\mathbf{I}_n)$ to the image domain $\mathcal{X}$. The optimization problem that describes sample generation reads as follow:

\begin{equation}\label{eq:max-ent-gan}
    \begin{aligned}
    & \text{maximize} & & (H \circ h_{\theta^k})(x) \\
    & \text{subject to} & & x = G(z),
    \end{aligned}
\end{equation}

where $H(q) \coloneqq -\sum_{i=1}^m P(c = i|q) \log [P(c = i|q)]$ and $m$ is the number of categories. Removing the constraint $x\in \mathcal{P}$ by including the generator simplifies the problem but changes its solution.
New samples are no longer selected from the pool but are artificially generated.
We solve the optimization problem in two steps:
(i) we use the chain rule and gradient descent to minimize the objective with respect to $z$ and
(ii) we use $G$ to recover a synthetic sample $x$ from $z$. Thus, solving problem~\eqref{eq:max-ent-gan} has a constant run-time complexity $\mathcal{O}(1)$ because it is independent of the pool size. Note, that traditional uncertainty sampling achieves a linear run-time complexity $\mathcal{O}(n)$ (where $n = |\mathcal{P}|$ is the pool size) because it requires scanning each sample in the pool $\mathcal{P}$.

\subsection{Sample Matching}\label{sec:sample-matching}

Sample matching compares real pool samples to generated synthetic samples in a feature space and retrieves the closest matches. Therefore, we require a feature extractor $F: \mathcal{X} \rightarrow \mathcal{F}$, that maps data samples to a feature space, and a distance function $d: \mathcal{F} \times \mathcal{F} \rightarrow \mathbb{R}^+_0$, that computes the distance between two data samples in this feature space.

\textbf{Feature Extractors} After the model generated meaningful features the task is selecting similar samples from the pool. In order to find suitable matches we need a feature space where nearby samples achieve a similar entropy (if entropy is the AL score used for sample generation). 
Naturally we would use the output of the $(l-1)$ layers of a CNN with $l$ layers as features because entropy depends on these features : $F_{\textrm{CLS}}(x) = h^{l-1}_{\theta^k}(x)$. Unfortunately, the feature representation of each data sample becomes obsolete as soon as the weights $\theta$ of the CNN are updated. Thus, using the classifier to extract features for sample matching requires recomputing the feature representation of each data sample after each training iteration. This leads to a linear run-time complexity $\mathcal{O}(n)$.
Thus, we propose to use feature extractors that are independent of the current weights $\theta^k$ such that we need to compute the features for each data sample only once in a pre-processing step. A feature space independent of the classifier does not guarantee entropy smoothness,~\ie samples with a nearby representation may have different
entropy. However, perfect matches will have exactly the same entropy. Therefore, the closer the
representations, the more likely they will score a similar entropy. However, this requires representative features for both: the real samples and the synthetic samples. Furthermore, we require, that the data set is
sufficiently dense because for a sparse data set even the closest matches could be far away.

The image space is a simple feature space that uses the raw values of each pixel as one feature (RGB or gray values) $F_\textrm{gray/rgb}(x) = x$.
The drawback is its large number of dimensions and that two visually close images with similar entropy that for example differ because of background intensity, small noise component, different scaling or small translations lead to far apart representations.
Hence, we require a feature extractor that is mostly invariant to such perturbations. Thus, we propose to use the encoder of an auto-encoder to extract data set specific features. Furthermore, we can train these methods to extract features that are invariant to small perturbations in input images. We define the encoder and the decoder as follows $\phi: \mathcal{X} \rightarrow \mathcal{F}$ and
$\psi: \mathcal{F} \rightarrow \mathcal{X}$ and minimize $\sum_{x \in X} \|x - (\phi \circ \psi)(x)\|_2^2$ to train the encoder and decoder. Thus, the feature extractor reads as: $F_{\textrm{auto}}(x) = \phi(x)$.
Another feature spaces is defined by the features extracted by the discriminator. Training a \gls{gan} includes a discriminator that uses data-set specific features to solve the task of differentiating synthetic from real samples by assigning each input sample a probability how likely it is real or fake, $D: \mathcal{X} \rightarrow [0,1]$. Thus, the features of the discriminator are not only suitable for real but also for synthetic samples and are data set specific. We propose to use the output of the $(j-1)$ layer of a discriminator with $j$ layers as features: $F_{\textrm{disc}}(x) = D^{j-1}(x)$.

\begin{algorithm}[t]
    \caption{ASAL}
    \KwIn{Initialize the set $X,Y$ by adding random pool samples to $X^0$ and their labels to $Y^0$. Train the generator $G$ and the feature extractor $F$. Precompute the PCA, $\mu$ and the set $\mathcal{S} = \{F_\textrm{PCA}(x) \mid x \in X\}$.}
    \KwResult{Trained Classifier $h_{\theta_k}$}
    \Repeat {Labelling budget is exhausted} 
    {
        \begin{enumerate}
            \item Train classifier $h_{\theta^k}$ to minimize empirical risk\\$R(h_{\theta^k}) = \frac{1}{|X^k|}\sum_{(x,y) \in (X^k,Y^k)} l(h_{\theta^k}(x), y)$.
            \item Generate synthetic samples $\tilde x$ with high\\entropy by solving Eq.~\eqref{eq:max-ent-gan}.
            \item Compute the feature representations $\tilde f$ of the generated samples:
            $ \tilde f = F_\textrm{PCA}(\tilde x)$
            \item Retrieve real samples $x$ that match $\tilde x$\\ $x = \{p_i \in \mathcal{P} \mid i = \argmin_{f \in \mathcal{S}} d(f, \tilde{f})\}$.
            \item Annotate the samples $x$ with labels $y$.
            \item Update the sets $X^{k+1} = X^k \cup \{x\}$, $Y^{k+1} = Y^k \cup \{y\}$.
        \end{enumerate}
    }
    \label{alg:asal}
\end{algorithm}

\begin{figure}[t]
\centering 
\def\svgwidth{0.7\columnwidth}
{
    \fontsize{5pt}{7pt}\selectfont
    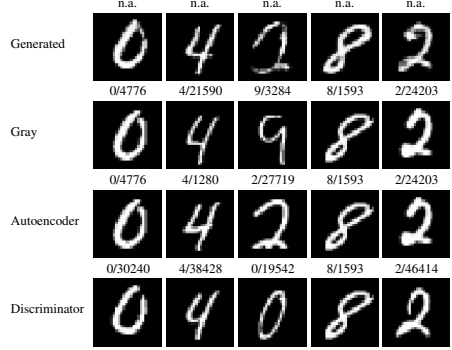
}
\caption{The rows show either generated or matched samples using different feature sets for \emph{MNIST - ten classes}. The brackets denote (label id / sample id).}
\label{fig:mnist-matching}
\end{figure}

\textbf{Efficient Feature Matching} In order to find the best match we extract the features of the synthetic sample. Then, we retrieve the pool sample that has the most similar feature representation with respect to the distance function $d$. Sample matching reads as follows: $x = \argmin_{x \in X} d(F(x), F(\tilde x))$, where $\tilde x$ is a synthetic sample. We propose to use the euclidean distance function $d(f_1, f_2) = \|f_1 - f_2\|_2$. This problem is equivalent to finding the nearest-neighbour of the synthetic sample in feature space.
Therefore, we use multi-dimensional binary search trees (k-d-trees)~\cite{bentley1975} for efficient nearest neighbour selection because their run-time complexity to search a nearest neighbour is sub-linear $\mathcal{O}(\log n)$
with respect to the pool size $n = |\mathcal{P}|$.
However, the run time depends on the number of dimension of the feature space. Therefore, we achieve fast feature matching if $\textrm{dim}(\mathcal{F}) \ll \textrm{dim}(\mathcal{X})$. A property of auto-encoders is that they allow to compress samples from a high dimensional input spaces into a much lower dimensional latent space such that they enable fast sample matching. However, all other discussed feature spaces have typically a similar number of dimensions as the image space. Therefore, we propose \gls{pca} to reduce the number of dimensions and to produce fewer features that contain most of the variance: $F_\textrm{PCA}(x) = \textrm{PCA}(F(x) - \mu)$, where $\mu = \frac{1}{|X|}\sum_{x\in X}F(x)$ and $F$ is one of the previously introduced feature extractors. Then, the full sample matching reads as

\begin{equation}
    x = \bigg\{p_i \in \mathcal{P} \,\bigg | \,i = \argmin_{f \in S} d(f, F_\textrm{PCA}(\tilde x)\bigg\},
\end{equation}
where $S = \{F_\textrm{PCA}(x) \mid x \in X\}$ can be precomputed before starting \gls{al}. We compress the features independent of the extractor to the same number of dimensions using \gls{pca} to ensure very similar absolute run-times of the nearest neighbour method.
Alg.~\ref{alg:asal} shows a detailed description of the different steps of \gls{asal} and Fig.~\ref{fig:mnist-matching} shows examples of synthetic samples with high entropy and their closest matches using the proposed matching with different features. We show more examples in the supplementary.

\begin{table*}[t]
  \renewcommand{\arraystretch}{1.2}
  \newcommand{\head}[1]{\textnormal{\textbf{#1}}}

  \colorlet{tableheadcolor}{gray!25} 
  \newcommand{\headcol}{\rowcolor{tableheadcolor}} %
  \colorlet{tablerowcolor}{gray!10} 
  \newcommand{\rowcol}{\rowcolor{tablerowcolor}} %
  \setlength{\aboverulesep}{0pt}
  \setlength{\belowrulesep}{0pt}

  \newcommand*{\rulefiller}{
    \arrayrulecolor{tableheadcolor}
    \specialrule{\heavyrulewidth}{0pt}{-\heavyrulewidth}
    \arrayrulecolor{black}}
  \centering
  \caption{Summary of all experiments. We run the experiment on CelebA four times but consider a different face attribute each time. \textit{Budget} denotes the maximum amount of samples in the AL data set, \textit{New} denotes the number of newly labelled samples in each AL cycle and \textit{Initial} denotes the number of samples in the data set when AL begins. These three number allow computing the number of AL cycles.}
  \label{table:experiments}
  \resizebox{0.85\linewidth}{!}{
  \begin{tabular}{lcrcccrrrrr}
    \toprule
    \headcol \head{Data set} & \head{Classes} & \head{Train Size} & \head{Classifier} & \head{GAN} & \head{Feature matching} & \head{Budget} & \head{New} & \head{Initial} & \head{AL Cycles} & \head{Seeds}\\
    \midrule
    
    MNIST    & two  & 10k   & Linear     & WGAN-GP & Gray/Disc/Auto   & 500    & 10            & 50             & 45  & 5      \\
    CIFAR-10 & two  & 10k   & Linear     & WGAN-GP & RGB/Disc/Auto    & 1000   & 10            & 50             & 95  & 5      \\
    \midrule
    MNIST    & ten  & 50k   & CNN        & WGAN-GP & Gray/Disc/Auto   & 10k    & 50            & 100            & 198 & 5      \\
    CIFAR-10 & ten  & 50k   & CNN        & WGAN-CT & RGB/Disc/Auto    & 30k    & 1000          & 1000           & 29  & 3      \\
    CelebA   & two  & 160k  & CNN        & WGAN-GP & Auto-Encoder     & 2k     & 10            & 1000           & 190 & 5      \\
    SVHN     & ten  & 604k  & CNN        & WGAN-GP & Auto-Encoder     & 50k    & 1000          & 1000           & 49  & 3      \\
    LSUN     & ten  & 9604k & CNN        & WGAN-GP & Auto-Encoder     & 30k    & 1000          & 1000           & 29  & 3      \\
    \bottomrule
  \end{tabular}
  }
\end{table*}

\section{Experiments}\label{sec:experiments}

\subsection{Datasets}
For the experiments we use five datasets: MNIST~\cite{lecun1998}, CIFAR-10~\cite{krizhevsky2009}, CelebA~\cite{liu2015}, SVHN~\cite{netzer2011svhn} and LSUN Scenes~\cite{yu2015lsun}. 
The MNIST data set contains ten digits unevenly distributed. Each image has a resolution of $28\times28$ gray-scale pixels. The data set consists of 50k training, 10k validation and 10k testing samples. CIFAR-10 consists of 50k training and 10k validation $32\times32$ color images with uniformly distributed label categories. We use the validation set for testing.
CelebA consists of roughly 160k training, 20k validation and 20k testing $64\times64$ color images and a list specifying the presence or absence of 40 face attributes for each image. SVHN consists of 73k training, 26k testing and 531k extra $32\times32$ color images with unevenly distributed label categories. We use the training and extra images to build the pool for \gls{al}. LSUN Scenes consists of roughly 10M training images with unevenly distributed labels. We split it into training and testing sets and centrally crop all the images to $64\times64$ color images.

For a fair comparison to \gls{gaal}, we follow Zhu and Bento~\cite{zhu2017} and construct the same binary data sets, consisting of the MNIST digits 5 \& 7 and the CIFAR-10 classes \textit{automobile} \& \textit{horse}. In addition we validate \gls{adba} on the same MNIST data set.
Furthermore, we validate \gls{asal} on the full MNIST, CIFAR-10, SVHN and LSUN data sets and use four face attributes to build four different CelebA classification benchmarks.
Each benchmark contains all 200k images, labelled according to the presence or absence of the attribute: \textit{Blond\_Hair, Wearing\_Hat, Bangs, Eyeglasses}.

\subsection{Experimental Settings}
\begin{figure*}[t]
\centering 
\vspace{-10pt}
\subfloat[Accuracy (MNIST - two classes)]{\includegraphics[height=0.23\textwidth, keepaspectratio]{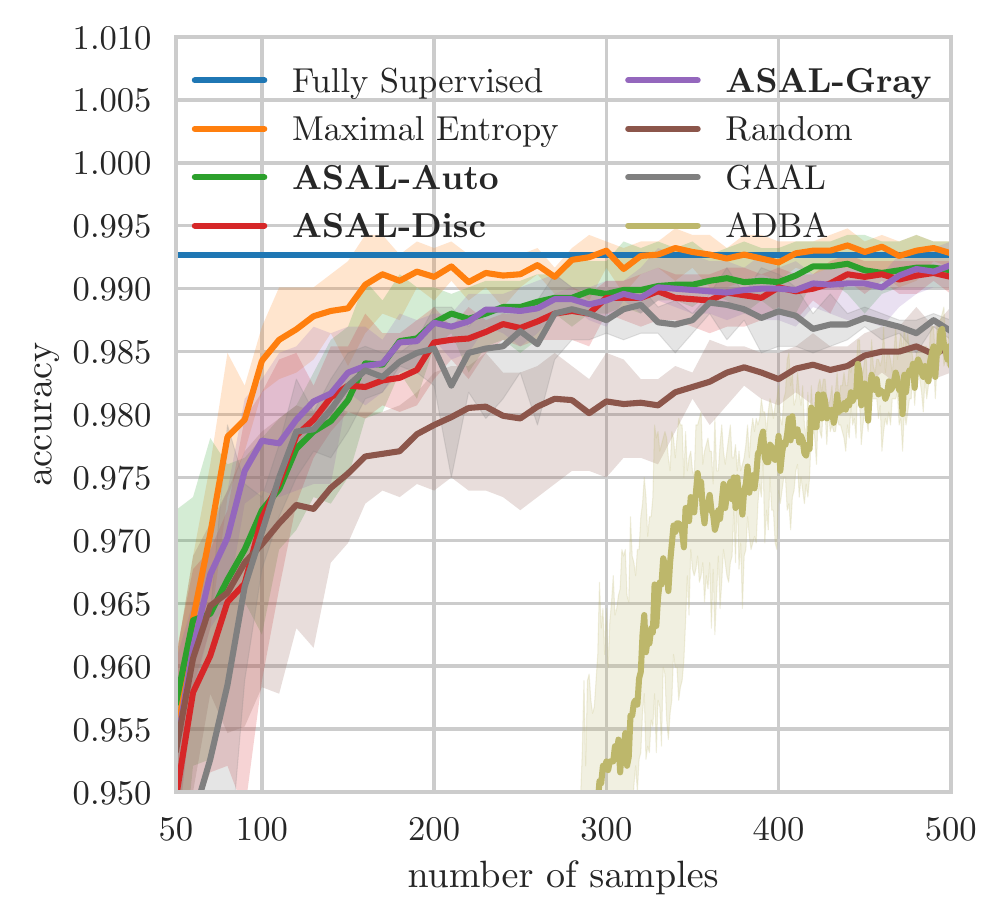}\label{fig:mnist-binary-test}}
\subfloat[Entropy of new samples.]{\includegraphics[height=0.23\textwidth,keepaspectratio]{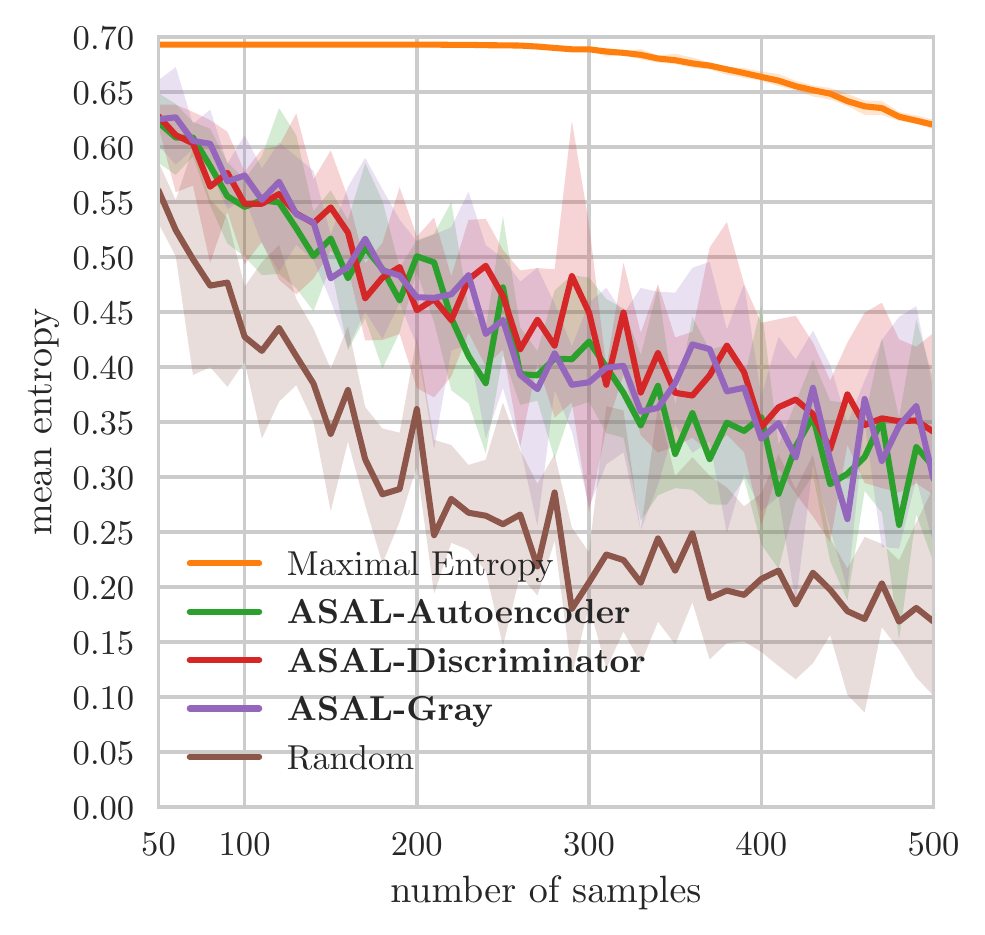}\label{fig:mnist-binary-entropy}}
\subfloat[Accuracy (MNIST - ten classes)]{\includegraphics[height=0.23\textwidth, keepaspectratio]{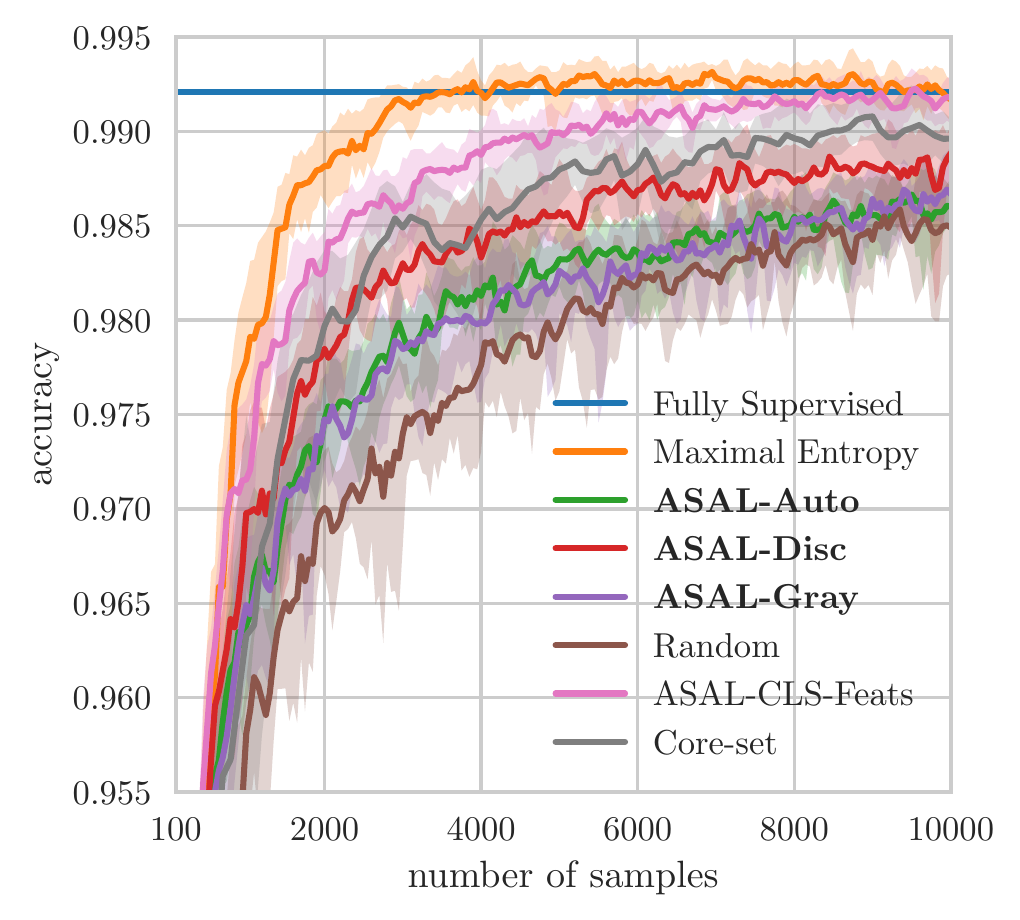}\label{fig:mnist-all-test}}
\subfloat[Small ablation study for PCA.]{\includegraphics[height=0.23\textwidth, keepaspectratio]{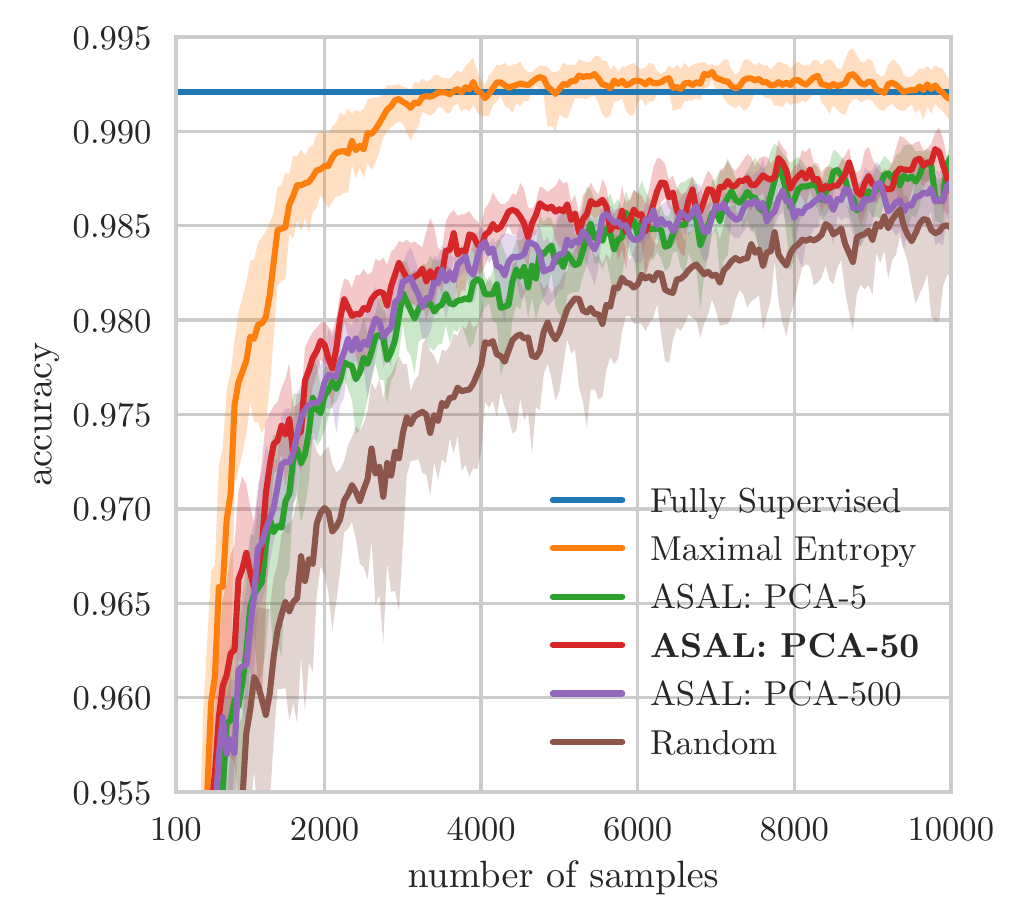}\label{fig:mnist-all-pca}}

\caption{Test accuracy and entropy for different methods, data sets and benchmarks.}

\end{figure*}

First, we produce different references to assess the performance of \gls{asal}. (i) \textit{Maximum-entropy sampling} (upper bound) because \gls{asal} tries to approximate this strategy in sub-linear run-time complexity. (ii) \textit{Random sampling} (lower bound, baseline) and (iii) the \textit{fully supervised model} (upper bound). In addition we report for a subset of the experiments the results of Core-set based \gls{al} (MNIST \& SVHN). We examine three different versions of \gls{asal} using the previously introduced set of features: \emph{\gls{asal}-Gray/RGB},  \emph{\gls{asal}-Autoencoder}, and \emph{\gls{asal}-Discriminator}. For some settings we compare to \emph{\gls{asal}-CLS-Features} that uses the classifier features for matching. We reduce the dimension of the feature space to 50 using \gls{pca}. We experimentally verified that more dimensions only increase the run-time but lead to similar accuracy. Fig.~\ref{fig:mnist-all-pca} shows the test accuracy of \emph{\gls{asal}-Autoencoder} with three different number of \gls{pca} dimensions. To synthesize new samples we use Adam~\cite{kingma2014} and apply 100 gradient steps to maximize the entropy with respect to the latent space variable, see Eq.~\eqref{eq:max-ent-gan}. We directly optimize for multiple latent space variables at the same time by embedding them in one batch. We always draw samples from the pool without replacement. We do not use data augmentation for any experiment except LSUN and train all models from scratch in each \gls{al} cycle. We run all experiments for five different runs with different random seeds and report the mean (solid line) except the computationally demanding experiments on \emph{CIFAR-10 - ten classes} SVHN and LSUN, that we run for three random seeds. The shaded areas correspond to the maximum and minimum value for each operating point considering all random seeds. 
Please refer to the supplementary for the model architectures (classifiers, auto-encoders, and \glspl{gan}), the training strategies and parameters.
Tab.~\ref{table:experiments} summarizes all experimental setups. For the linear models  ($h(x) = Wx + b$) we use directly the raw pixel values as input features for the classifier. We use Wasserstein \glspl{gan}~\cite{arjovsky2017} with gradient penalty~\cite{gulrajani2017} and add a consistency term~\cite{wei2018} for \emph{CIFAR-10 - ten classes} because it produces synthetic samples with higher visual quality~\cite{wei2018}. Note, that we use the same setup for CelebA for four different experiments but only change the target classification labels.

\begin{figure*}[t]
\centering  
\subfloat[CIFAR-10 - two classes]{\includegraphics[height=0.23\textwidth, keepaspectratio]{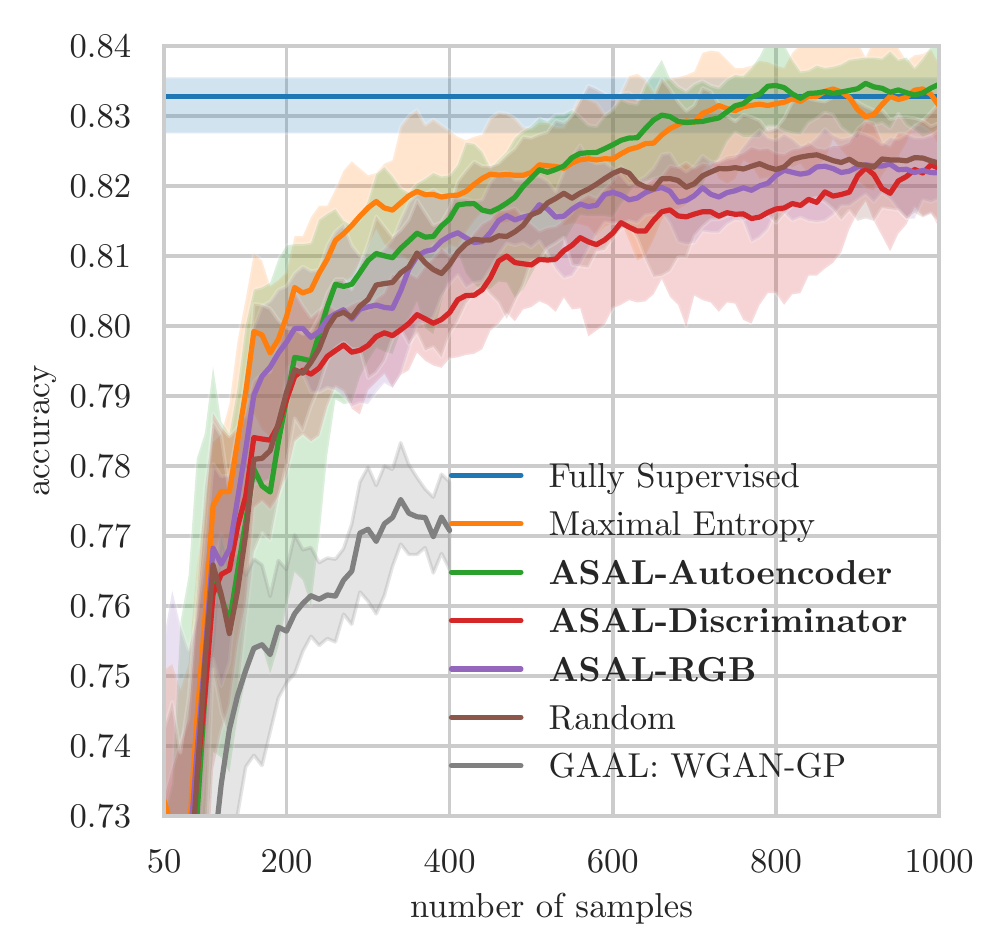}\label{fig:cifar-binary}}
\subfloat[CIFAR-10 - ten classes.]{\includegraphics[height=0.23\textwidth, keepaspectratio]{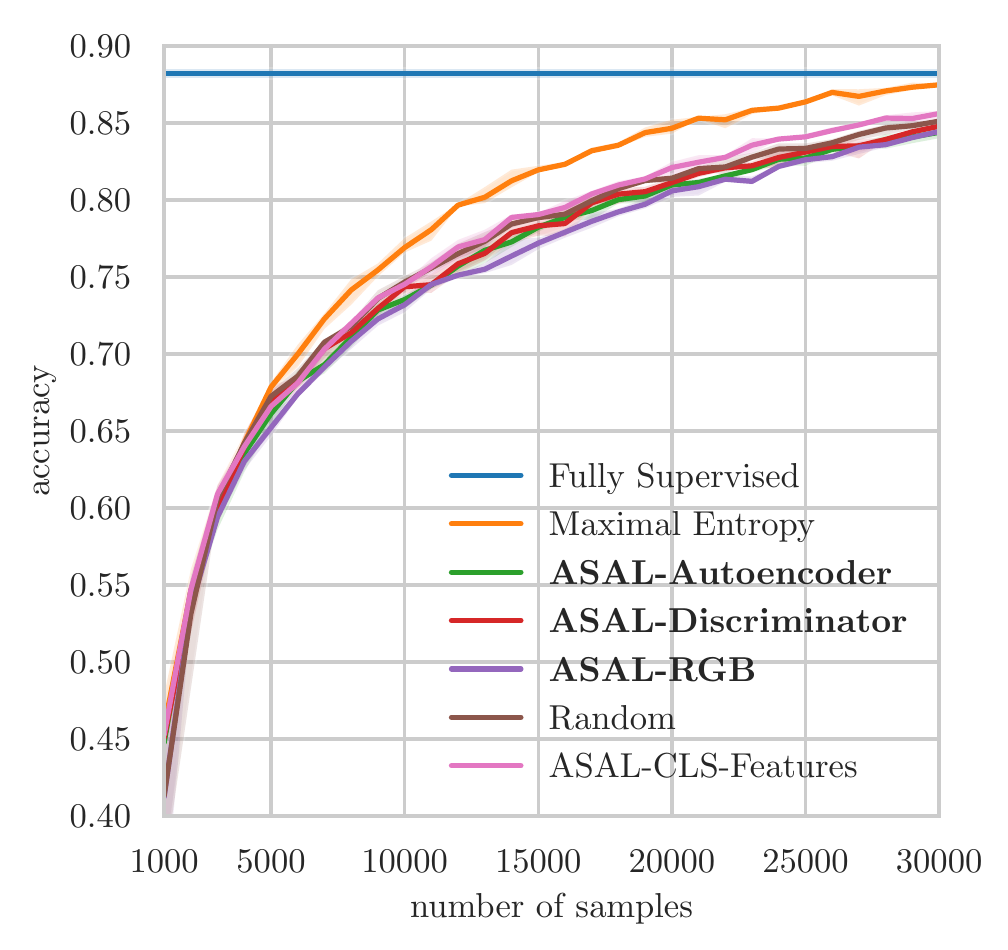}\label{fig:cifar-all-val}}
\subfloat[SVHN - ten classes]{\includegraphics[height=0.23\textwidth, keepaspectratio]{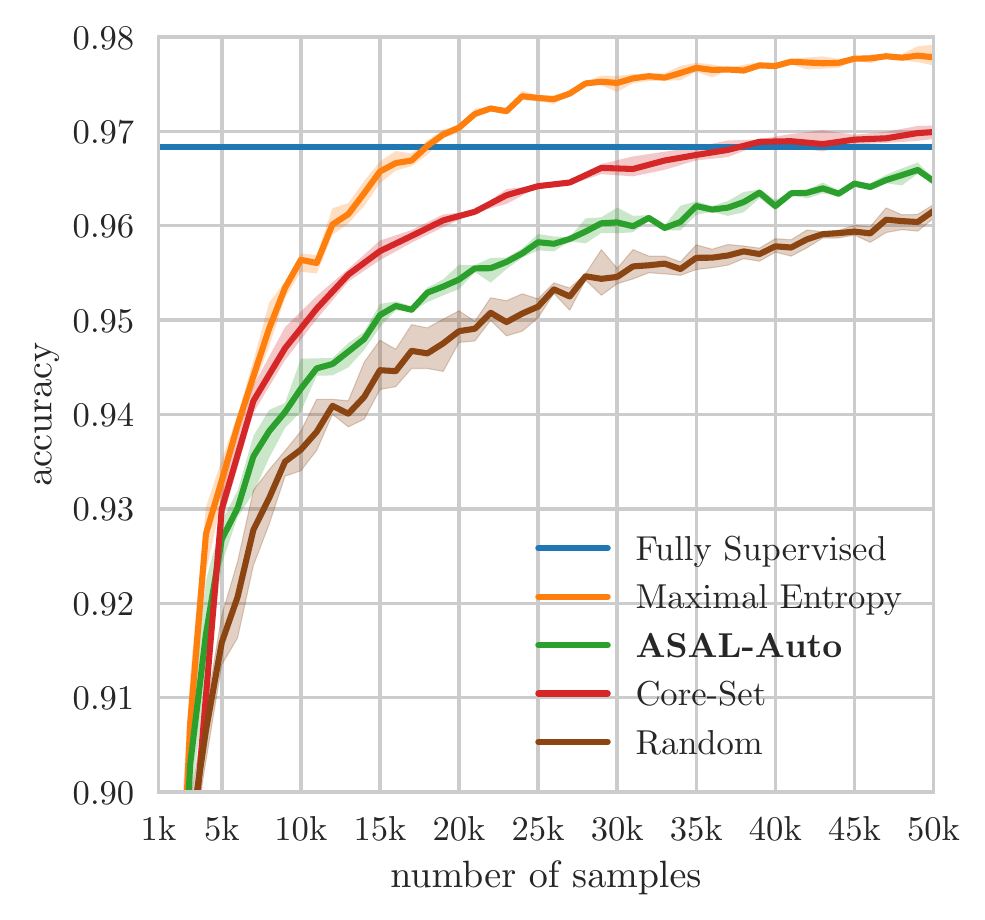}\label{fig:svhn}}
\subfloat[LSUN - ten classes]{\includegraphics[height=0.23\textwidth, keepaspectratio]{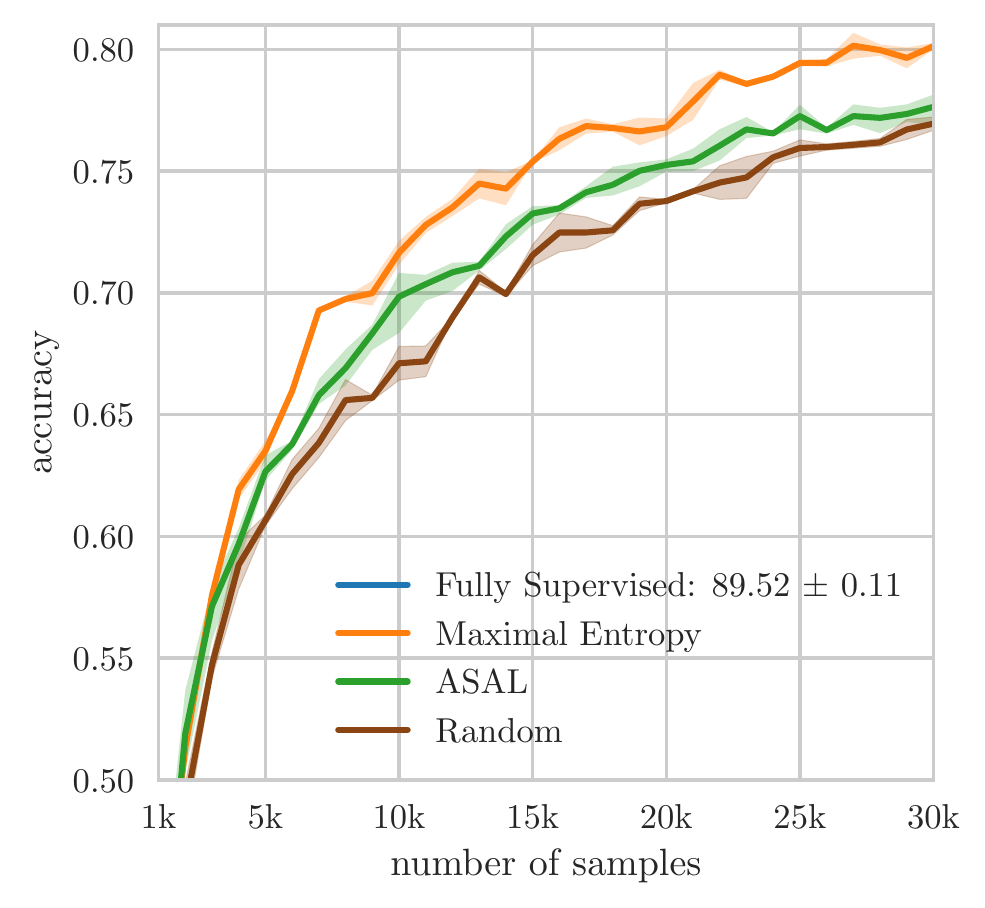}\label{fig:lsun}}

\caption{Comparison of classification accuracy between different methods on four different benchmarks.}
\end{figure*}
\begin{figure*}[t]
\centering 
\vspace{-10pt}
\subfloat[\textit{Wearing\_Hat}]{\includegraphics[height=0.23\linewidth, keepaspectratio]{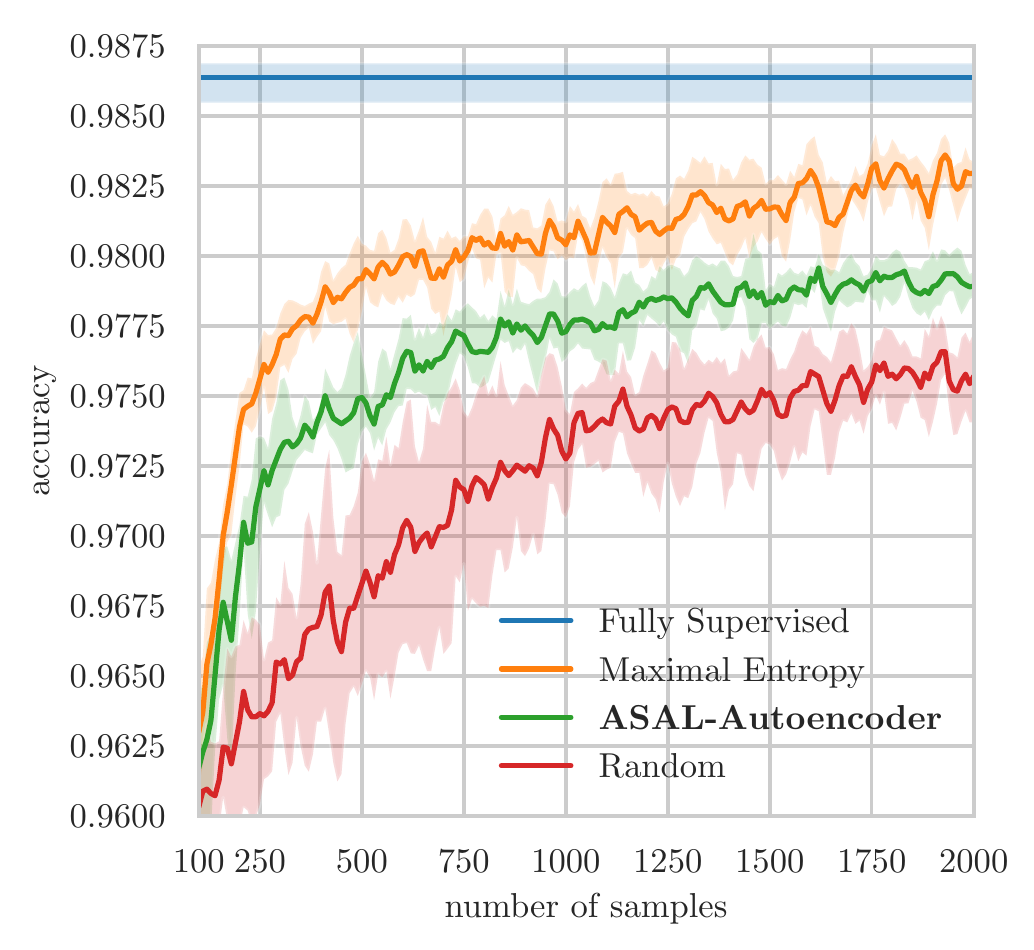}}
\subfloat[\textit{Blond\_Hair}]{\includegraphics[height=0.23\linewidth,keepaspectratio]{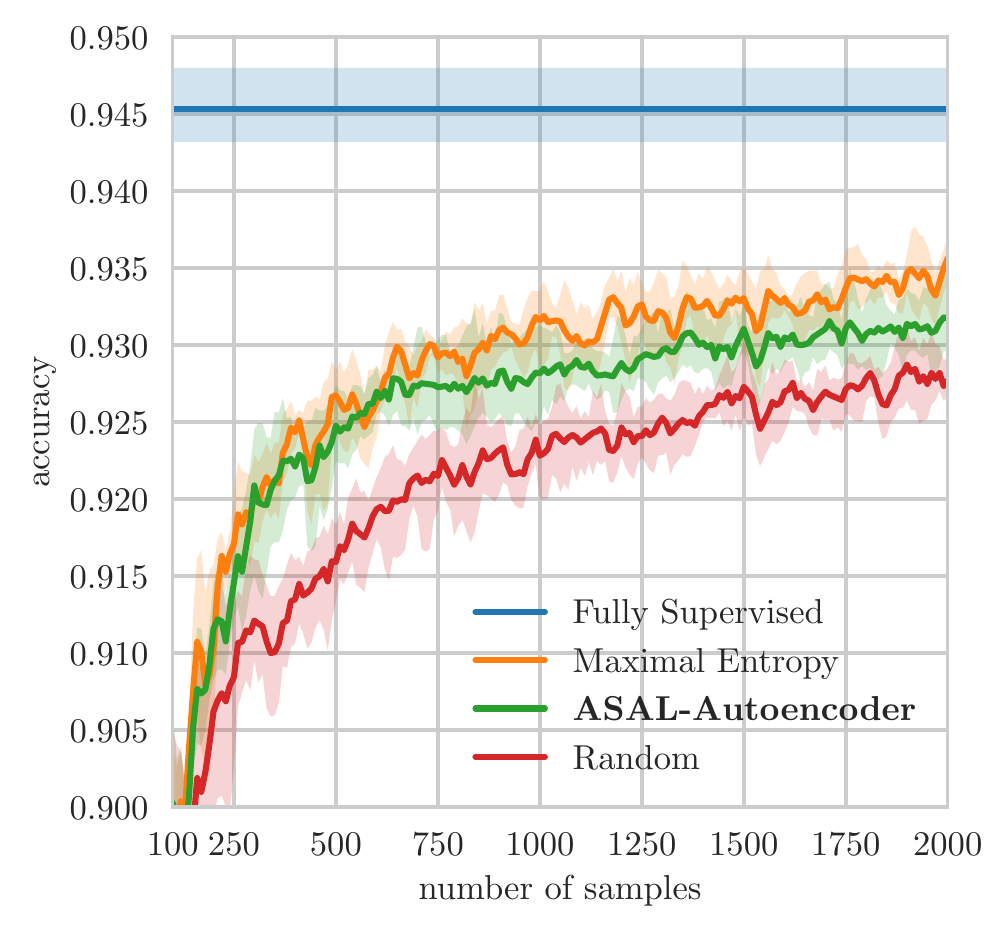}}
\subfloat[\textit{Bangs}]{\includegraphics[height=0.23\linewidth, keepaspectratio]{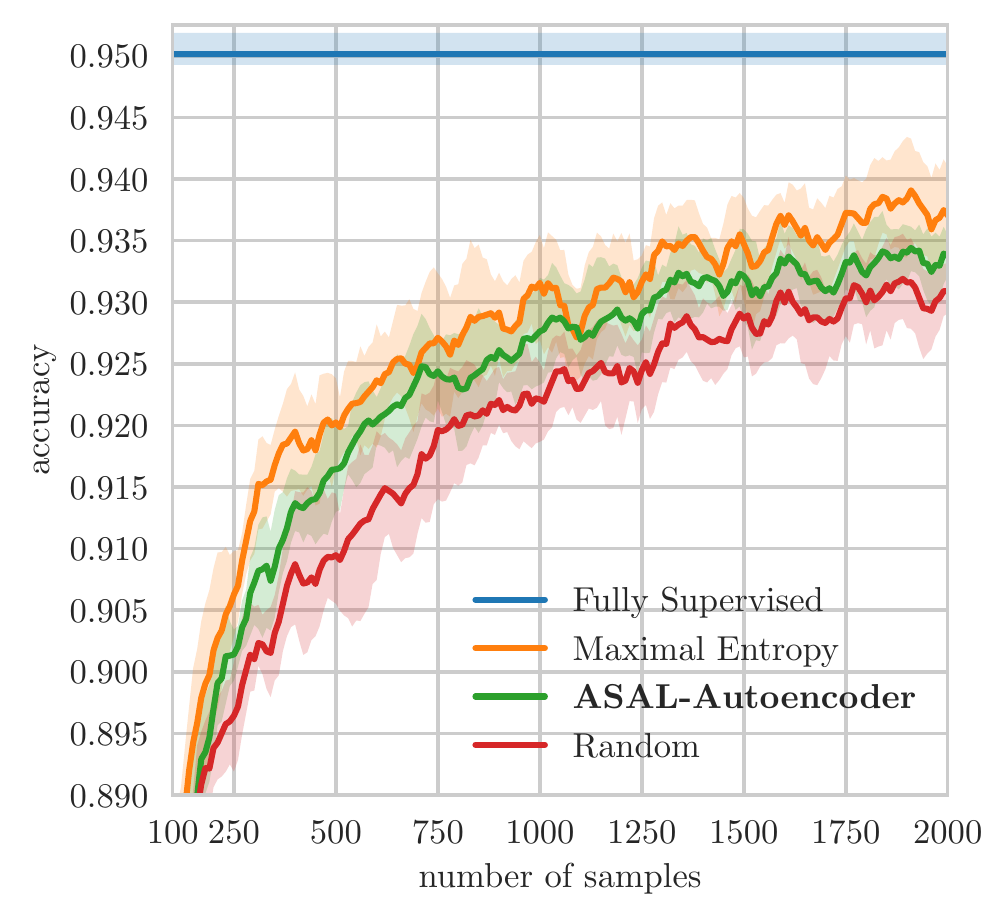}}
\subfloat[\textit{Eyeglasses}]{\includegraphics[height=0.23\linewidth, keepaspectratio]{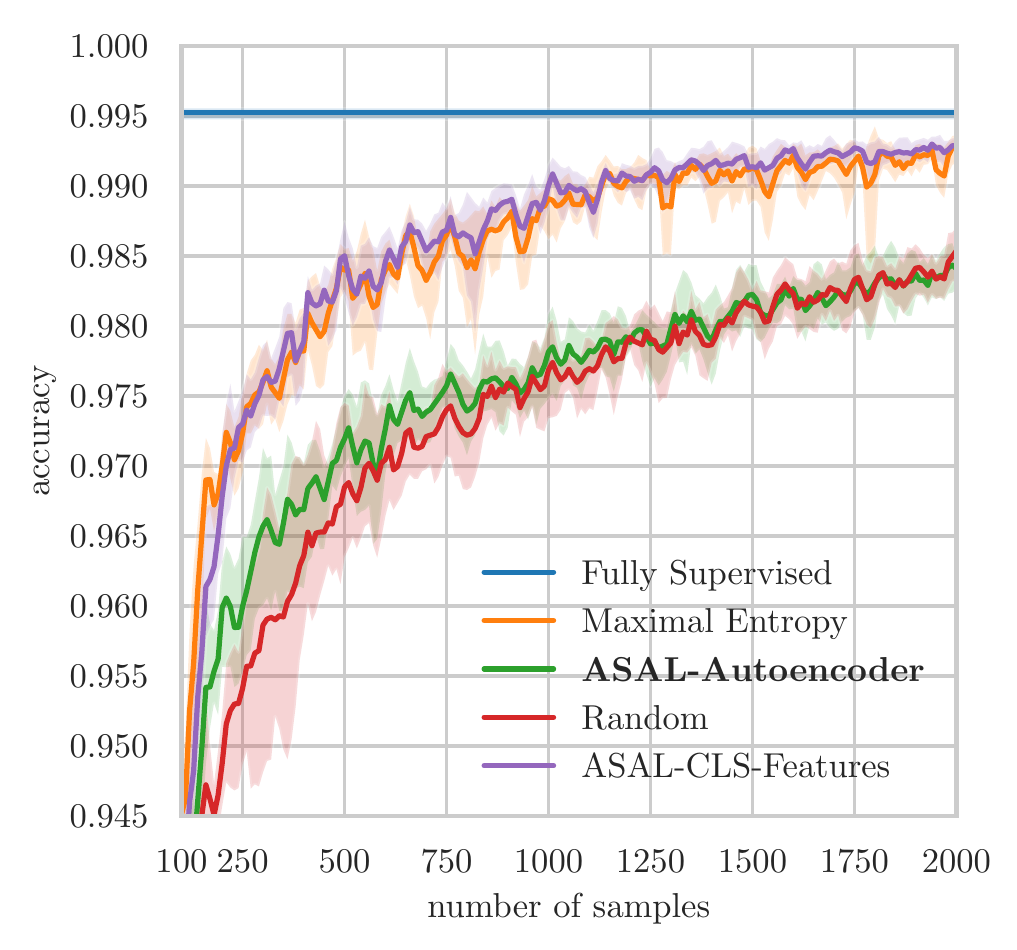}\label{fig:celeba-eyeglasses}}
\caption{Test accuracy on four \textit{CelebA} benchmarks. The target classes correspond to absence or presence of face attributes.}\label{fig:celeba-test}

\end{figure*}

\section{Results}
\subsection{Linear Models}

\gls{asal} outperforms random sampling and approaches maximum entropy sampling quickly on \textit{MNIST - two classes}. We observe, that all three proposed feature spaces for sample matching perform equally well. All \gls{asal} strategies reach a classification accuracy of 98.5\% with only 200 labelled samples, whereas random sampling requires 500 labelled samples, see Fig.~\ref{fig:mnist-binary-test}.

On \textit{CIFAR-10 - two classes} only \emph{\gls{asal}-Autoencoder} exceeds the performance of random sampling. However, using auto-encoder features for sample matching reaches the classification accuracy of maximum entropy sampling already with 500 labelled samples, whereas random sampling requires approximately twice the amount of samples to reach a comparable performance, see Fig.~\ref{fig:cifar-binary}.

\subsection{Convolutional Neural Networks}

\gls{asal} clearly outperforms random sampling on \textit{MNIST - ten classes}. In contrast to the binary setting we used a CNN where the weights and thererfore the extracted features change in each \gls{al} cycle. Nonetheless, all three feature spaces, used for sample matching, exceed the performance of random sampling. However, the discriminator features lead to the highest classification accuracy, see Fig.~\ref{fig:mnist-all-test}. Furthermore, we observe that \emph{\gls{asal}-discriminator}  achieves almost similar test accuracy as core-set based \gls{al} but at a smaller run-time complexity.
Unfortunately, \gls{asal} performs similar to random sampling on \textit{CIFAR-10 - ten classes} independent of feature space used for sample matching but classical uncertainty sampling exceeds the performance of random sampling, see Fig~\ref{fig:cifar-all-val}.
The four experiments using different target labels on CelebA emphasize, that \gls{asal} outperforms random sampling and approaches uncertainty sampling with a better run-time complexity. However, for \textit{CelebA - Eyeglasses} \gls{asal} performs only marginally better than random sampling. \gls{asal} exceeds random sampling during the first cycles but equals its performance when using more than 750 labelled samples, see Fig.~\ref{fig:celeba-test}.
The results on SVHN in Fig~\ref{fig:svhn} show that \gls{asal} outperforms random sampling but achieves lower test accuracy than the more costly core-set based \gls{al} and uncertainty sampling. Similarly, Fig.~\ref{fig:lsun} shows that \gls{asal} outperforms random sampling on LSUN. We omit the comparison with core-set based \gls{al} on LSUN because it is demanding with respect to memory and leads to an even higher run time than uncertainty sampling.
To summarize, \gls{asal} outperforms random sampling on eight out of ten benchmarks. On \textit{CelebA - Blond\_Hair} and \textit{CIFAR-10 - two classes} \gls{asal} achieves almost the same performance as maximum entropy sampling. We will analyze  the successful and the failure cases in 
Sec.~\ref{sec:discussion} and give intuition when \gls{asal} works.
\subsection{Comparison between \acrshort{asal}, \acrshort{gaal} and \acrshort{adba}}

The most similar method to \gls{asal} is \gls{gaal}~\cite{zhu2017}. Even though Zhu \& Bento~\cite{zhu2017} report that \gls{gaal} clearly performs worse than random sampling, we reproduced their results. For fairer comparison we replace their DCGAN with our Wasserstein \gls{gan} that we also use for \gls{asal} and generate images with higher quality. Fig.~\ref{fig:mnist-binary-test} shows that \gls{gaal} achieves a higher accuracy than random sampling and an accuracy almost as high as \gls{asal} at the beginning of \gls{al}. However, after adding more than 350 samples, the classification accuracy does not even saturate but drops and approaches the quality of random sampling. The reason for this decrease are generated samples where identifying the correct labels is very difficult such that the annotations get unreliable. Nonetheless, Fig.~\ref{fig:mnist-matching} shows that labelling synthetic samples is possible and therefore \gls{gaal} can work.
Furthermore, we implement \gls{adba} and use again the same \gls{gan} for sample and line generation. \gls{adba} requires labeling transition points in generated lines of samples such that directly comparing the labeling effort is difficult. We count annotating one line as one annotation. Annotating one line leads to eleven labeled synthetic images. Thus, 500 samples in Fig.~\ref{fig:mnist-binary-test} correspond to 5500 labeled synthetic samples. The classifier is trained in the latent instead of the image space and requires much more annotations to achieve competitive results. Thus, we conclude that \gls{adba} achieves worse performance than all other methods and is limited to binary classification with linear models in latent space. Hence, we omit further comparison with \gls{adba}.

We reproduce \gls{gaal} on \textit{CIFAR-10 - two classes} and observe that reliably labelling uncertain synthetic images is very difficult even with a state-of-the-art \gls{gan}. Fig.~\ref{fig:cifar-binary} reports the performance of \gls{gaal} on \textit{CIFAR-10 - two classes}. We observe that \gls{gaal} performs clearly worse than random sampling and \gls{asal}. We run \gls{gaal} only up to 400 labelled samples because the trend is clear and because manually labelling synthetic images is very costly and tedious. Thus, we conclude, that \gls{asal} outperforms \gls{gaal} and \gls{adba}.

\subsection{Discussion}\label{sec:discussion}

When designing \gls{asal} we make several assumptions and approximations to enable a sub-linear run-time complexity. In this section we analyze the experiments of \gls{asal} and investigate for the failure cases which of these assumptions hold and which do not. We assume that: \textit{(i)} the \gls{gan} can generate synthetic images with high entropy that match the true data distribution, \textit{(ii)} the data set is sufficiently large, such that there exists always a real sample that is close to each synthetic image, \textit{(iii)} there exists a fixed feature space (independent of the classifier), where nearby representations have a similar entropy.

Fig.~\ref{fig:mnist-binary-entropy} shows that all three \gls{asal} strategies retrieve on average samples with 63\% higher entropy than random sampling on \textit{MNIST - two classes}. We conclude that for this data set all assumptions hold. Especially the relatively large and dense data set with 10k samples that cover many variations of the digits enables reliable sample matching and leads to a well trained \gls{gan}.  
In Sec.~\ref{sec:sample-matching} we described the feature extractor $F_\textrm{CLS}$ that uses directly the CNN features. This feature space guarantees entropy smoothness such that nearby representations share a similar entropy. Using the best feature extractor $F_\textrm{CLS}$ increases the run-time but avoids assumption (iii). Thus, if \gls{asal} works only when using $F_\textrm{CLS}$ we require a different feature extractor than the proposed.
Furthermore, if \gls{asal} fails with $F_\textrm{CLS}$ features it indicates that the data set is too small such that training the \gls{gan} to generate uncertain samples with realistic features and matching is unfeasible.
Indeed, Fig.~\ref{fig:mnist-all-test} shows that \gls{asal} on \textit{MNIST - ten classes} using $F_\textrm{CLS}$ approaches quickly the quality of maximum entropy sampling and verifies that $F_\textrm{CLS}$ performs better than fixed features. It shows that using a non-optimal feature extractor reduces the performance but verifies that sample matching with the data set and synthetic samples works. 
We redo the same experiment on \textit{CIFAR-10 - ten classes} and observe that using $F_\textrm{CLS}$ only marginally exceeds the quality of random sampling. Therefore, our proposed sample matching and feature spaces are not the reason why \gls{asal} fails but rather the small data set size. Hence, we expect that either the \gls{gan} generates synthetic samples with unrealistic characteristics or that the generated samples would be useful but close matches are missing in the data set.
We redo the same experiments for the benchmark \textit{CelebA - Eyeglasses} where \gls{asal} fails. However, we already verified that \gls{asal} works on three benchmarks on this data set and know that the quality of the synthetic uncertain images is sufficient. Fig.~\ref{fig:celeba-eyeglasses} shows that $F_\textrm{CLS}$ achieves the same performance as maximum entropy sampling. Hence, the performance drop is caused by using the fixed instead of the varying features. Furthermore, the amount of images in the data set, that contain eyeglasses is very small such that the synthetic image might contain a face with an eyeglass and the matching retrieves a very similar face without eyeglasses. The issue is that the proposed feature extractor concentrates on many face attributes for matching but uncertainty depends only on a small subset.

SVHN is a less diverse data set than CelebA and CIFAR-10 but contains many more samples. Thus, the quality of generated samples is high and similar matching retrieves meaningful samples. Hence, all three assumptions hold. LSUN is the biggest tested data set in this paper but training a \gls{gan} to generate high quality samples is still challenging. Nonetheless, \gls{asal} outperforms \textit{random sampling}. Thus, the \gls{gan} is able to generate samples with features that help training and that are present in the data set too. Furthermore, matching is able to select similar samples because LSUN contains on average 1M samples per class. Thus, we conclude that \gls{asal} can work with lower quality synthetic samples as long as they contain meaningful characteristics because we label matched real instead of generated samples.

\begin{figure}[t]
\centering 
\vspace{-10pt}
\subfloat[Timings for sample selection]{\includegraphics[width=0.5\columnwidth, keepaspectratio]{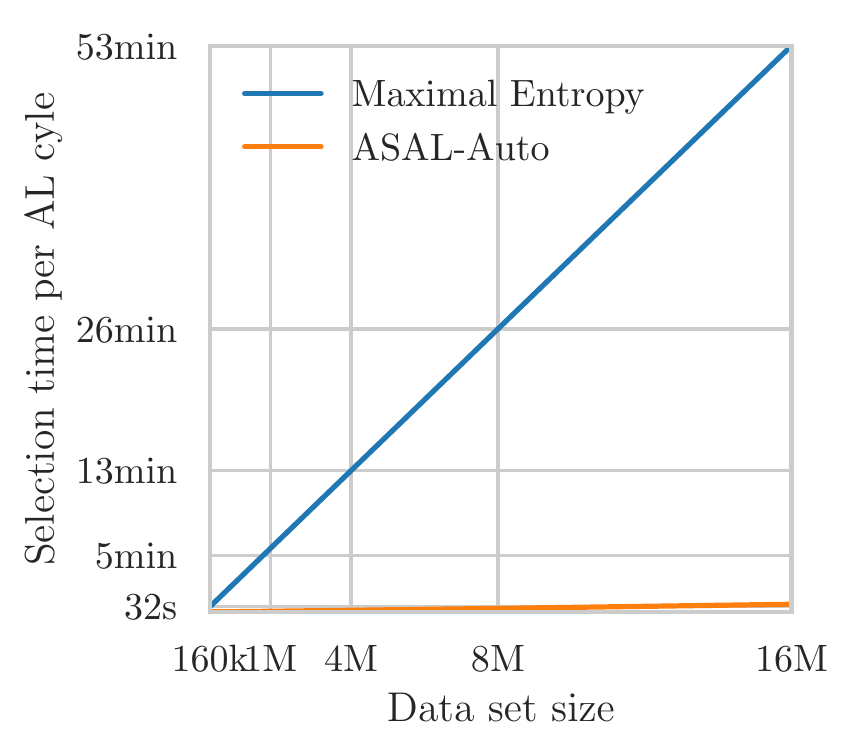}\label{fig:run-time}}
\subfloat[Transition point]{\includegraphics[width=0.5\columnwidth, keepaspectratio]{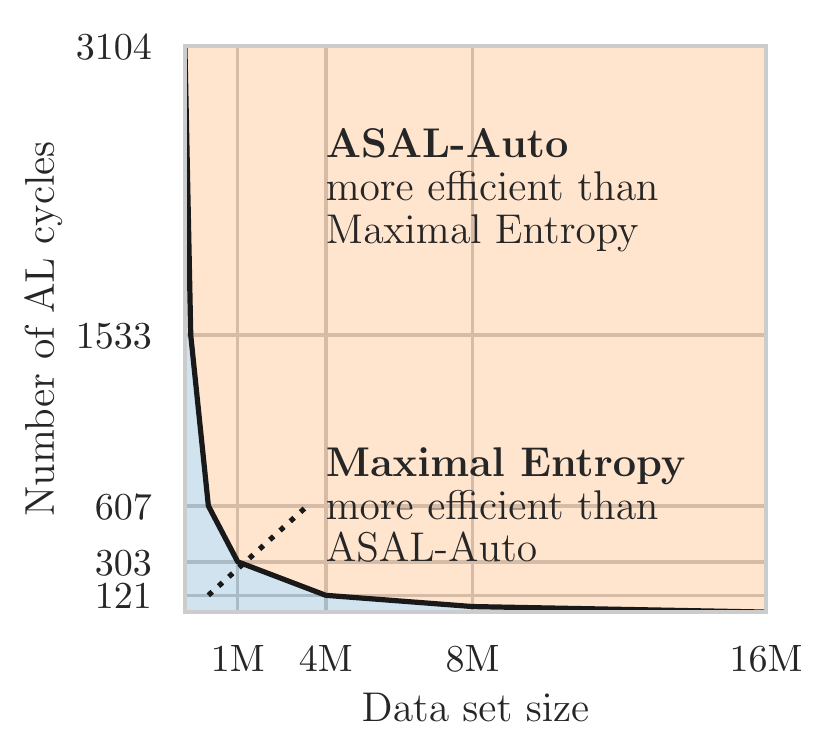}\label{fig:breaking-point}}

\caption{Run-time of uncertainty sampling and \acrshort{asal} to select 10 samples with respect to the data set size. The transition point denotes the number of AL cycles when \acrshort{asal} gets more efficient than maximum entropy sampling}\label{fig:celeba-timings-breaking}

\end{figure}

\subsection{Timings}

In this section we report timings on CelebA and LSUN. For CelebA we first concentrate on the run time of one \gls{al} cycle with respect to different data set sizes. Next, we report the transition point after how many \gls{al} cycles \gls{asal} gets more efficient than \textit{uncertainty sampling} in case pre-processing time is taken into account. Finally, we report the run time of \gls{asal} and other \gls{al} methods on LSUN including I/O-time. All measurements omit classifier training time because it is equivalent for all \gls{al} methods.  We use a Nvidia TITAN X GPU and an Intel Xeon CPU E5-2680 v4.

Fig.~\ref{fig:run-time} reports the time required to select ten new samples in each \gls{al} cycle with respect to data set size. We randomly augmented the original data set (160k) to create larger data sets containing up to 16M samples. Whereas it is possible to keep all images in  memory for 160k (1.98GB) this is hardly possible for 16M images (198GB). For experiments on CelebA we omit including I/O-time. The sample matching proposed in \gls{asal} stores only 50 float features per image (32MB for 160k and 3.2GB for 16M images). This saving enables to keep the features in memory and to build the nearest-neighbor model even for huge data sets.

\gls{asal} has a sub-linear run-time complexity to select new samples. However, it requires several pre-processing steps such as training the \gls{gan} ($\sim25$h) and auto-encoder ($\sim1.6$h), extracting the features ($\sim32$s per 160k samples) and fitting the nearest-neighbor model ($\sim5$min for 16M samples). The sample selection time is $\sim44$s for 16M. Conversely, maximum entropy sampling avoids any pre-processing cost but has a much higher sample selection time: $\sim53$min for 16M. Fig.~\ref{fig:run-time} shows that \gls{asal} is much faster than uncertainty sampling but requires pre-processing.
However, the time savings for \gls{asal} in each \gls{al} cycle is large and allows to compensate for the initial pre-computation time when running \gls{asal} for sufficiently many \gls{al} cycles. 
Fig.~\ref{fig:breaking-point} shows the transition point, the point where \gls{asal} achieves a higher efficiency than \textit{maximum entropy sampling} depending on the data set size and number of \gls{al} cycles.
Note, that the sample selection time for \emph{uncertainty sampling} is independent of the number of selected samples but the run time of \gls{asal} increases when selecting more samples. However, the sample selection time for \gls{asal} is still much smaller than for uncertainty sampling even when querying much more samples. The reason is that we can generate many artificial samples within one batch at once. Note that selecting fewer samples reduces the risk of correlated uncertain samples.

Tab.~\ref{tab:timings-lsun} reports timings for \gls{al} methods including I/O time. We measure the time for \gls{asal}, random and uncertainty sampling. For the other methods we predict the run time based on previous measurements: Learning-Loss~\cite{yoo2019learningloss} requires propagating each sample through the network each \gls{al} cycle and computing the learned loss. Hence, the run-time complexity  is linear and the run time is similar to maximal entropy sampling. Similarly, Core-Set based \gls{al}\cite{sener2018} requires extracting the features for each sample every \gls{al} cycle to select new core samples. Note that in each \gls{al} cycle sample selection takes longer than training a classifier with less than 15k samples for methods with linear run time.

\begin{table}[t]
  \renewcommand{\arraystretch}{1.3}
  \newcommand{\head}[1]{\textnormal{\textbf{#1}}}

  \colorlet{tableheadcolor}{gray!25} 
  \newcommand{\headcol}{\rowcolor{tableheadcolor}} %
  \colorlet{tablerowcolor}{gray!10} 
  \newcommand{\rowcol}{\rowcolor{tablerowcolor}} %
  \setlength{\aboverulesep}{0pt}
  \setlength{\belowrulesep}{0pt}

  \newcommand*{\rulefiller}{
    \arrayrulecolor{tableheadcolor}
    \specialrule{\heavyrulewidth}{0pt}{-\heavyrulewidth}
    \arrayrulecolor{black}}

  \caption{Run-time complexity and runtime for one AL cycle on LSUN including I/O-time ($n = |\mathcal{P}|$ refers to the pool size and $k=|X^k|$ to the number of labeled samples).}
  \label{tab:timings-lsun}
  \centering
  \resizebox{1.\linewidth}{!}{%
  \begin{tabular}{lccr}
    \toprule
    \headcol \head{AL Method} & \head{Feature Extraction} & \head{Sample Selection} & \head{Runtime}\\
    \midrule
    Random & --- & --- & $<1$s\\
    Maximal-Entropy & --- & $O(n)$ & 3660s\\
    Learning-Loss~\cite{yoo2019learningloss} & --- & $O(n)$ & $\sim3660$s\\
    Core-Set~\cite{sener2018} & $O(n)$ & $O(kn)$ & $>3660$s\\
    \midrule
    \textbf{ASAL-Auto (ours)} & $O(1)$ & $O(\log n)$ & 471s\\
    \bottomrule
  \end{tabular}}
\end{table}
\section{Conclusion}

We proposed a new pool-based \gls{al} sampling method that uses sample synthesis and matching to achieve a sub-linear run-time complexity. 
We demonstrated, that \gls{asal} outperforms random sampling on eight out of ten benchmarks. 
We analyzed the failure cases and conclude, that the success of \gls{asal} depends on the structure and size of the data set and the consequential quality of generated images and matches. \gls{asal} works exactly on the large data sets where it is most needed. There the sub-linear run-time compensates quickly for any pre-processing. \gls{asal} is suitable for interactive \gls{al} where pre-processing is acceptable but small sampling times are required. In future research we propose to test \gls{asal} for other \gls{al} scores such as the Learned-Loss~\cite{yoo2019learningloss} to accelerate their run time. Although auto encoder features work well for natural images we suggest to study VGG~\cite{simonyan2015} or AlexNet~\cite{krizhevsky2012} feature in future research.

\nocite{gulrajani2018github}
\nocite{wei2018github}
\nocite{springenberg2014}

{\small
\bibliographystyle{ieee}
\bibliography{egbib}

\begin{thebibliography}{10}\itemsep=-1pt

\bibitem{arjovsky2017}
M.~Arjovsky, S.~Chintala, and L.~Bottou.
\newblock Wasserstein gan.
\newblock {\em arXiv preprint arXiv:1701.07875}, 2017.

\bibitem{bentley1975}
J.~L. Bentley.
\newblock Multidimensional binary search trees used for associative searching.
\newblock {\em Communications of the ACM}, 18(9):509--517, 1975.

\bibitem{campbell2000}
C.~Campbell, N.~Cristianini, A.~Smola, et~al.
\newblock Query learning with large margin classifiers.
\newblock In {\em ICML}, pages 111--118, 2000.

\bibitem{goodfellow2014}
I.~Goodfellow, J.~Pouget-Abadie, M.~Mirza, B.~Xu, D.~Warde-Farley, S.~Ozair,
  A.~Courville, and Y.~Bengio.
\newblock Generative adversarial nets.
\newblock In {\em Advances in neural information processing systems}, pages
  2672--2680, 2014.

\bibitem{gulrajani2018github}
I.~Gulrajani.
\newblock {improved\_wgan\_training}.
\newblock \url{\\https://github.com/igul222/improved\_wgan\_training}, 2018.

\bibitem{gulrajani2017}
I.~Gulrajani, F.~Ahmed, M.~Arjovsky, V.~Dumoulin, and A.~C. Courville.
\newblock Improved training of wasserstein gans.
\newblock In {\em Advances in Neural Information Processing Systems}, pages
  5767--5777, 2017.

\bibitem{hospedales2013}
T.~M. Hospedales, S.~Gong, and T.~Xiang.
\newblock Finding rare classes: Active learning with generative and
  discriminative models.
\newblock {\em IEEE transactions on knowledge and data engineering},
  25(2):374--386, 2013.

\bibitem{huijser2017}
M.~Huijser and J.~C. van Gemert.
\newblock Active decision boundary annotation with deep generative models.
\newblock In {\em The IEEE International Conference on Computer Vision (ICCV)},
  Oct 2017.

\bibitem{jain2010}
P.~Jain, S.~Vijayanarasimhan, and K.~Grauman.
\newblock Hashing hyperplane queries to near points with applications to
  large-scale active learning.
\newblock In {\em Advances in Neural Information Processing Systems}, pages
  928--936, 2010.

\bibitem{joshi2009}
A.~J. Joshi, F.~Porikli, and N.~Papanikolopoulos.
\newblock Multi-class active learning for image classification.
\newblock In {\em Conference on Computer Vision and Pattern Recognition}, pages
  2372--2379. IEEE, 2009.

\bibitem{kingma2014}
D.~P. Kingma and J.~Ba.
\newblock Adam: A method for stochastic optimization.
\newblock In {\em Proceedings of the International Conference on Learning
  Representations}, 2015.

\bibitem{krizhevsky2009}
A.~Krizhevsky.
\newblock Learning multiple layers of features from tiny images.
\newblock 2009.

\bibitem{krizhevsky2012}
A.~Krizhevsky, I.~Sutskever, and G.~E. Hinton.
\newblock {ImageNet Classification with Deep Convolutional Neural Networks}.
\newblock In F.~Pereira, C.~J.~C. Burges, L.~Bottou, and K.~Q. Weinberger,
  editors, {\em Advances in Neural Information Processing Systems 25}, pages
  1097--1105. Curran Associates, Inc., 2012.

\bibitem{lecun1998}
Y.~LeCun, L.~Bottou, Y.~Bengio, and P.~Haffner.
\newblock Gradient-based learning applied to document recognition.
\newblock {\em Proceedings of the IEEE}, 86(11):2278--2324, 1998.

\bibitem{liu2015}
Z.~Liu, P.~Luo, X.~Wang, and X.~Tang.
\newblock Deep learning face attributes in the wild.
\newblock In {\em Proceedings of International Conference on Computer Vision
  (ICCV)}, 2015.

\bibitem{miyato2018vat}
T.~Miyato, S.-i. Maeda, S.~Ishii, and M.~Koyama.
\newblock Virtual adversarial training: a regularization method for supervised
  and semi-supervised learning.
\newblock {\em IEEE transactions on pattern analysis and machine intelligence},
  2018.

\bibitem{netzer2011svhn}
Y.~Netzer, T.~Wang, A.~Coates, A.~Bissacco, B.~Wu, and A.~Y. Ng.
\newblock Reading digits in natural images with unsupervised feature learning.
\newblock In {\em Workshop on Deep Learning and Unsupervised Feature Learning,
  NIPS}, 2011.

\bibitem{nguyen2004}
H.~T. Nguyen and A.~Smeulders.
\newblock Active learning using pre-clustering.
\newblock In {\em Proceedings of the twenty-first international conference on
  Machine learning}, page~79. ACM, 2004.

\bibitem{radford2015}
A.~Radford, L.~Metz, and S.~Chintala.
\newblock Unsupervised representation learning with deep convolutional
  generative adversarial networks.
\newblock In {\em Proceedings of the International Conference on Learning
  Representations}, 2015.

\bibitem{sener2018}
O.~Sener and S.~Savarese.
\newblock Active learning for convolutional neural networks: A core-set
  approach.
\newblock In {\em International Conference on Learning Representations}, 2018.

\bibitem{simonyan2015}
K.~Simonyan and A.~Zisserman.
\newblock {Very Deep Convolutional Networks for Large-Scale Image Recognition}.
\newblock In {\em International Conference on Learning Representations (ICLR)},
  pages 1--14, 2015.

\bibitem{springenberg2014}
J.~T. Springenberg, A.~Dosovitskiy, T.~Brox, and M.~Riedmiller.
\newblock Striving for simplicity: The all convolutional net.
\newblock In {\em Proceedings of the International Conference on Learning
  Representations Workshops}, 2014.

\bibitem{tong2001}
S.~Tong and D.~Koller.
\newblock Support vector machine active learning with applications to text
  classification.
\newblock {\em Journal of machine learning research}, 2(Nov):45--66, 2001.

\bibitem{wei2018github}
X.~Wei, B.~Gong, Z.~Liu, W.~Lu, and L.~Wang.
\newblock {CT-GAN}.
\newblock \url{https://github.com/biuyq/CT-GAN}, 2018.

\bibitem{wei2018}
X.~Wei, B.~Gong, Z.~Liu, W.~Lu, and L.~Wang.
\newblock Improving the improved training of wasserstein gans: A consistency
  term and its dual effect.
\newblock In {\em Proceedings of the International Conference on Learning
  Representations}, 2018.

\bibitem{yang2016}
Y.~Yang and M.~Loog.
\newblock Active learning using uncertainty information.
\newblock In {\em Pattern Recognition (ICPR), 2016 23rd International
  Conference on}, pages 2646--2651. IEEE, 2016.

\bibitem{yang2015}
Y.~Yang, Z.~Ma, F.~Nie, X.~Chang, and A.~G. Hauptmann.
\newblock Multi-class active learning by uncertainty sampling with diversity
  maximization.
\newblock {\em International Journal of Computer Vision}, 113(2):113--127,
  2015.

\bibitem{yoo2019learningloss}
D.~Yoo and I.~S. Kweon.
\newblock Learning loss for active learning.
\newblock In {\em The IEEE Conference on Computer Vision and Pattern
  Recognition (CVPR)}, June 2019.

\bibitem{yu2015lsun}
F.~Yu, A.~Seff, Y.~Zhang, S.~Song, T.~Funkhouser, and J.~Xiao.
\newblock Lsun: Construction of a large-scale image dataset using deep learning
  with humans in the loop.
\newblock {\em arXiv preprint arXiv:1506.03365}, 2015.

\bibitem{zhu2017}
J.-J. Zhu and J.~Bento.
\newblock Generative adversarial active learning.
\newblock In {\em Advances in Neural Information Processing Systems Workshops},
  2017.

\end{thebibliography}
}
\cleardoublepage
\appendix
\section{Model Architectures and Experimental Details}

Tabs.~\ref{tab:auto-mnist},~\ref{tab:auto-cifar} and~\ref{tab:auto-celeba} show the architectures of the auto-encoders used for the three different data sets.
Tabs.~\ref{tab:models-mnist},~\ref{tab:models-cifar},~\ref{tab:models-celeba} show the architectures of the classifiers, generator and discriminators. Tab.~\ref{tab:resnet-cifar} shows the structure of the residual generators and discriminators. Tab.~\ref{tab:gan-training} shows the number of training iterations used for each GAN and data set. Wherever possible we use the same training parameters as the author of the accompanying paper. Therefore we follow~\cite{gulrajani2017} and~\cite{wei2018} to train all the GANs. We only use a different amount of training iterations depending on the data set. Futhermore, we follow~\cite{springenberg2014} to train the classifier used for \textit{CIFAR-10 - ten classes}.

\begin{table}[h]
  \renewcommand{\arraystretch}{1.3}
  \newcommand{\head}[1]{\textnormal{\textbf{#1}}}

  \colorlet{tableheadcolor}{gray!25} 
  \newcommand{\headcol}{\rowcolor{tableheadcolor}} %
  \colorlet{tablerowcolor}{gray!10} 
  \newcommand{\rowcol}{\rowcolor{tablerowcolor}} %
  \setlength{\aboverulesep}{0pt}
  \setlength{\belowrulesep}{0pt}

  \newcommand*{\rulefiller}{
    \arrayrulecolor{tableheadcolor}
    \specialrule{\heavyrulewidth}{0pt}{-\heavyrulewidth}
    \arrayrulecolor{black}}

  \caption{Number of training iterations for the different GANs and data sets.}
  \label{tab:gan-training}
  \centering
  \resizebox{1\columnwidth}{!}{%
  \begin{tabular}{lrrrrrrr}
    \toprule
    \headcol \head{GAN} & \multicolumn{2}{c}{\head{MNIST}} & \multicolumn{2}{c}{\head{CIFAR-10}} & \head{CelebA}  & \head{SVHN} & \head{LSUN}\\
    \rulefiller \cmidrule(lr){2-3} \cmidrule(lr){4-5}
    \headcol & two & ten & two & ten & & &\\
    \midrule
    DCGAN & 40k  & 100k & 100k\footnotemark & 200k & --- & --- & ---\\
    WGAN-GP & 40k & 100k & 200k & 200k & 100k & 100k & 100k \\
    Res-WGAN-GP & --- & --- & 100k & 100k & --- & --- & ---\\
    Res-WGAN-CT & --- & --- & 100k & 100k & --- & --- & ---\\
    
    \bottomrule
  \end{tabular}
  }
\end{table}
\footnotetext{We observe mode collapse for \emph{CIFAR-10 - two classes} using DCGAN when training the GAN for more than 100k iterations.}

\subsection{MNIST - two classes}
For binary digit classification we train a linear model with cross entropy loss.
We train the model for 10 epochs using the Adam optimizer~\cite{kingma2014} with a batch size of 10 and learning rate of 0.001. 
We train the Wasserstein GAN~\cite{gulrajani2017} with gradient penalty to synthesize only the digits 5 \& 7. Similarly, we train the auto-encoder for 40 epochs with a batch size of 100 using the Adam optimizer with a learning rate of 0.001 using only the digits 5 \& 7. For the binary experiments we decrease the number of channels in the auto encoder (see Tab.~\ref{tab:auto-mnist}) by a factor of three.

\subsection{CIFAR-10 - two classes}
For \textit{CIFAR-10 - two classes} we train a linear model with cross entropy loss.
We train the model for 10 epochs using the Adam optimizer~\cite{kingma2014} with a batch size of 50 and learning rate of 0.001. 
We train the Wasserstein GAN~\cite{gulrajani2017} with gradient penalty to synthesize only the two target classes. Similarly, we train the auto-encoder for 100 epochs with a batch size of 128 using the Adam optimizer with a learning rate of 0.001 using only the images with the two target classes.

\subsection{MNIST - ten classes}
For ten digit classification we use LeNet~(\cite{lecun1998}) with cross entropy. We train the model for 10 epochs using the Adam optimizer with a batch size of 50 and learning rate of 0.001.
We train the Wasserstein GAN~\cite{gulrajani2017} with gradient penalty and the auto-encoder for 40 epochs with a batch size of 100 using the Adam optimizer with a learning rate of 0.001.

\subsection{CIFAR-10 - ten classes}
Using all classes of CIFAR-10 complicates the classification task and we require a deep model to achieve close to state-of-the-art results. Therefore, we use the All-CNN model proposed by \cite{springenberg2014} with a reported classification error of 9.08\%. We use the proposed architectures and training strategies and use stochastic gradient descent with constant momentum of 0.9 and a learning rate of 0.01 that we decay by a factor of 10 at the 130th and the 140th epoch.
We train the model for 150 epochs with a batch size of 128 without data augmentation and report a classification error of 11.8\%. 
The All-CNN contains  $\sim$1.4 million different parameters. Hence, we require larger initial training sets than for the previous models. Thus we include 100 randomly selected images per class. We add 1000 samples to the data set every AL cycle until the budget of 30k samples is reached. We generate ten times a batch containing 100 samples because optimizing for all samples at the same time is unfeasible.
In addition to the previous experiments we test a traditional layered and a residual structure for the GAN. We train both with gradient penalty with or without a soft consistency term.

\subsection{CelebA}
For classification we use the CNN presented in Tab.~\ref{tab:models-celeba}. We use the Adam optimizer with a learning rate of 0.001 and a batch size of 50 and train for 30 epochs. We start active learning with 100 labelled samples, where the number of samples per class corresponds to the data distribution. We train the auto-encoder for 100 epochs with a batch size of 64 using the Adam optimizer with a learning rate of 0.0001.

\subsection{SVHN}
For classification we use the Conv-Small CNN proposed by Miyato\etal~\cite{miyato2018vat} and use the auto encoder and GAN architectures presented in Tabs.~\ref{tab:auto-cifar} and~\ref{tab:models-cifar}. We use the Adam optimizer with a learning rate of 0.001 and a batch size of 128 and train for 120 epochs. We start active learning with 1000 labelled samples, where the number of samples per class corresponds to the data distribution. We train the auto-encoder for 100k iterations with a batch size of 64 using the Adam optimizer with a learning rate of 0.0001.

\subsection{LSUN}
For classification we use the Conv-Small CNN proposed by Miyato\etal~\cite{miyato2018vat} because the architecture is designed for $32\times32$ color images we change the first layer to use $5\times5$ convolution kernels with stride two.
The auto encoder and GAN architectures are presented in Tabs.~\ref{tab:auto-celeba} and~\ref{tab:models-celeba}.
We use the Adam optimizer with a learning rate of 0.001 and a batch size of 128 and train for 120 epochs. We start active learning with 1000 labelled samples, where the number of samples per class corresponds to the data distribution. We train the auto-encoder for 100k iterations with a batch size of 64 using the Adam optimizer with a learning rate of 0.0001.

\section{Additional Results: MNIST - two classes}
\begin{figure*}[t]
\centering
\includegraphics[width=0.9\textwidth, keepaspectratio]{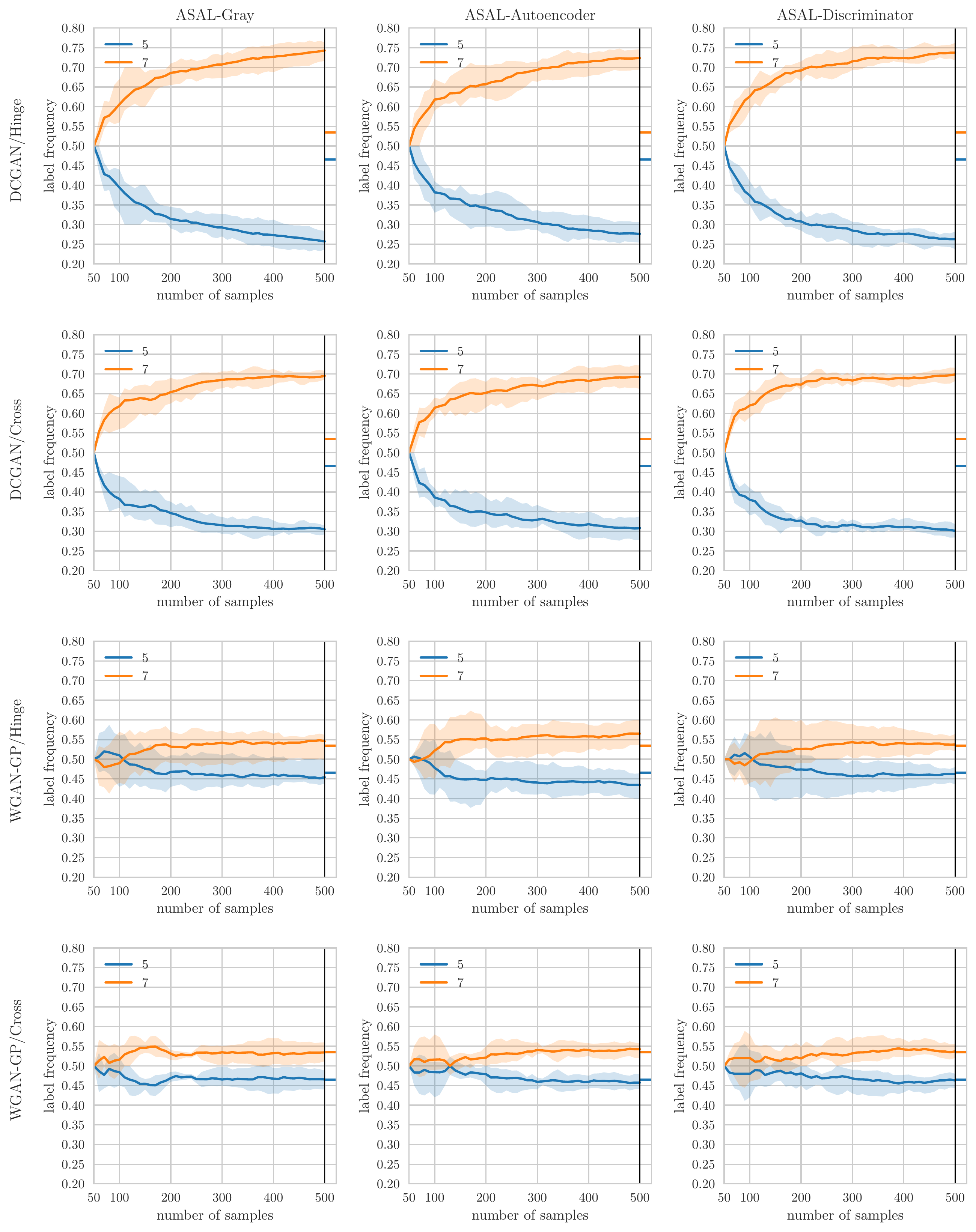} 
\caption{Label distribution for active learning using different matching strategies, uncertainty measures and \acrshortpl{gan} for \emph{MNIST - two classes}. The ticks on the right show the true label distribution in the pool.
\gls{asal} using WGAN-GP (third and fourth row) reaches a label distribution of the training data that is similar to the true label distribution in the pool. Conversely, \gls{asal} using DCGAN (first and second row) leads to a training set that contains almost three times as many images with the digit \textsf{7} than digit \textsf{5}. Most likely, the DCGAN is responsible for this behaviour because we already observed that it produces the digit \textsf{7} more frequently than the digit \textsf{5}.}\label{app:fig:mnist-binary-asal-label-dist}
\end{figure*}

\begin{figure*}[t]
\centering 
\subfloat[Random sampling with Hinge loss.]{\includegraphics[width=0.25\textwidth, keepaspectratio]{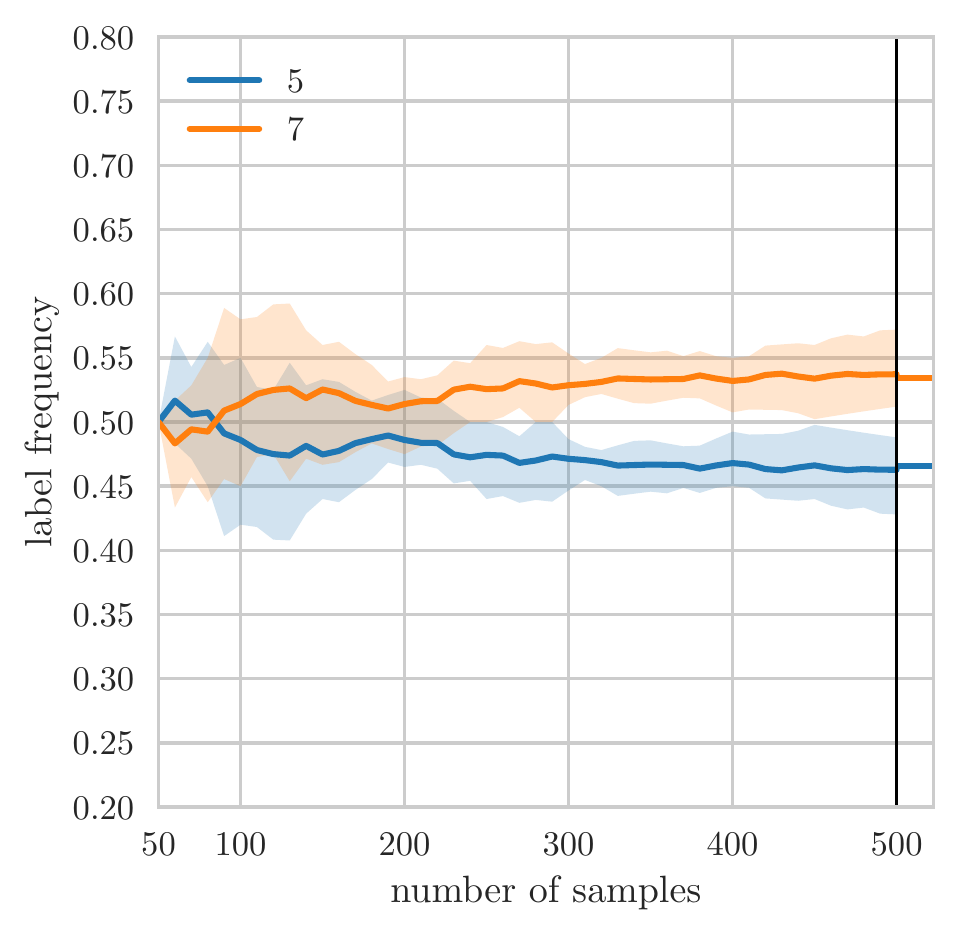}}
\subfloat[Random sampling with cross-entropy loss.]{\includegraphics[width=0.25\textwidth, keepaspectratio]{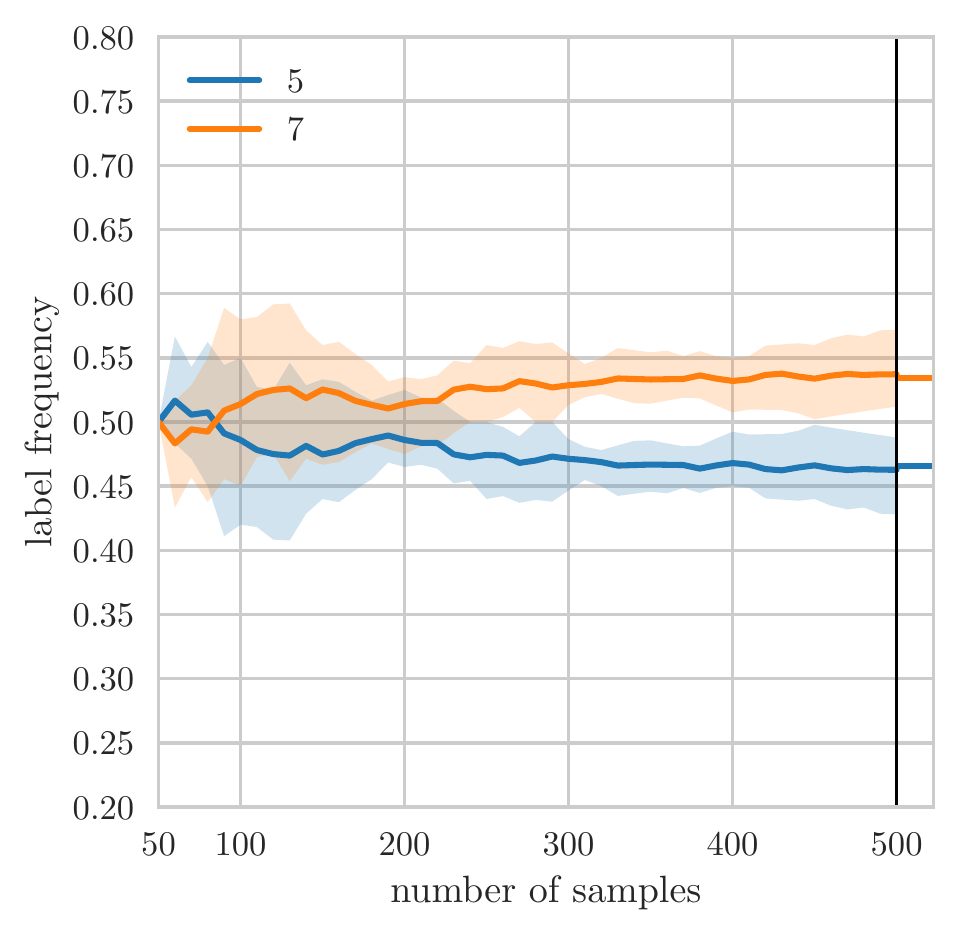}}
\subfloat[\emph{Minimum distance} sampling with Hinge loss.]{\includegraphics[width=0.25\textwidth, keepaspectratio]{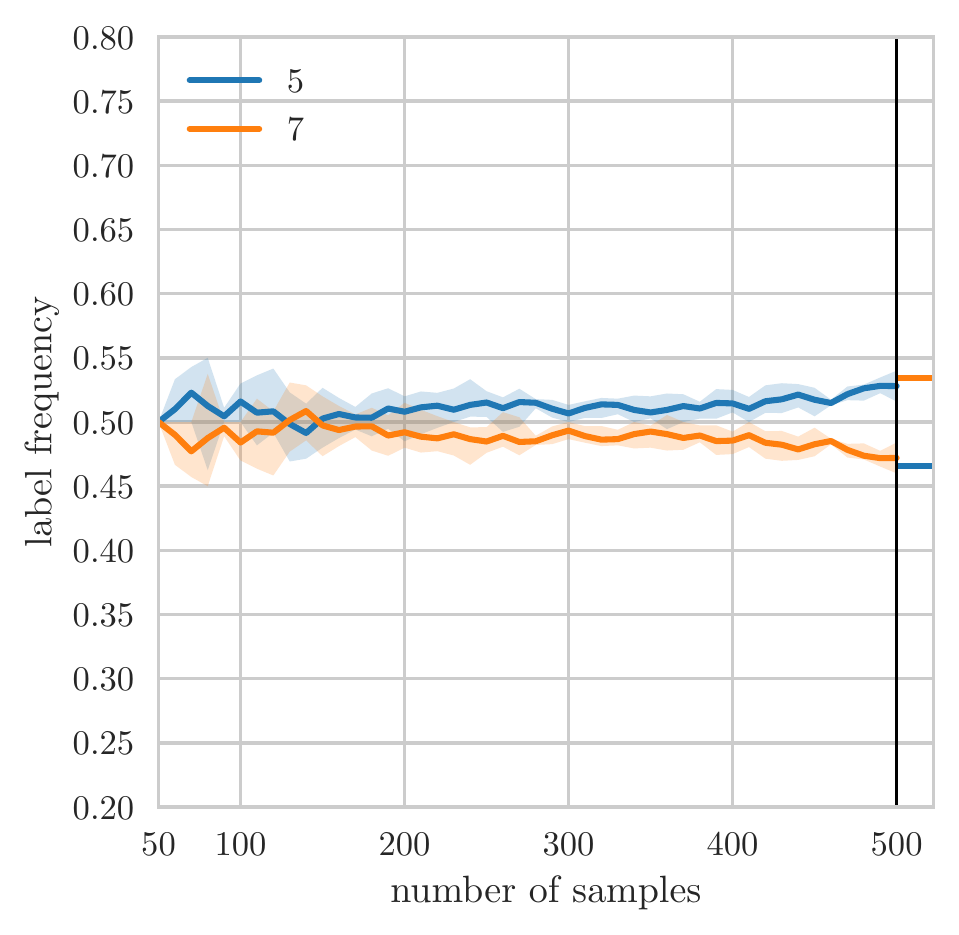}}
\subfloat[\emph{Maximum entropy} sampling with cross-entropy loss.]{\includegraphics[width=0.25\textwidth, keepaspectratio]{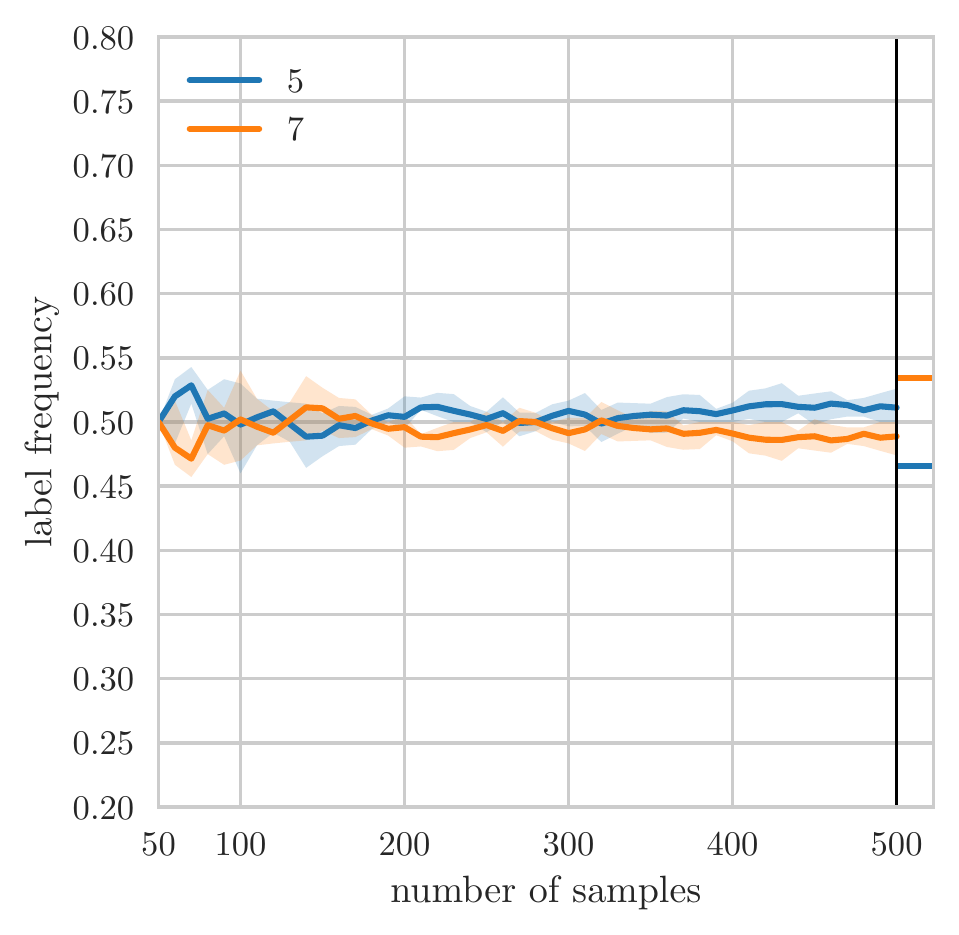}}

\caption{Label distribution for uncertainty sampling using maximum entropy and random sampling for \emph{MNIST - two classes} using different uncertainty measures and loss functions. The tick on the right show the true label distribution in the pool. The label distribution of the training set, assembled with random sampling (top), converges to the true label distribution of the pool. Conversely, uncertainty sampling leads to a training set that contains more frequently the label \textsf{5} than \textsf{7} compared to the pool that contains \textsf{7} more frequently. Apparently, images with the digit \textsf{5} lead to higher uncertainty of the used classifier.}\label{app:fig:mnist-binary-label-dist}
\end{figure*}

\begin{figure*}[t]
\centering
\subfloat[\emph{Minimum distance} with Hinge loss and DCGAN.]{\includegraphics[width=0.25\textwidth, keepaspectratio]{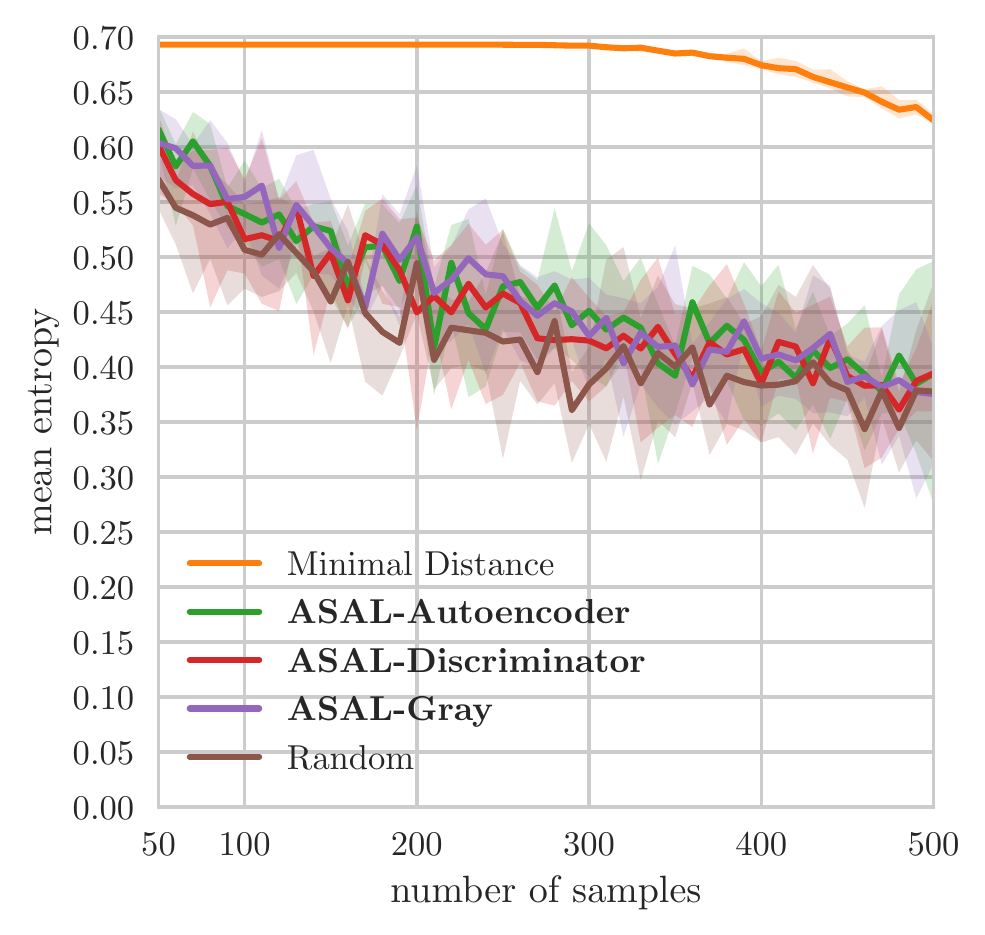}}
\subfloat[\emph{Maximum entropy} with cross-entropy loss and DCGAN.]{\includegraphics[width=0.25\textwidth, keepaspectratio]{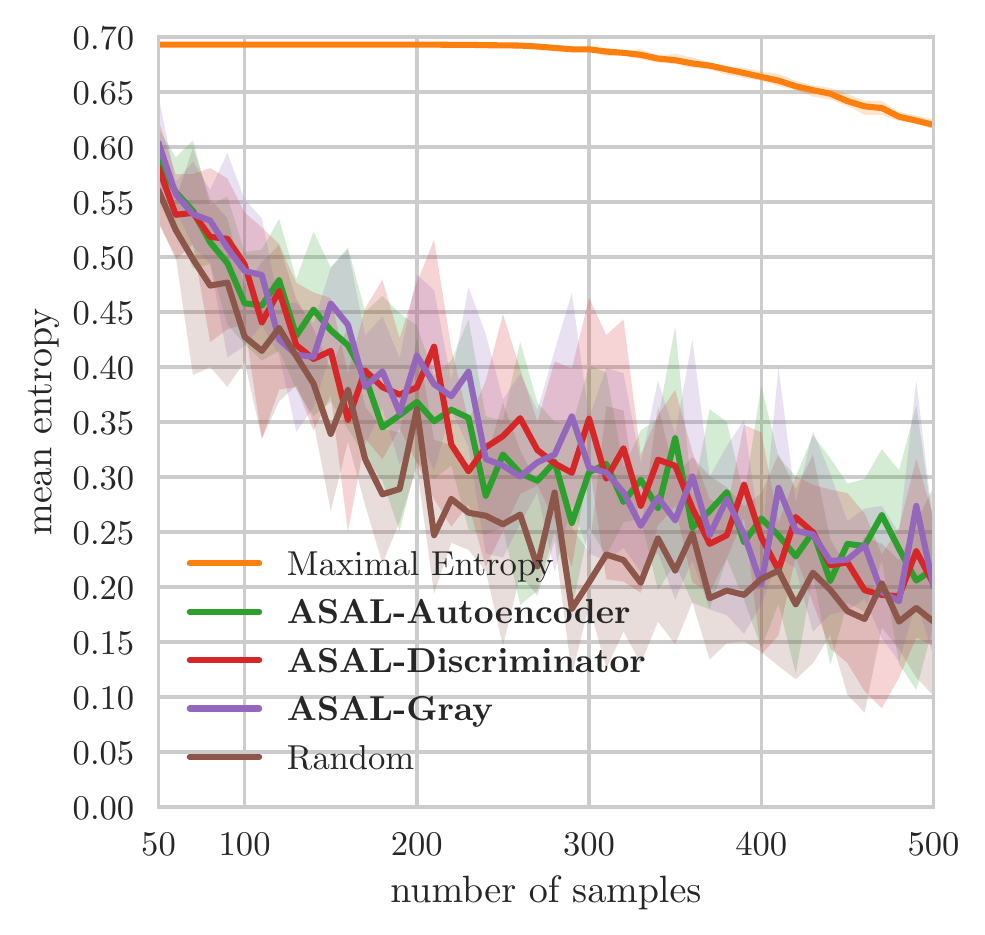}}
\subfloat[\emph{Minimum distance} with Hinge loss and WGAN-GP.]{\includegraphics[width=0.25\textwidth, keepaspectratio]{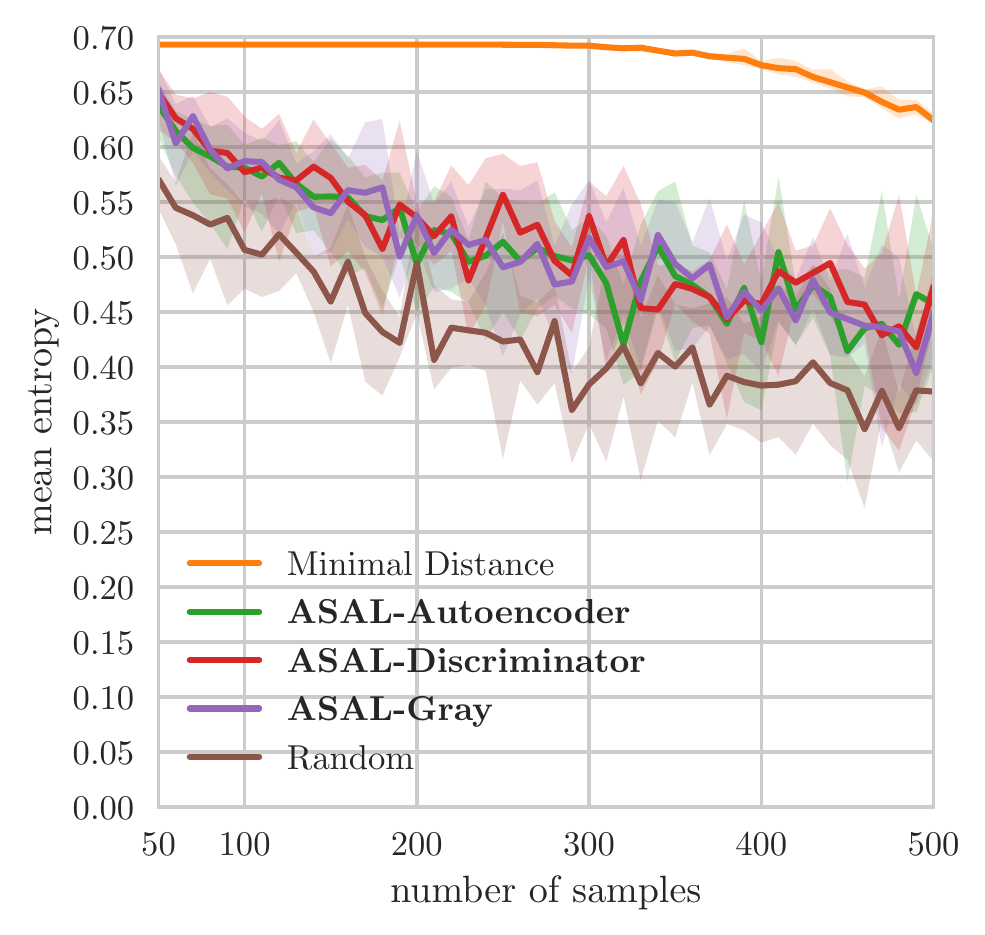}}
\subfloat[\emph{Maximum entropy} with cross-entropy loss and WGAN-GP.]{\includegraphics[width=0.25\textwidth, keepaspectratio]{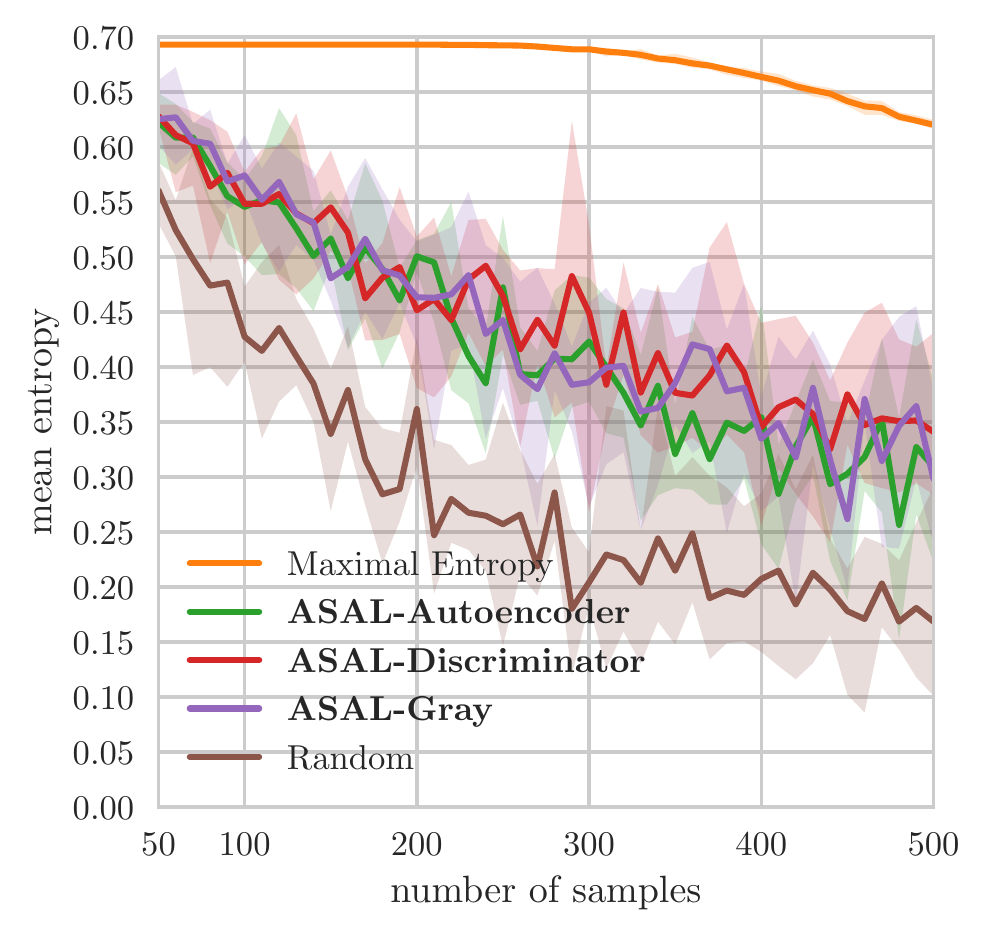}\label{app:fig:mnist-binary-new-entropies-wgan-max-ent}}

\caption{Average entropy of images that are selected and added to the training set for \emph{MNIST - two classes} using different \acrshortpl{gan}, uncertainty measures and loss functions. All figures show that \acrshort{asal} selects samples from the pool that have a higher entropy than randomly sampled images. However, maximum entropy sampling and WGAN-GP (\ref{app:fig:mnist-binary-new-entropies-wgan-max-ent}) lead to the largest entropy gap between selected and randomly sampled images. Maximum entropy sampling (right column) results in smaller average entropy of the classifier than minimum distance sampling (left column) because we use the cross-entropy loss that directly optimizes for small entropy, opposed to the hinge loss that minimizes the distance to the separating hyper-plane.}\label{app:fig:mnist-binary-entropy}
\end{figure*}

Fig.~\ref{app:fig:mnist-binary-test} shows the test accuracy of \gls{asal} for different uncertainty scores and \glspl{gan}. Fig.~\ref{app:fig:mnist-binary-asal-label-dist} and~\ref{app:fig:mnist-binary-label-dist} shows the label distribution during active learning for \gls{asal} an classical methods.
Fig.~\ref{app:fig:mnist-binary-entropy} shows the entropy of newly added samples.

\subsection{Agreement of Manual Annotations and Matched Labels}
\begin{figure}[t]
\centering
\includegraphics[width=\columnwidth, keepaspectratio]{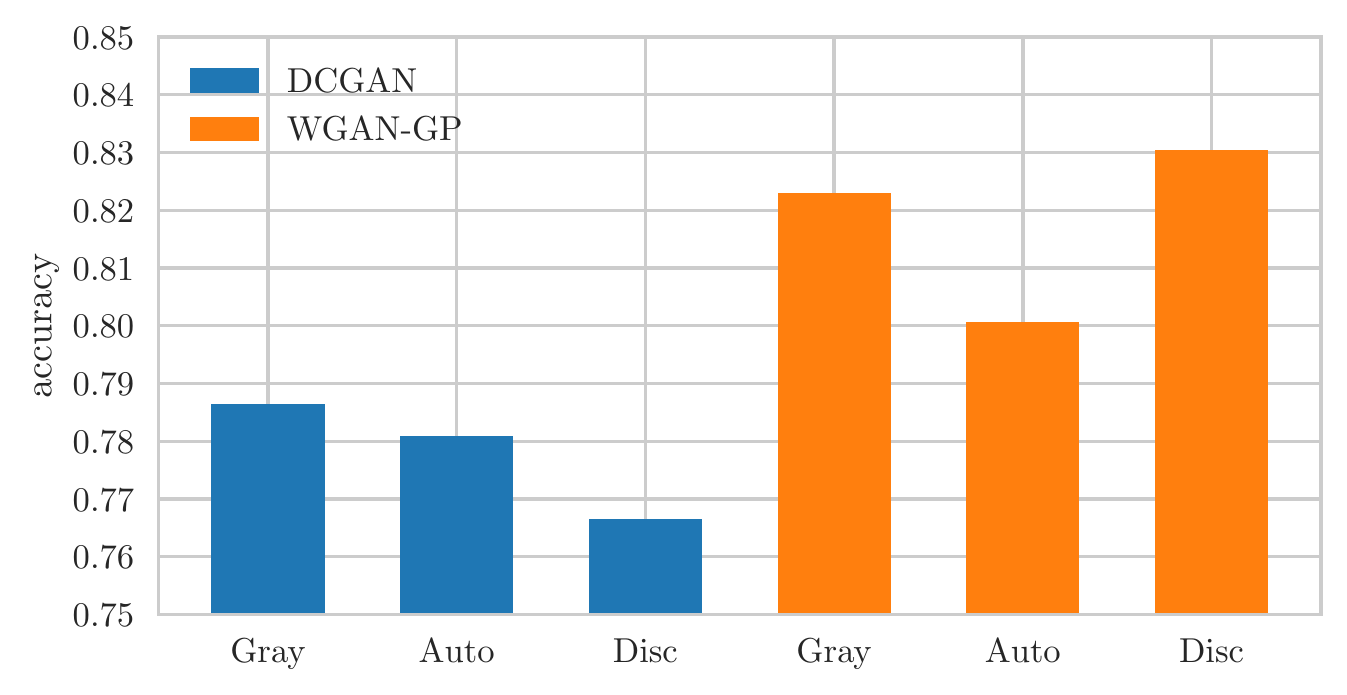}
\caption{Comparison of the agreement accuracy between manual annotations and matched labels. The matching strategies employed in \acrshort{asal} allow to select similar images from the pool and compare these labels to the manual annotations. For \emph{MNIST - two classes} the agreement for WGAN-GP is higher than for DCGAN.}
\label{app:fig:mnist-binary-gaal-label-matching}
\end{figure}

Instead of manually annotating images we propose to select similar images from the pool and ask for labels of these images. Similar images might show an object of the same class, have similar surrounding, colors, size or share other features. Thus, we compare the agreement of the manual class annotations of the generated images with the matched images, using the three different strategies. We use 1300 generated samples for each GAN, annotate the images manually and retrieve the closest match with the corresponding label from the pool. 
We assume that the final model will be measured on an almost evenly distributed test set similar to MNIST and USPS. However, the test set for this experiment contains the generated samples with manual annotations and the \gls{gan} may generate samples with unevenly distributed label frequency. Thus, we compute the accuracy for each class independently and average these values subsequently to obtain the final score.

\begin{figure*}[t]
\centering
\subfloat[\emph{Minimum distance} with Hinge loss and DCGAN.]{\includegraphics[width=0.25\textwidth, keepaspectratio]{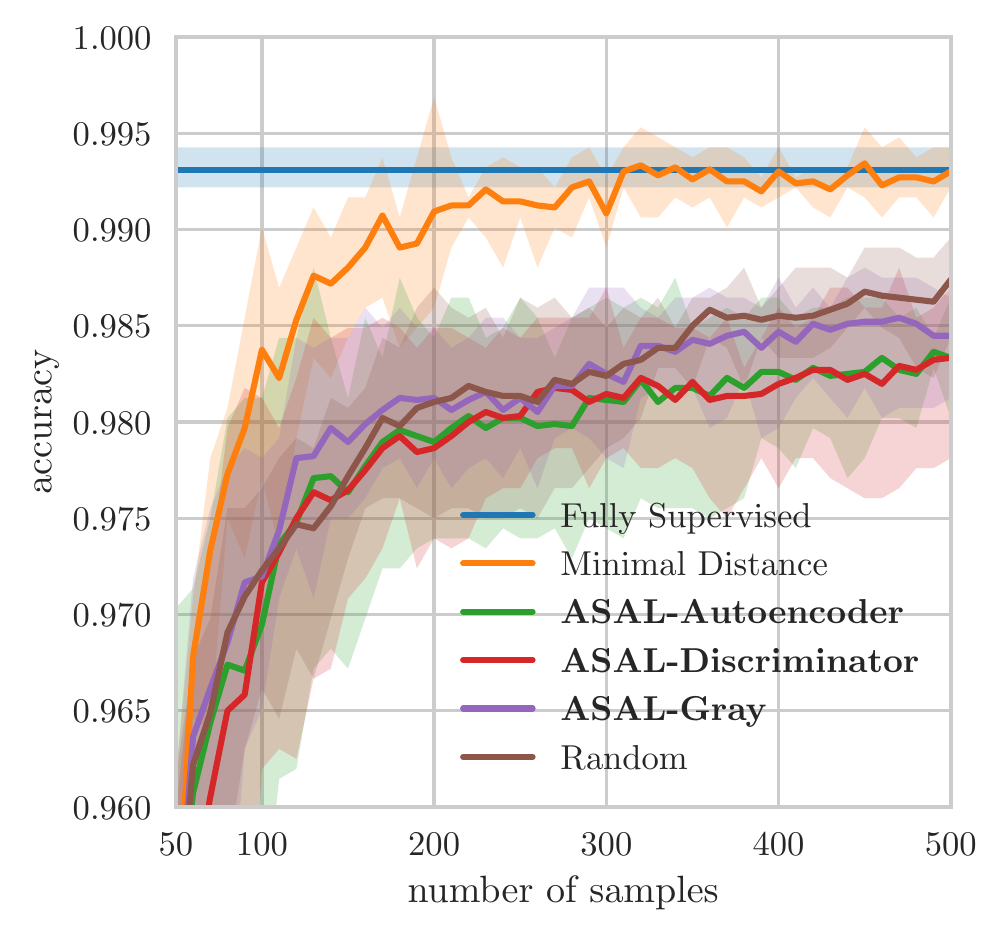}\label{app:fig:mnist-binary-test-acc-dcgan-hinge}}
\subfloat[\emph{Maximum entropy} with cross-entropy loss and DCGAN.]{\includegraphics[width=0.25\textwidth, keepaspectratio]{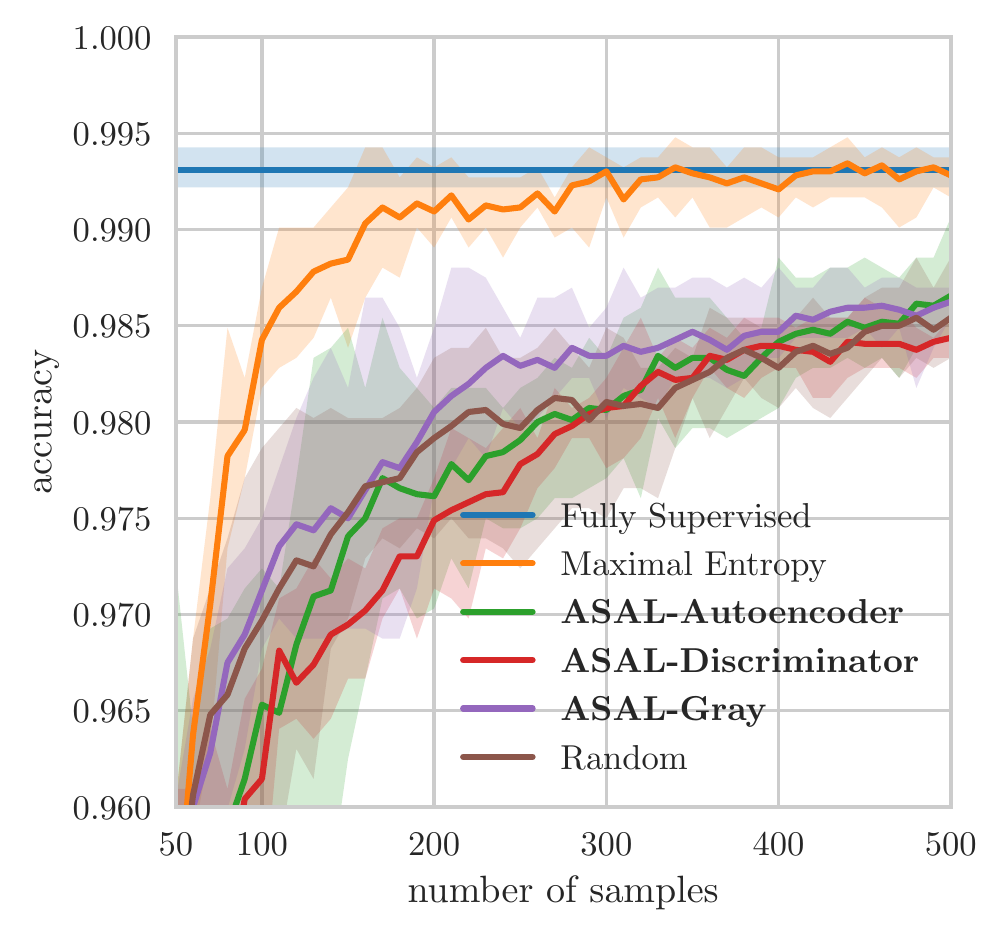}\label{app:fig:mnist-binary-test-acc-dcgan-maxent}}
\subfloat[\emph{Minimum distance} with Hinge loss and WGAN-GP.]{\includegraphics[width=0.25\textwidth, keepaspectratio]{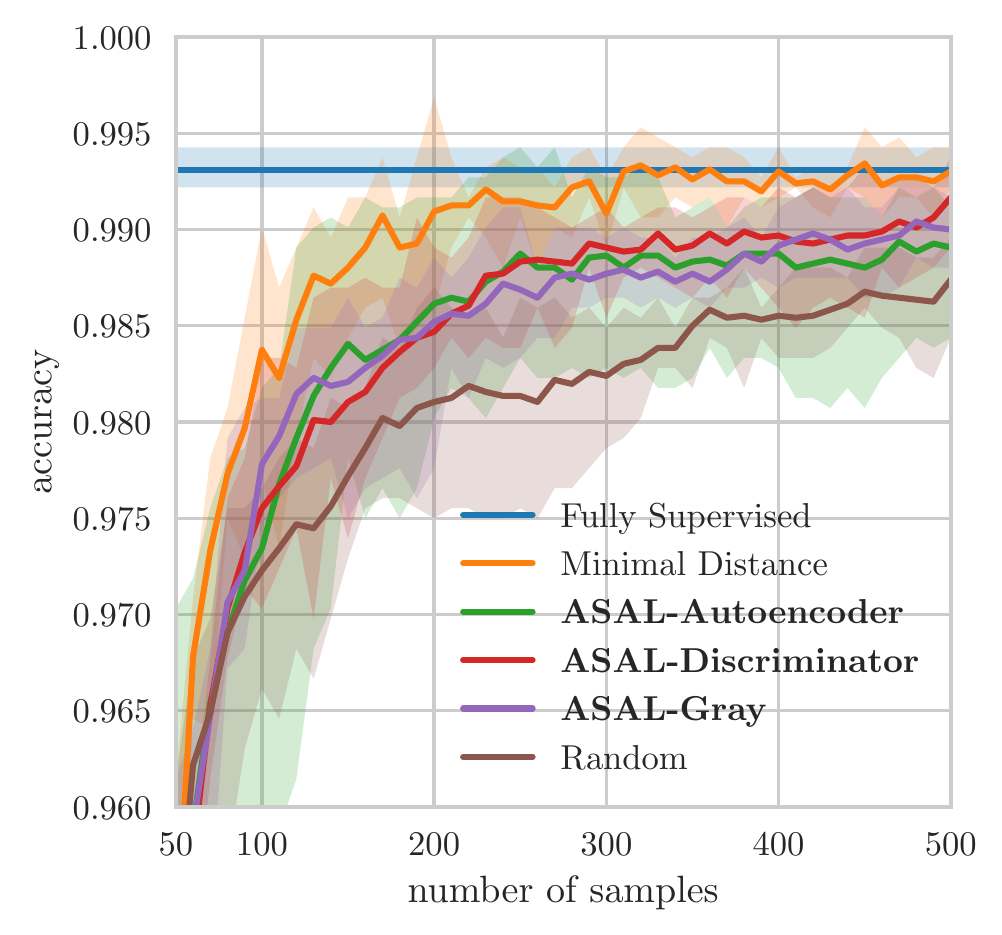}}
\subfloat[\emph{Maximum entropy} with cross-entropy loss and WGAN-GP.]{\includegraphics[width=0.25\textwidth, keepaspectratio]{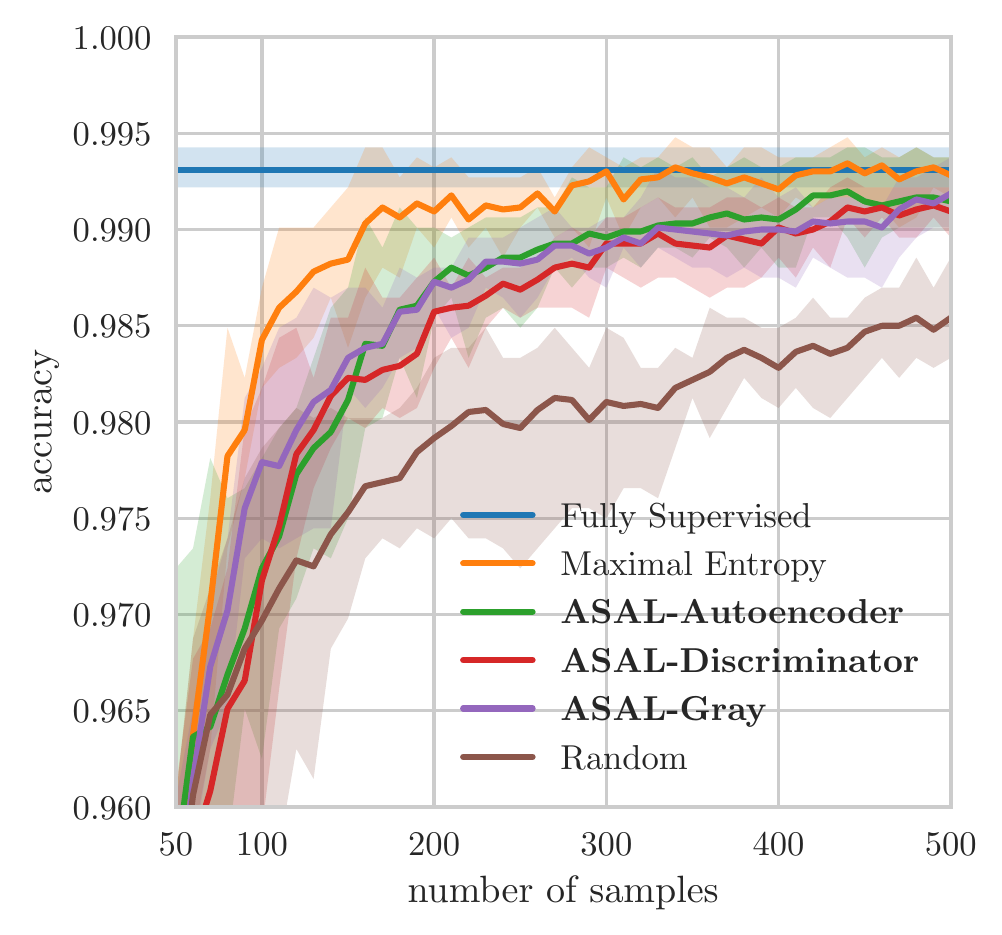}\label{app:fig:mnist-binary-test-acc-max-ent-cross}}

\caption{Test accuracy on \emph{MNIST - two classes} of a fully supervised model, for random sampling, uncertainty sampling and different ASAL using different GANs, uncertainty measures and loss functions. ASAL with WGAN-GP (bottom) clearly exceed the performance of ASAL using DCGAN (top). Maximum entropy sampling and using the cross entropy loss lead to the setup (\ref{app:fig:mnist-binary-test-acc-max-ent-cross}) that approaches the fully-supervised model with the fewest samples and reaches the smallest gap for all ASAL using 500 labelled samples.}
\label{app:fig:mnist-binary-test}
\end{figure*}

Fig.~\ref{app:fig:mnist-binary-gaal-label-matching} shows that the agreement is higher for \gls{asal} strategies using WGAN-GP than DCGAN. Furthermore, we observe that the matching based on gray values achieves the highest agreement. Similarly, Figs.~\ref{app:fig:mnist-binary-test-acc-dcgan-hinge} and~\ref{app:fig:mnist-binary-test-acc-dcgan-maxent} show best performance for ASAL-Gray.

\section{Additional Results: MNIST - ten classes}
\begin{figure*}[t]
\centering
\subfloat[DCGAN.]{\includegraphics[width=0.25\textwidth, keepaspectratio]{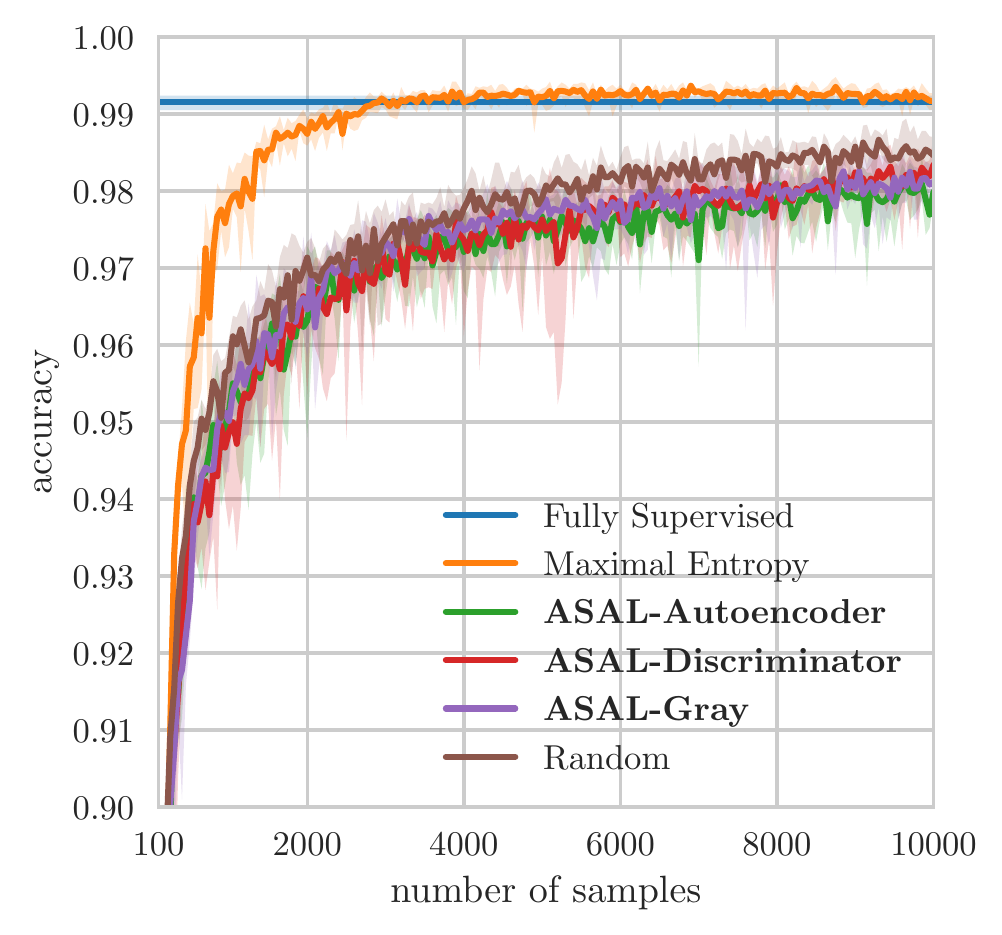}\label{app:fig:mnist-binary-test-acc-dcgan}}
\subfloat[WGAN-GP.]{\includegraphics[width=0.25\textwidth, keepaspectratio]{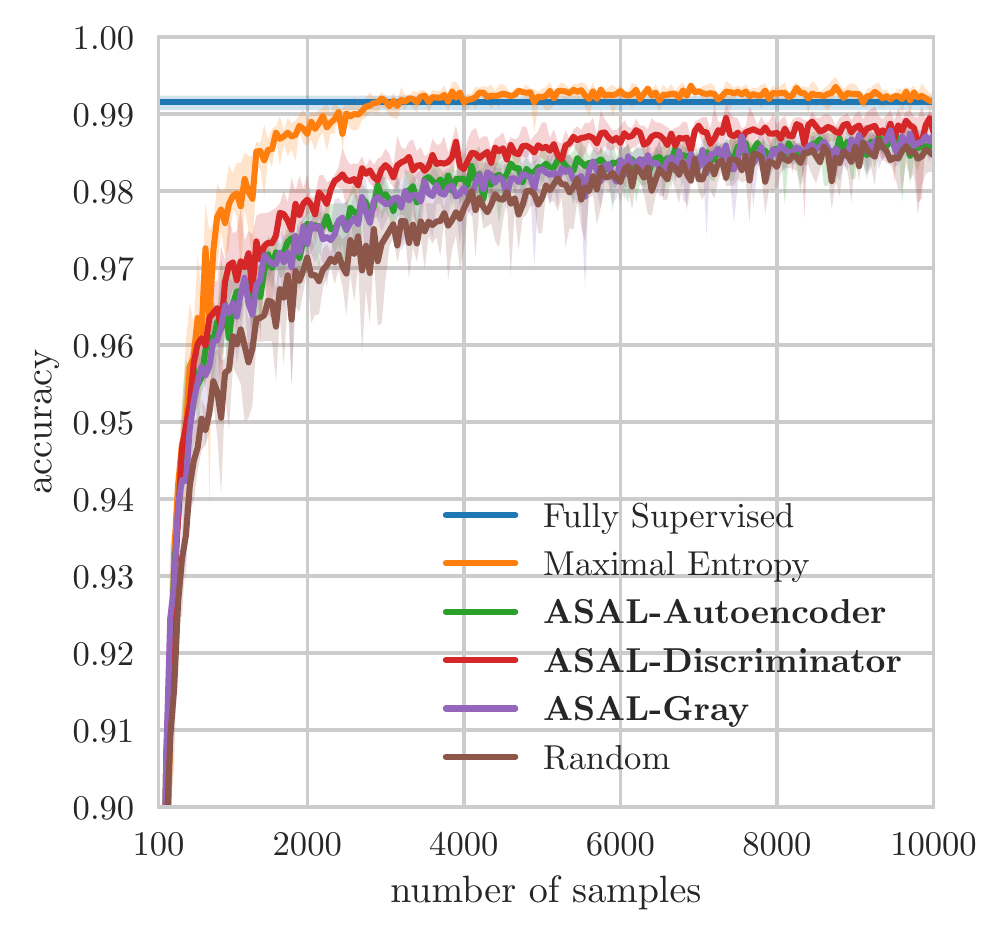}}

\caption{Test accuracy on \emph{MNIST - ten classes} of a fully supervised model, for random sampling, uncertainty sampling and different \acrshortpl{asal} using two different \acrshortpl{gan}. Selecting new images using random samples exceeds the performance of the proposed strategy when using the DCGAN. However, replacing the DCGAN with the WGAN-GP enables outperforming random sampling. \acrshort{asal}-Discriminator achieves the best quality.}\label{app:fig:mnist-all-test-acc}
\end{figure*}

\begin{figure*}[t]
\centering  
\subfloat[DCGAN.]{\includegraphics[width=0.25\textwidth, keepaspectratio]{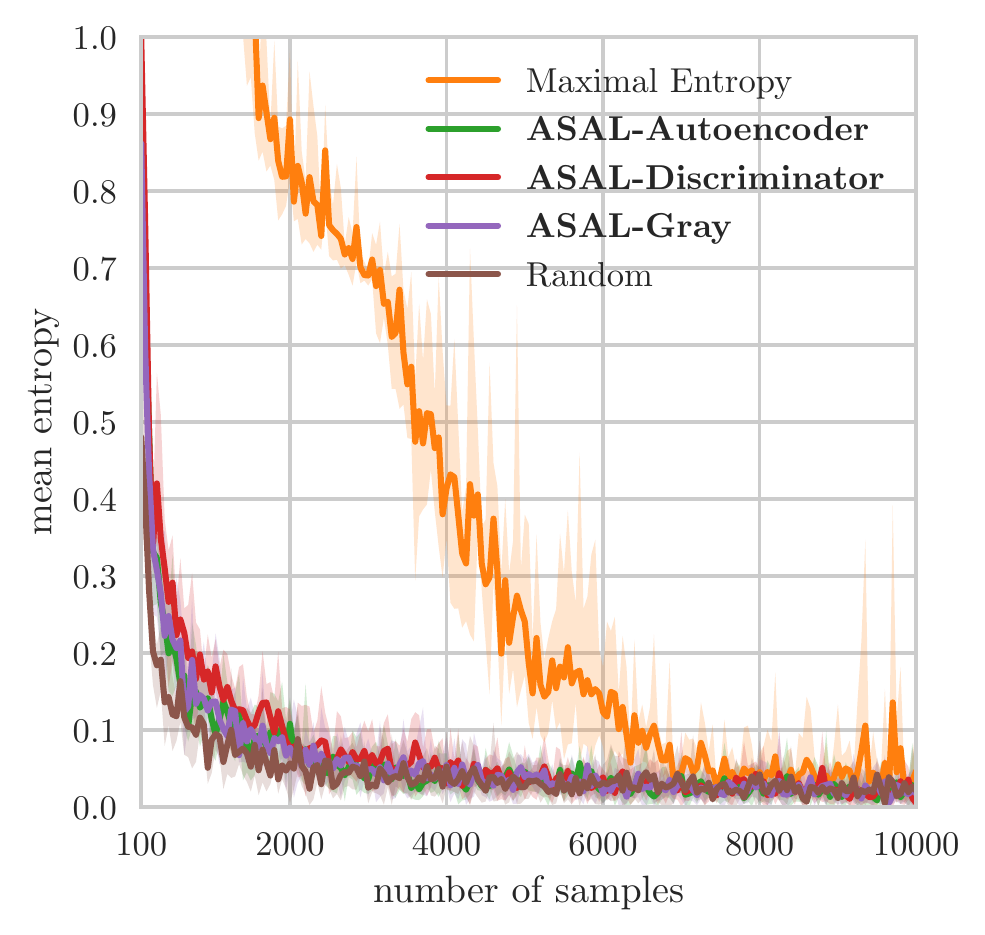}}
\subfloat[WGAN-GP.]{\includegraphics[width=0.25\textwidth, keepaspectratio]{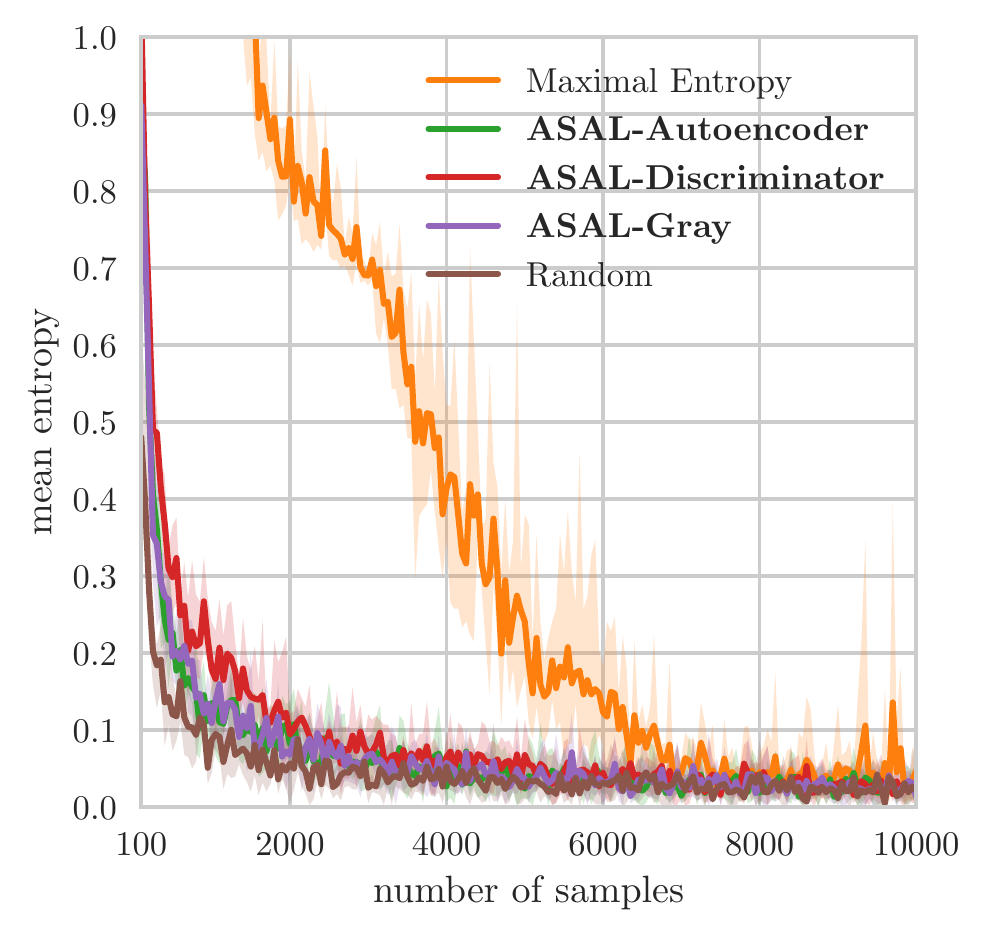}}

\caption{Average entropy of images that are selected and added to the training set for \emph{MNIST - ten classes} using different \acrshortpl{gan}. Both figures show that at the beginning \acrshort{asal} selects images with higher entropy than random sampling. In average WGAN-GP leads to a larger gap than DCGAN. However, this gap rapidly shrinks when increasing the training set.}\label{app:fig:mnist-all-entropy}
\end{figure*}

\begin{figure*}[t]
\centering
\subfloat[Random sampling.]{\includegraphics[width=0.33\textwidth, keepaspectratio]{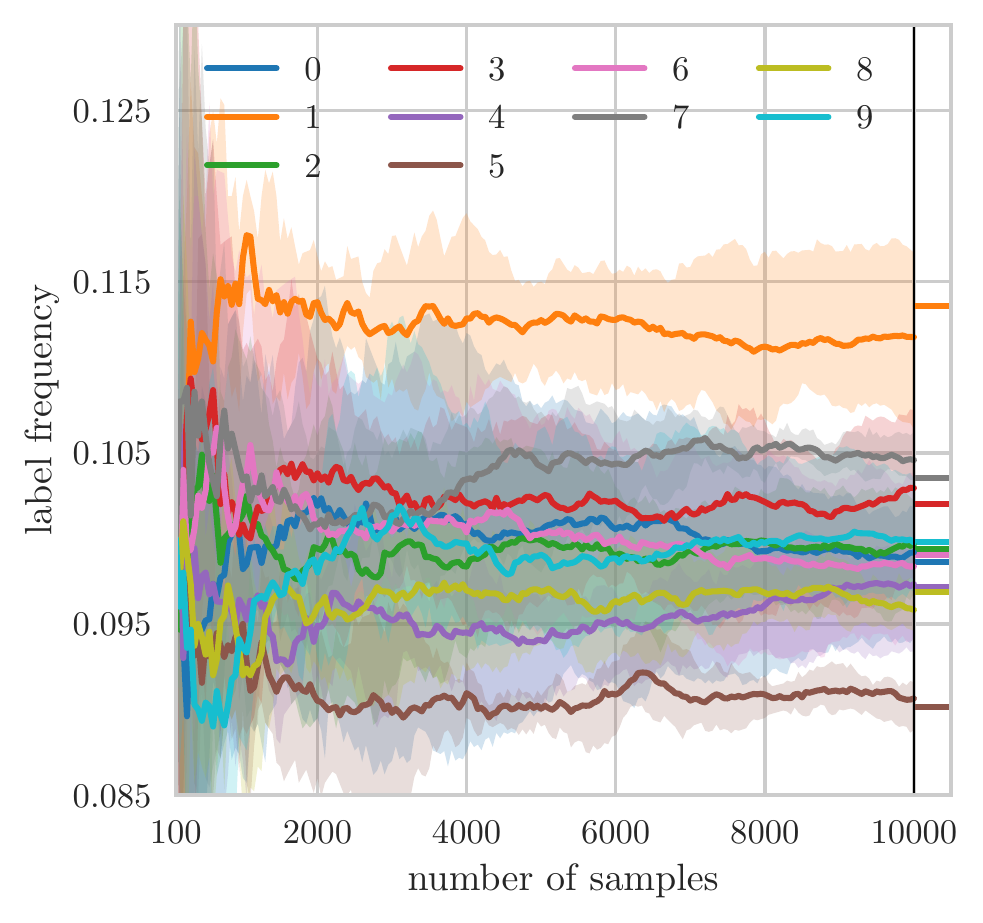}}
\subfloat[Maximum entropy sampling.]{\includegraphics[width=0.33\textwidth, keepaspectratio]{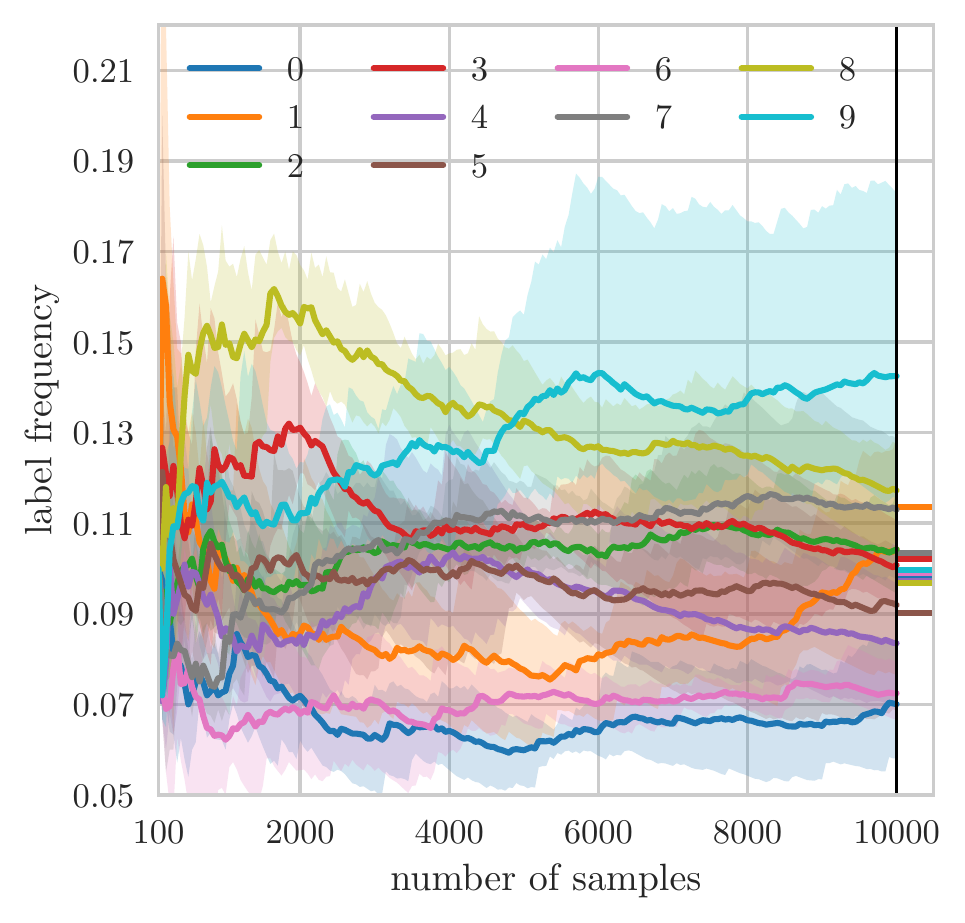}}
\\
\subfloat[ASAL-Gray with DCGAN.]{\includegraphics[width=0.33\textwidth, keepaspectratio]{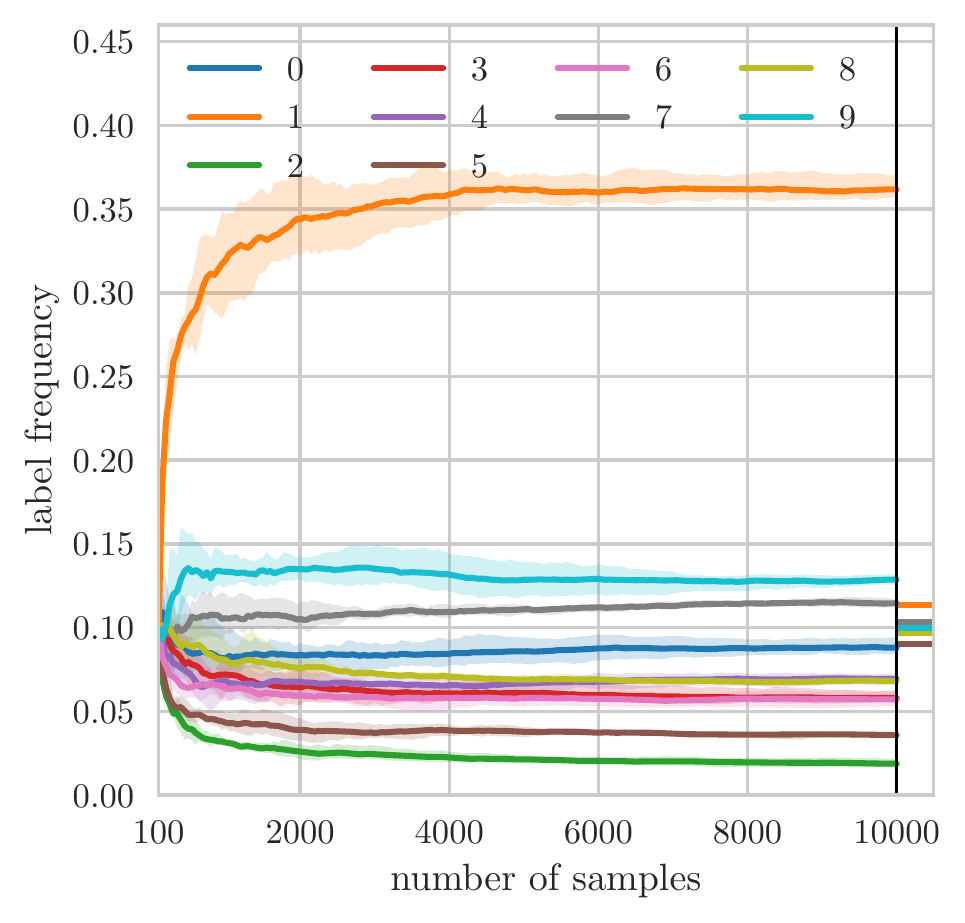}}
\hfill
\subfloat[ASAL-Autoencoder with DCGAN.]{\includegraphics[width=0.33\textwidth, keepaspectratio]{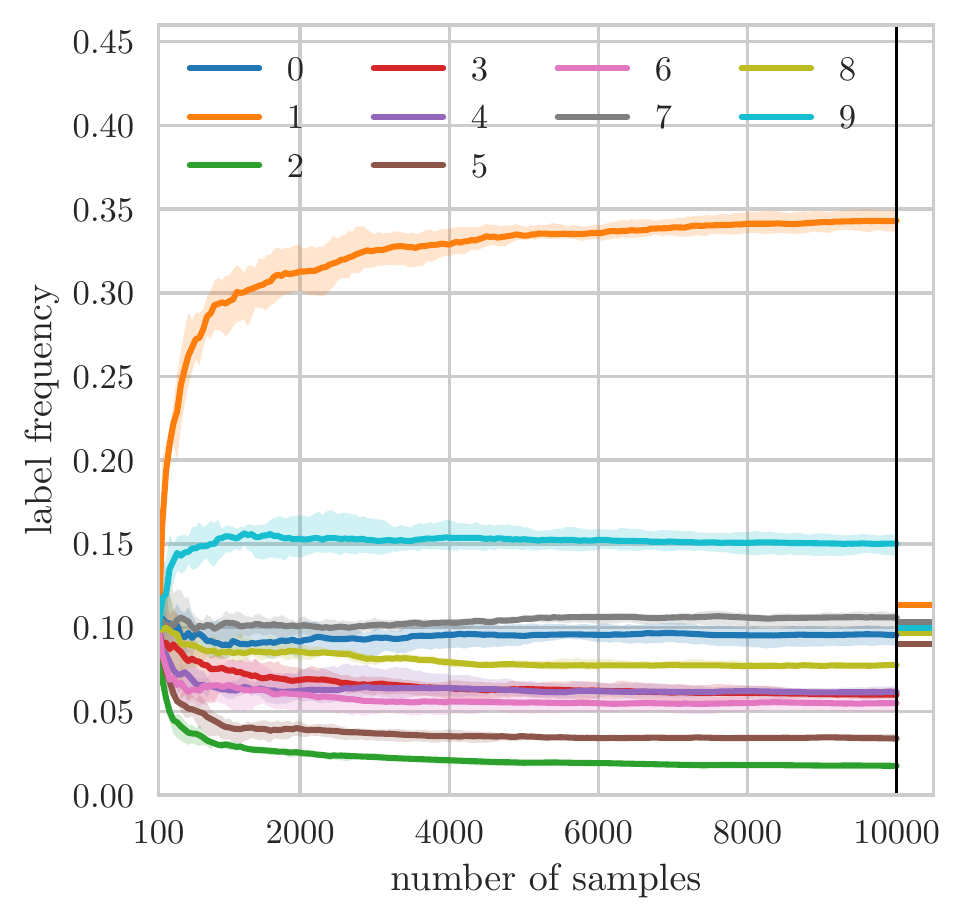}}
\hfill
\subfloat[ASAL-Discriminator with DCGAN.]{\includegraphics[width=0.33\textwidth, keepaspectratio]{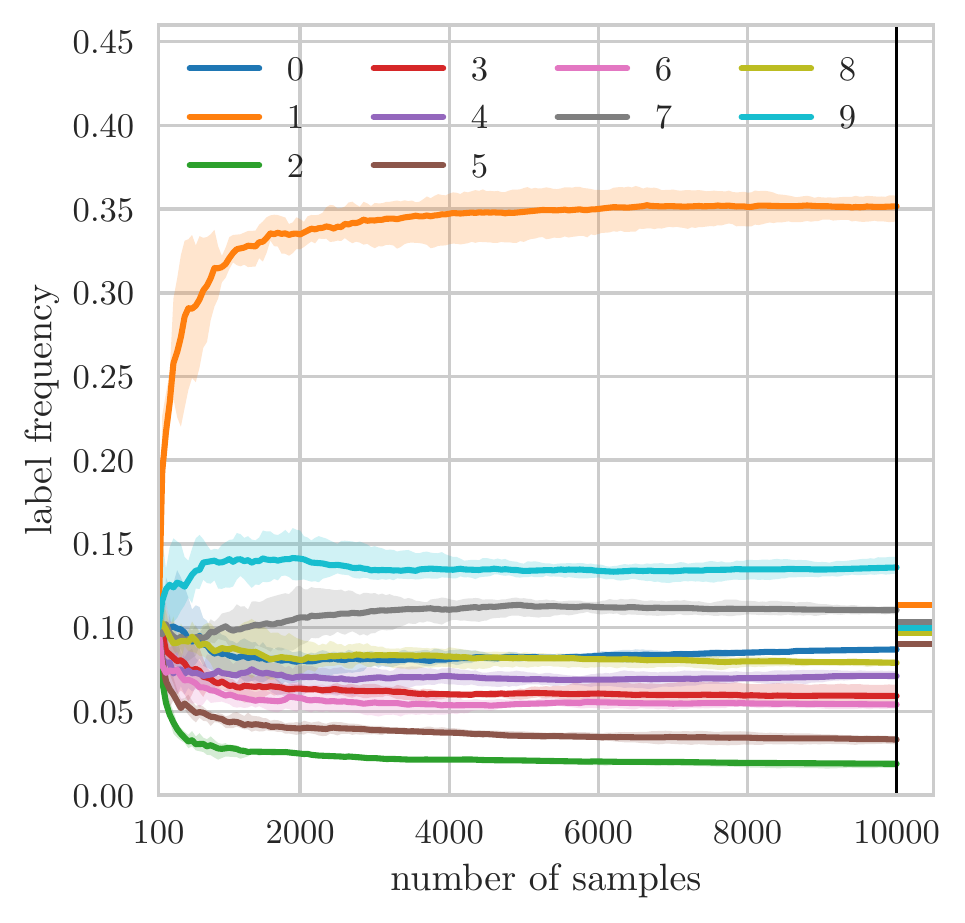}}
\\
\subfloat[ASAL-Gray with WGAN-GP.]{\includegraphics[width=0.33\textwidth, keepaspectratio]{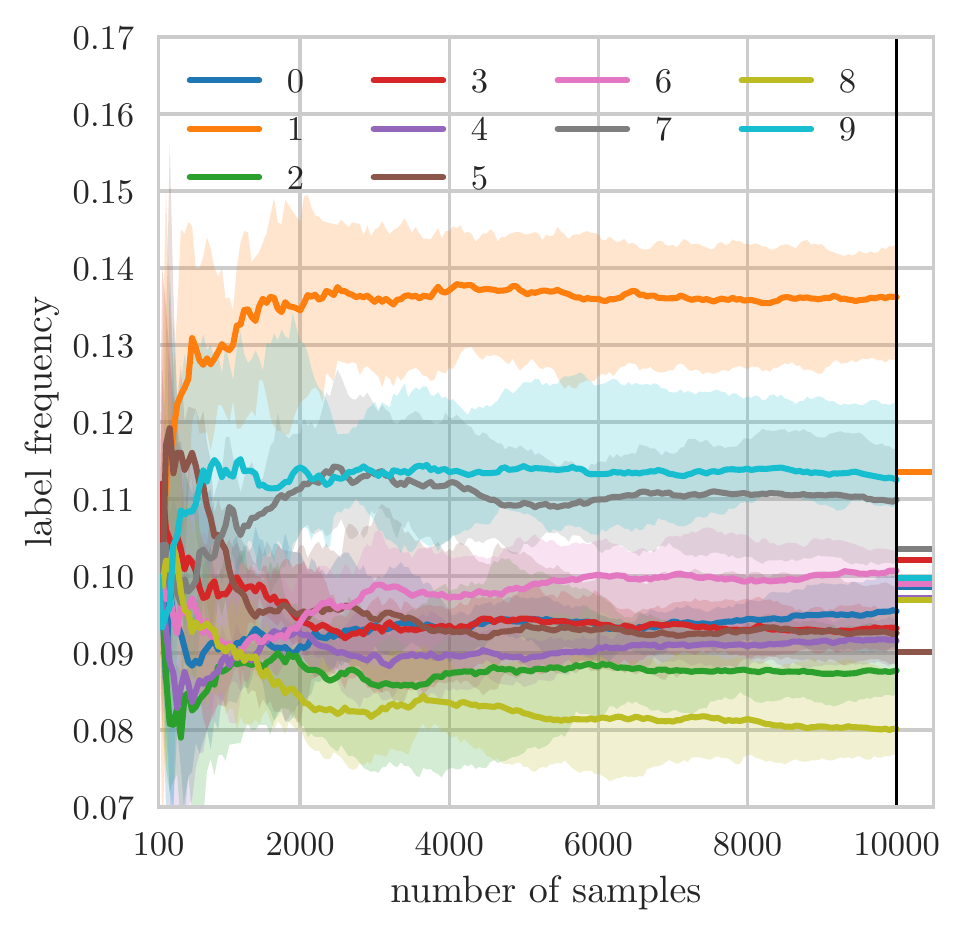}}
\hfill
\subfloat[ASAL-Autoencoder with WGAN-GP.]{\includegraphics[width=0.33\textwidth, keepaspectratio]{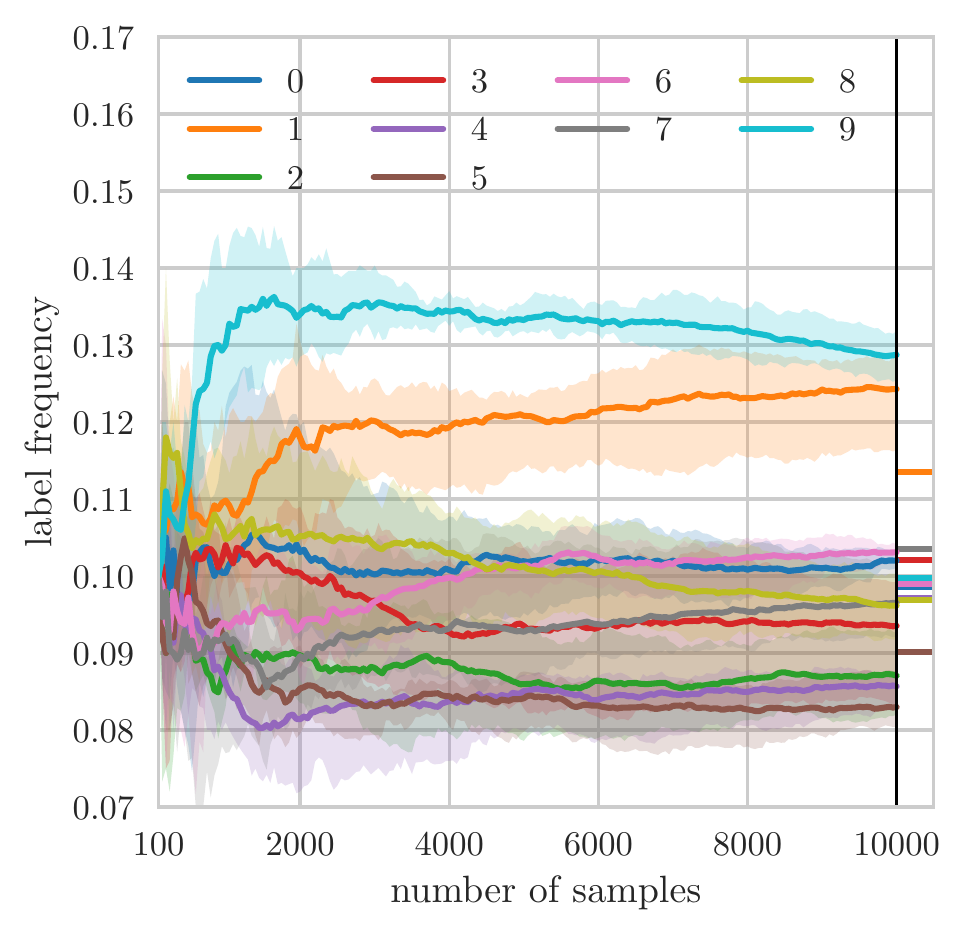}}
\hfill
\subfloat[ASAL-Discriminator with WGAN-GP.]{\includegraphics[width=0.33\textwidth, keepaspectratio]{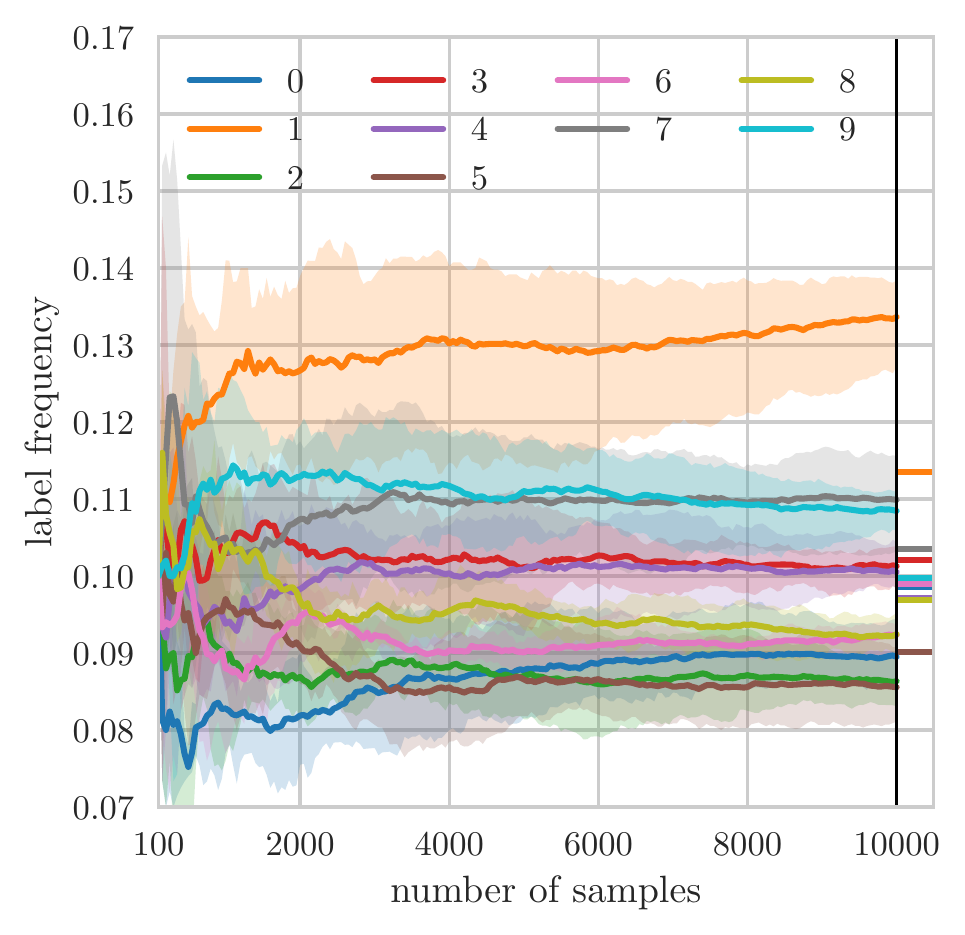}}

\caption{Label distribution for uncertainty sampling using maximum entropy, random sampling and active learning using different matching strategies and \acrshortpl{gan} for \emph{MNIST - ten classes}. The tick on the right show the true label distribution in the pool. Note the different scaling of the y-axis. Random sampling converges to the true label distribution in the pool and maximum entropy sampling leads to a training set with a higher ration of certain digits (\textsf{7,8,9}) or lower (\textsf{0,1,4,6}) than the pool. Similarly, \gls{asal} using WGAN-GP (bottom row) selects certain digits more frequently than others. Conversely, \gls{asal} using DCGAN (top row) leads to a training set that contains 30\% images with the digit \textsf{1}. Most likely, the DCGAN is responsible for this behaviour because we already observed that it produces the digit \textsf{1} more frequently than any other digit.}
\label{app:fig:mnist-all-label-dist}
\end{figure*}

\begin{figure*}[t]
\centering
\includegraphics[width=0.75\textwidth, keepaspectratio]{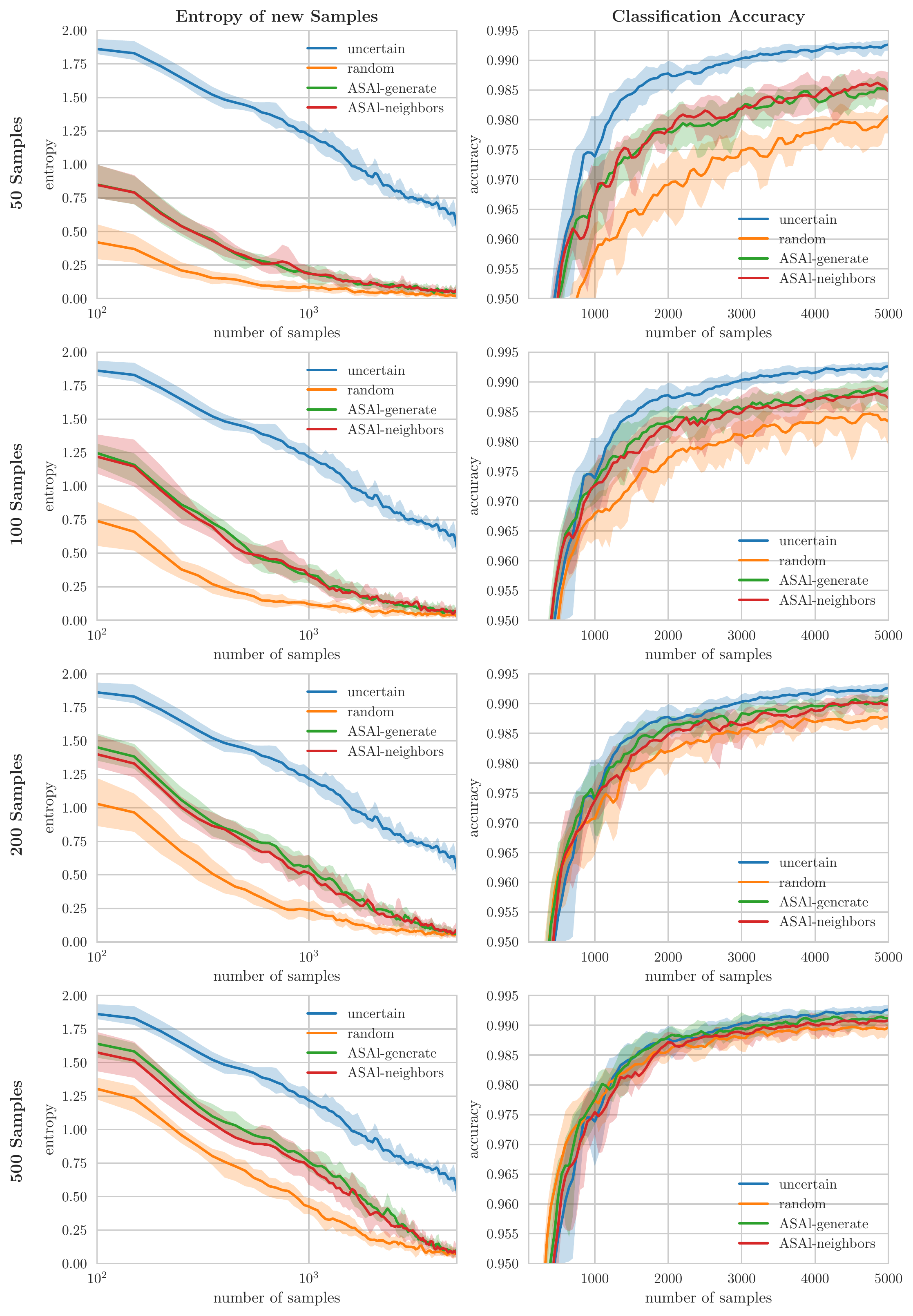} 
\caption{Comparison of random sampling combined with uncertainty sampling and \acrshort{asal} combined with uncertainty sampling. Instead of selecting only 50 random samples and label all of them, we randomly build a small subset containing a few hundred samples from the pool and retrieve the 50 most uncertain samples, we denote this setting as \textit{random}. For \textit{\acrshort{asal}-generate}, we generate more than the required 50 samples, match all of these samples and select the 50 most uncertain among all matched real samples. For \textit{\acrshort{asal}-neighbors}, we generate 50, match them but retrieve k instead of only the nearest neighbor and select the 50 most uncertain among all matched real neighbors. \textit{Uncertain} refers to classical uncertainty sampling that we show as a reference. We conclude, that using \acrshort{asal} to construct subsets and search for uncertain samples, leads to higher entropy of newly added samples and to a better classification accuracy on any subset size.}\label{app:fig:mnist-all-subsample}
\end{figure*}

Fig.~\ref{app:fig:mnist-all-test-acc} shows the test accuracy of \gls{asal} for different \glspl{gan}. Fig.~\ref{app:fig:mnist-all-label-dist} shows the label distribution during active learning for \gls{asal} an classical methods.
Fig.~\ref{app:fig:mnist-all-entropy} shows the entropy of newly added samples.
Fig.~\ref{app:fig:mnist-all-subsample} shows a comparison of two different methods: (i) randomly sub sampling the pool and scanning this subset for the sample with maximum entropy and (ii) using \gls{asal} to retrieve more samples than required and scan this subset for the samples with maximum entropy. Fig.~\ref{app:fig:mnist-all-subsample} shows, that for a fixed subset size, \gls{asal} combined with uncertainty sampling retrieves always  higher entropy samples than random sampling combined with uncertainty sampling. In addition, \gls{asal} achieves the best classification accuracy and approaches uncertainty sampling faster. 

\section{Additional Results: CIFAR - two classes}
\begin{figure*}[t]
\centering 
\subfloat[Random sampling with Hinge loss.]{\includegraphics[width=0.25\textwidth, keepaspectratio]{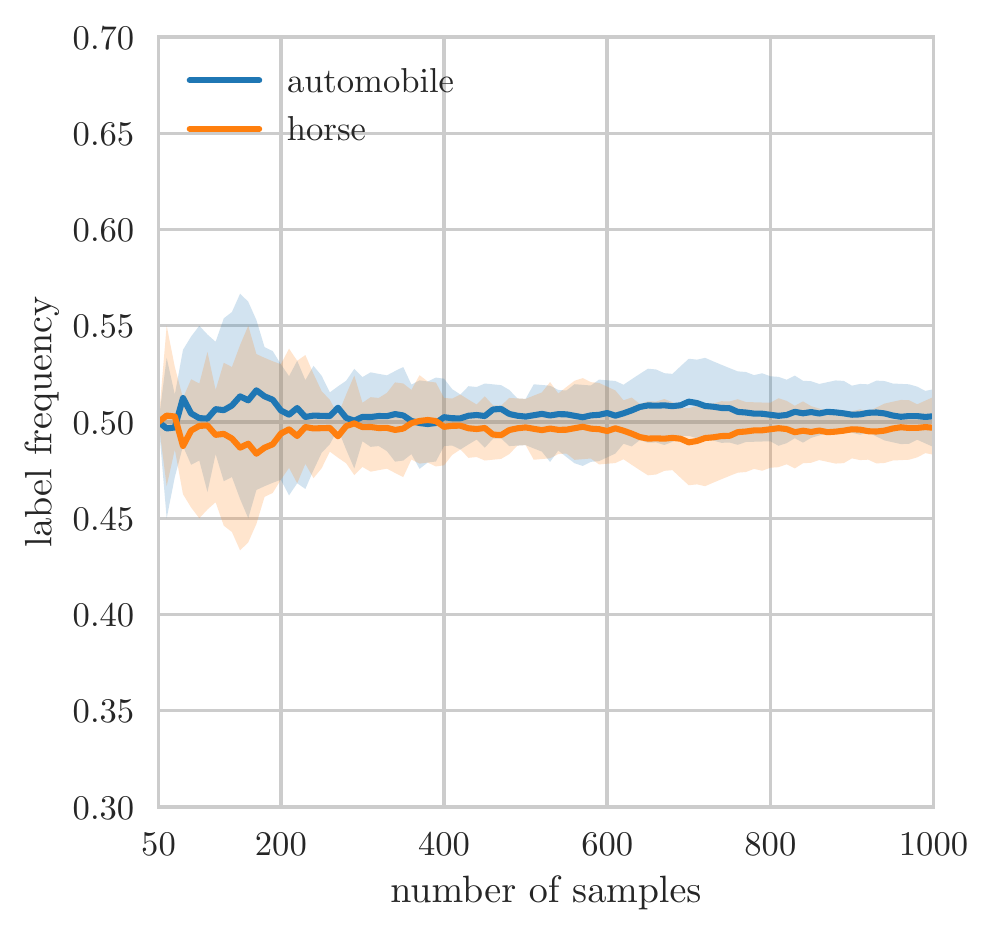}}
\subfloat[Random sampling with cross-entropy loss.]{\includegraphics[width=0.25\textwidth, keepaspectratio]{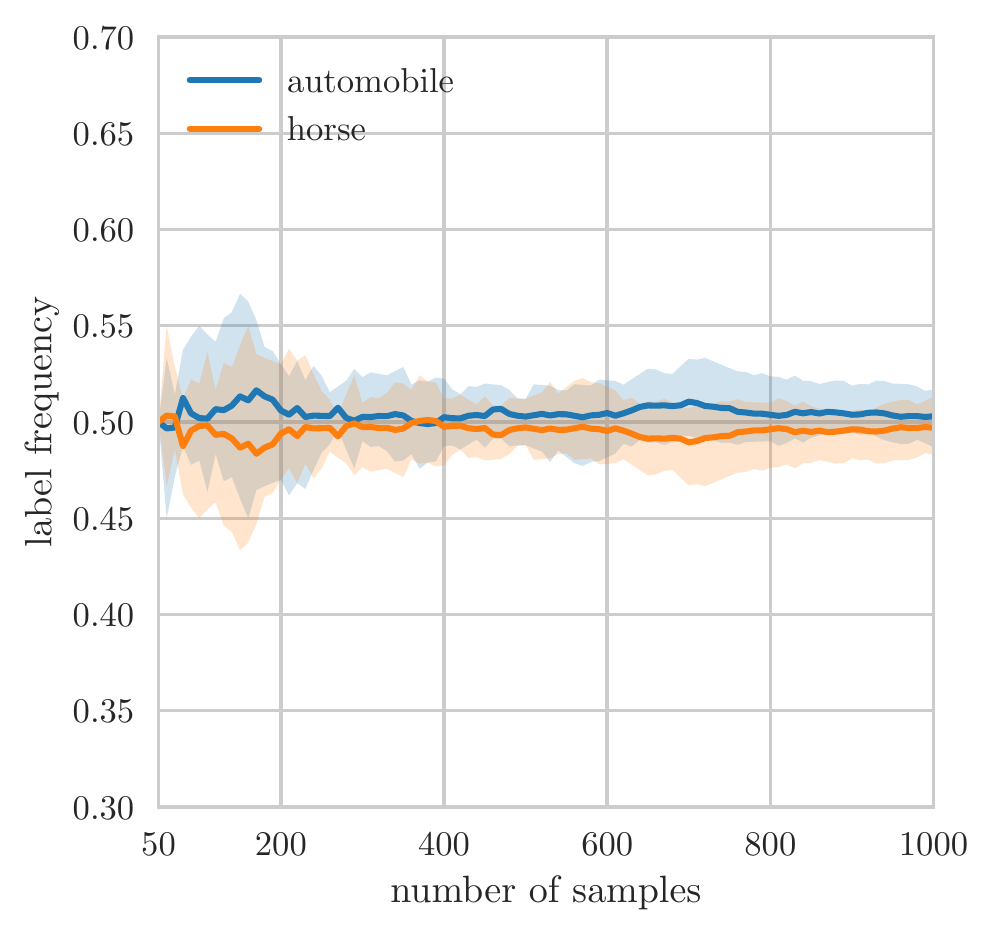}}
\subfloat[\emph{Minimum distance} sampling with Hinge loss.]{\includegraphics[width=0.25\textwidth, keepaspectratio]{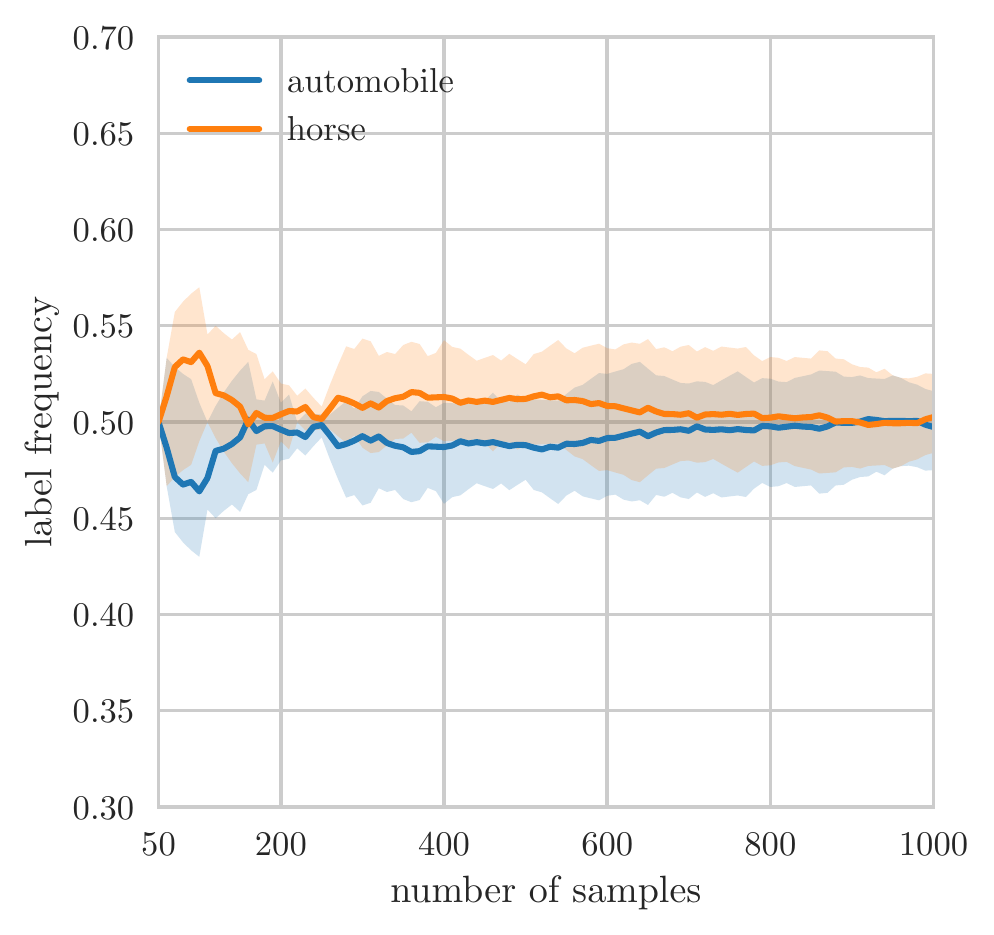}}
\subfloat[\emph{Maximum entropy} sampling with cross-entropy loss.]{\includegraphics[width=0.25\textwidth, keepaspectratio]{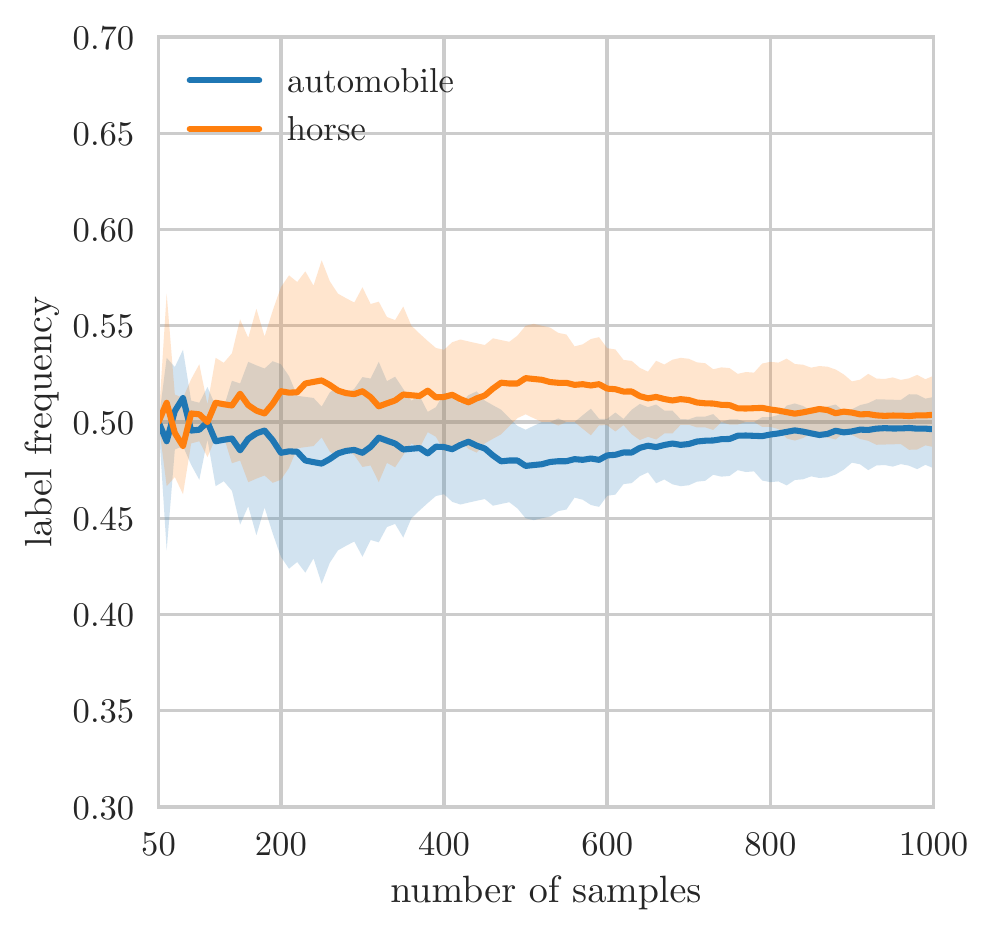}}

\caption{Label distribution for uncertainty sampling using maximum entropy and random sampling for \emph{CIFAR-10 - two classes} using different uncertainty measures and loss functions.The label distribution of the training set of all strategies converges to the true label distribution of the pool. However, in average over all active learning iterations the training set of the uncertainty sampling strategies most frequently contained the images with the label \textsf{horse}.}\label{app:fig:cifar-binary-label-dist}
\end{figure*}

\begin{figure*}[t]
\centering
\includegraphics[width=0.9\textwidth, keepaspectratio]{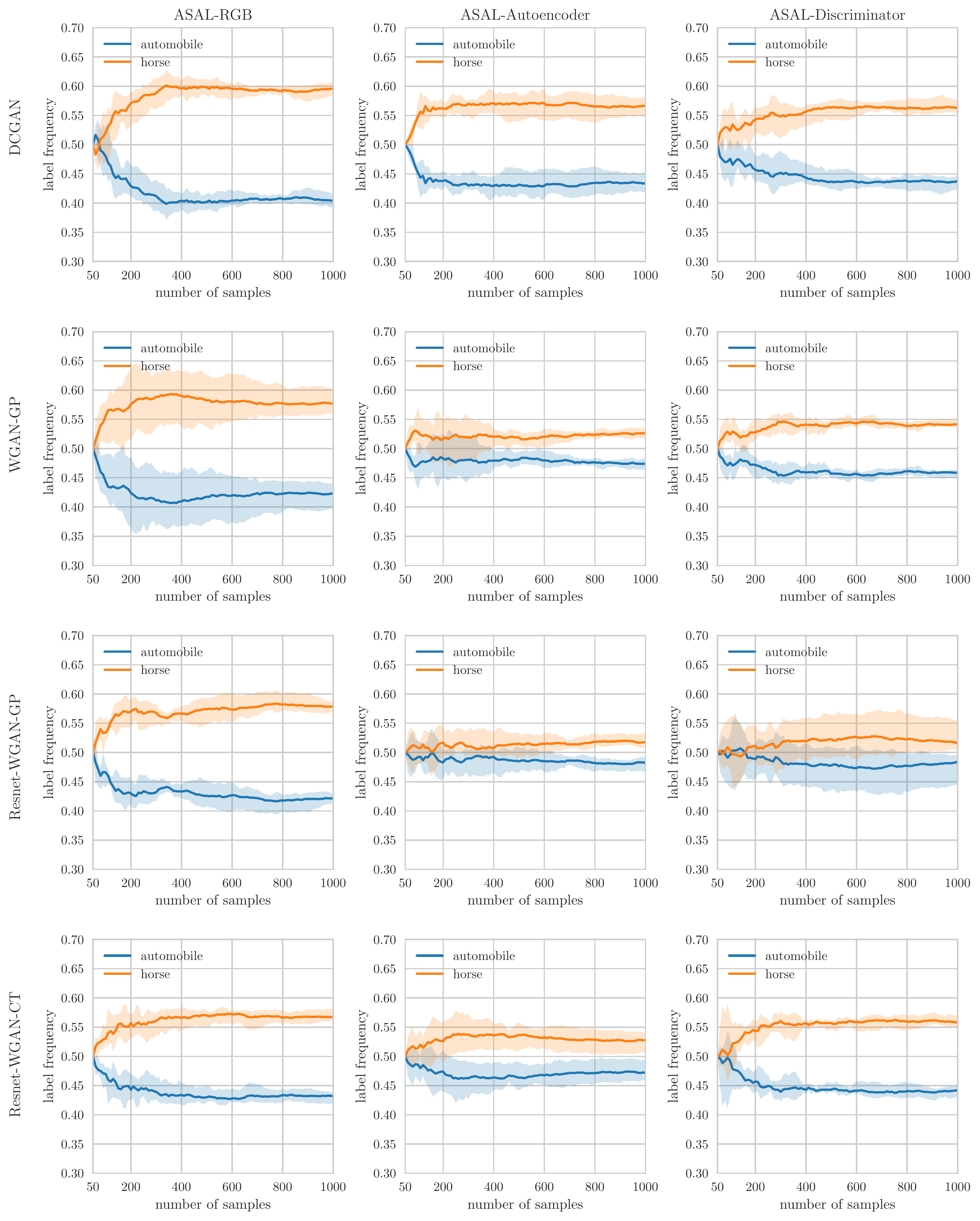} 
\caption{Label distribution for active learning with minimum distance sample generation and the Hinge loss, using different matching strategies and GANs for \emph{CIFAR-10 - two classes}. All setups assemble training sets containing the more image with the label \textsf{horse} than \textsf{automobile}.}\label{app:fig:cifar-binary-asal-label-dist-1}
\end{figure*}

\begin{figure*}[t]
\centering
\includegraphics[width=0.9\textwidth, keepaspectratio]{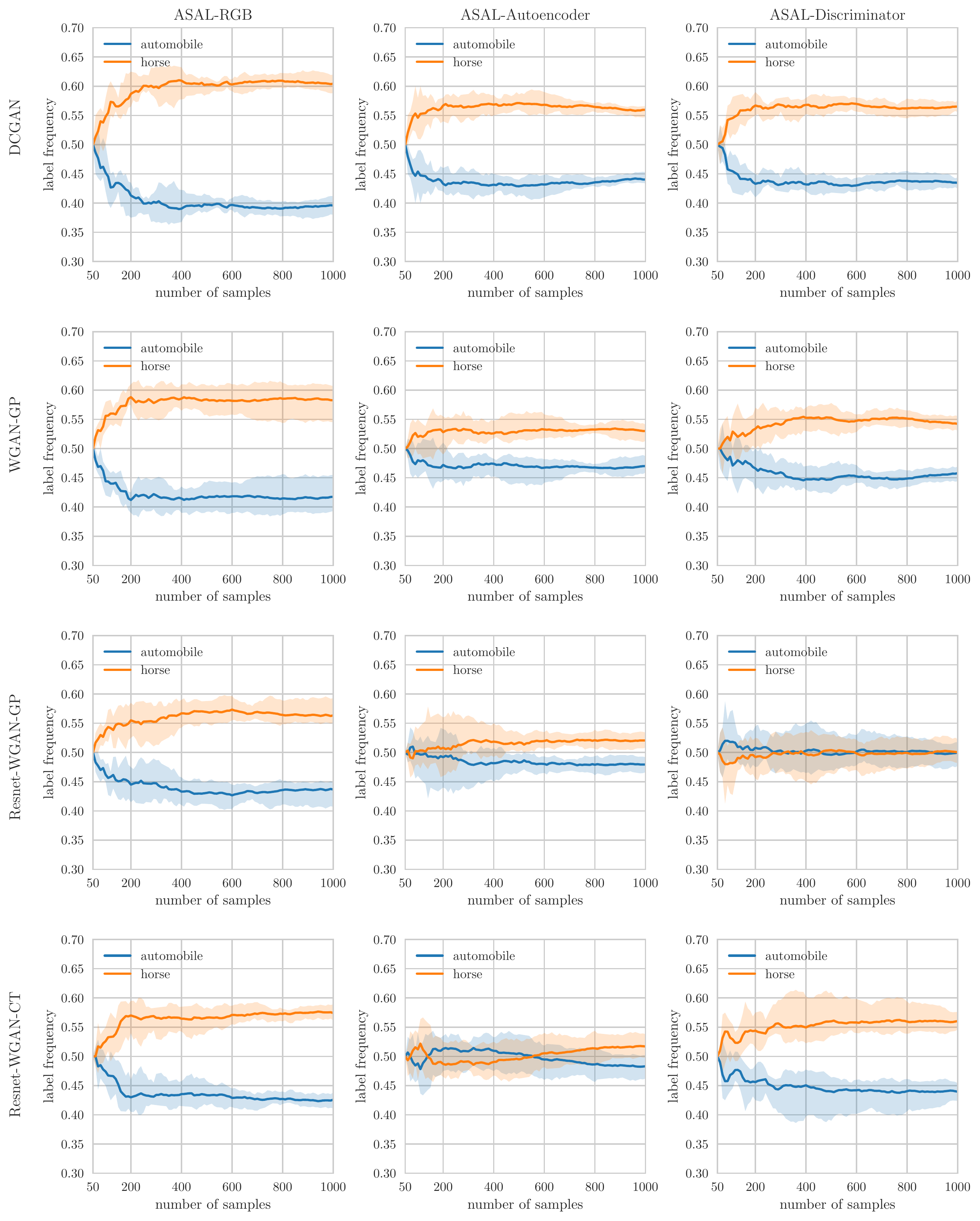} 
\caption{Label distribution for active learning with maximum entropy sample generation and the cross-entropy loss, using different matching strategies and GANs for \emph{CIFAR-10 - two classes}.}\label{app:fig:cifar-binary-asal-label-dist-2}
\end{figure*}

\begin{figure*}[t]
\centering
\includegraphics[width=0.6\textwidth, keepaspectratio]{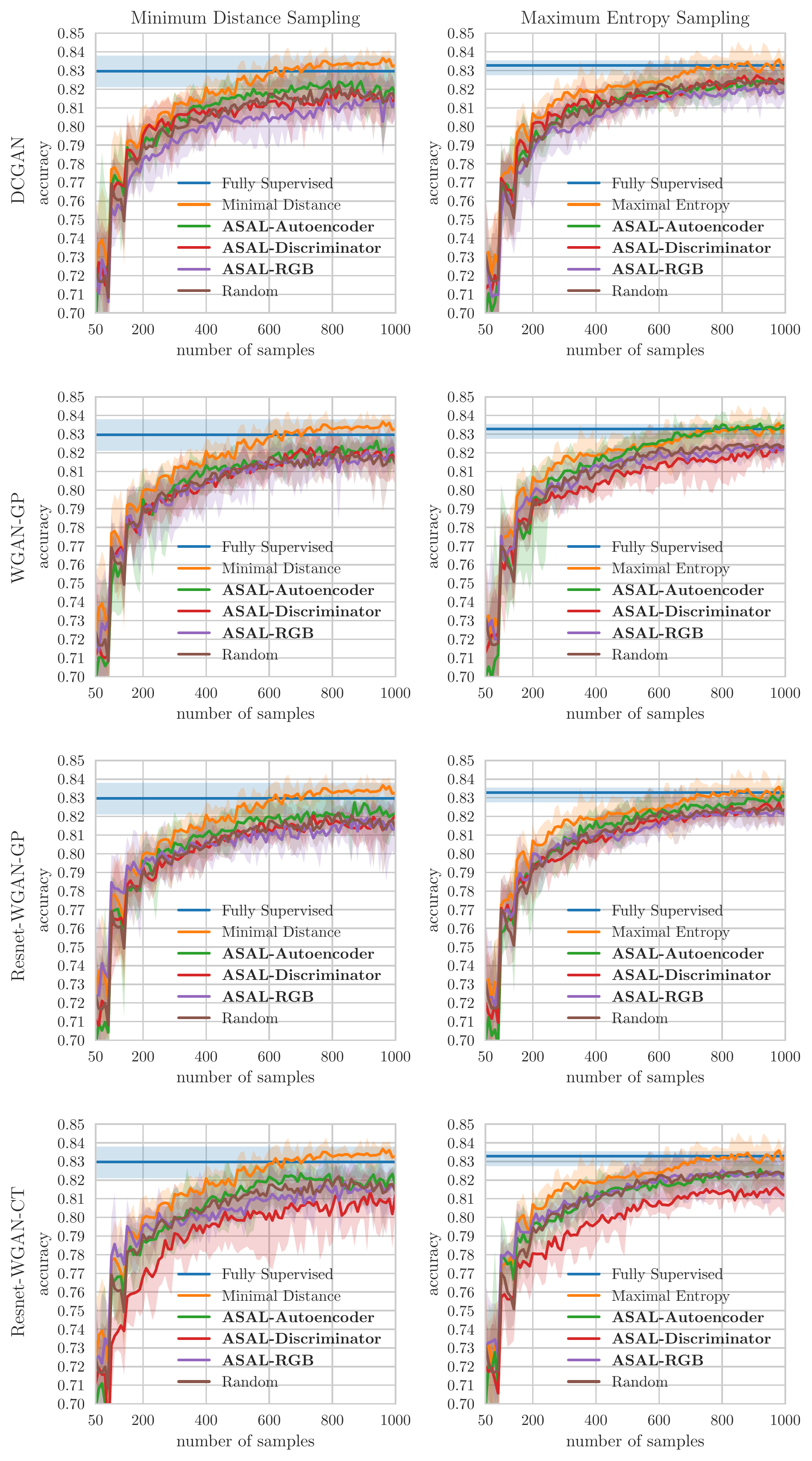}
\caption{Validation accuracy on \emph{CIFAR-10 - two classes} of a fully supervised model, for random sampling, uncertainty sampling and different \acrshortpl{asal} using different \acrshortpl{gan}. \acrshort{asal}-Autoencoder leads to the best performance. \acrshort{asal}-Disc. using Resnet-WGAN-CT performs worse that any other strategy because the sample matching using is unable to retrieve high entropy samples from the pool, see Fig.~\ref{app:fig:cifar-binary-new-mean-entropies}.}
\label{app:fig:cifar-binary-val-acc}
\end{figure*}

\begin{figure*}[t]
\centering
\includegraphics[width=0.6\textwidth, keepaspectratio]{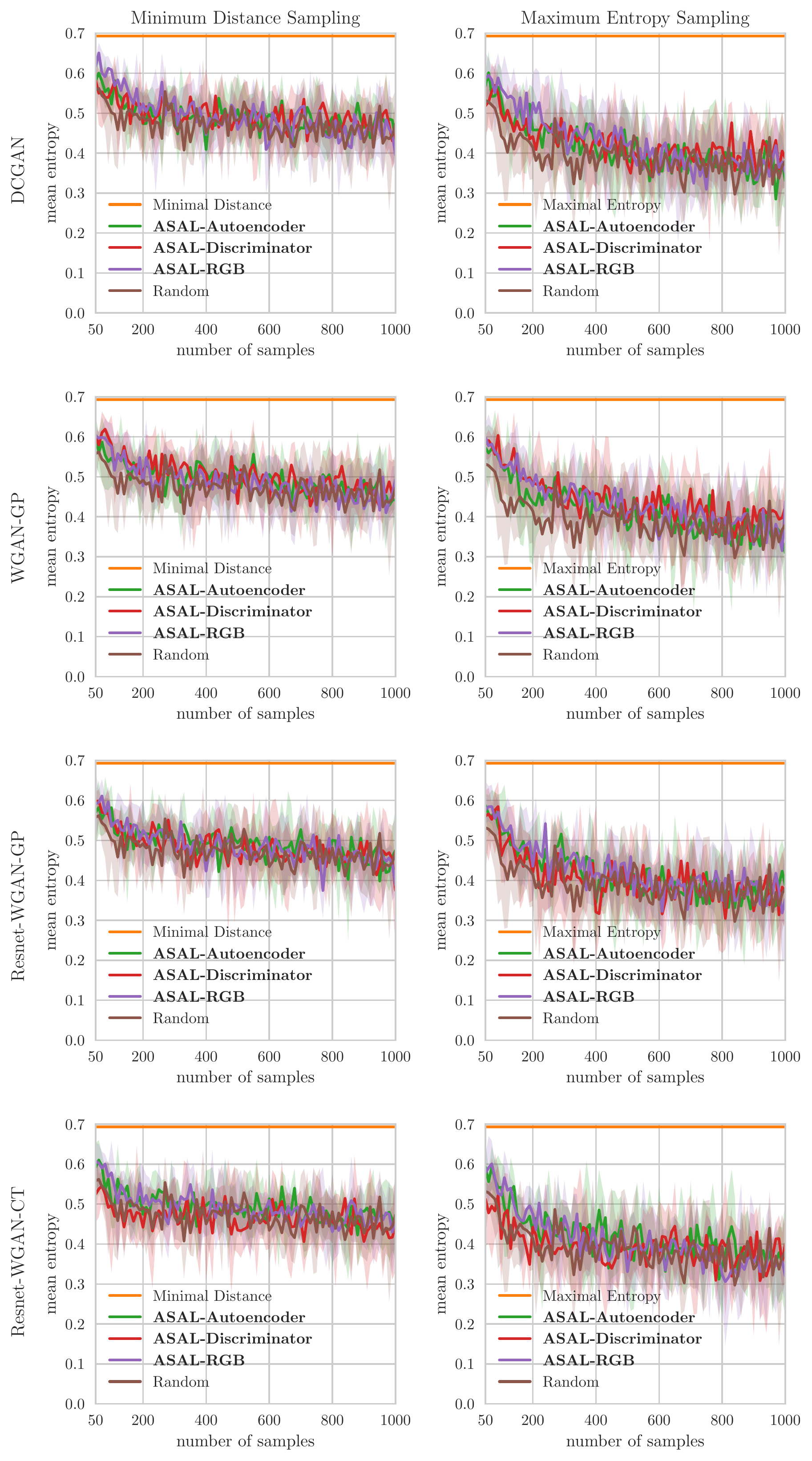}

\caption{Average entropy of images that are selected and added to the training set for \emph{CIFAR-10 - two classes} using different \acrshortpl{gan}. The mean entropy of the random sampling and the proposed method show hardly any difference. However, for maximum entropy sampling at least at the beginning \acrshort{asal} selects images with higher entropy than random sampling.}
\label{app:fig:cifar-binary-new-mean-entropies}
\end{figure*}

Fig.~\ref{app:fig:cifar-binary-val-acc} shows the test accuracy of \gls{asal} for different \glspl{gan}. Fig.~\ref{app:fig:cifar-binary-label-dist},~\ref{app:fig:cifar-binary-asal-label-dist-1} and~\ref{app:fig:cifar-binary-asal-label-dist-2}  shows the label distribution during active learning for \gls{asal} an classical methods.
For CIFAR-10, we do not indicate the true label distribution by a tick because the validation set contains the same number of samples for each class.
Fig.~\ref{app:fig:cifar-binary-new-mean-entropies} shows the entropy of newly added samples.

\section{Additional Results: CIFAR - ten classes}
\begin{figure*}[t]
\centering
\subfloat[DCGAN.]{\includegraphics[width=0.25\textwidth, keepaspectratio]{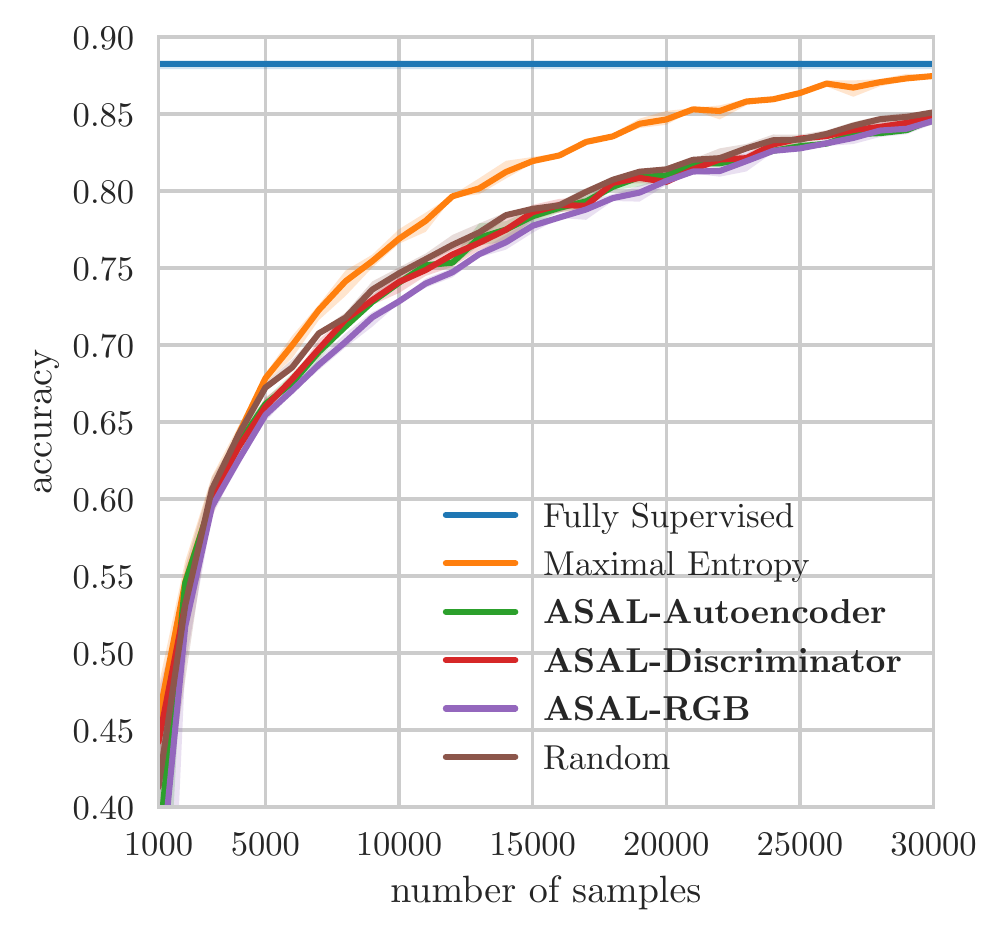}}
\subfloat[WGAN-GP.]{\includegraphics[width=0.25\textwidth, keepaspectratio]{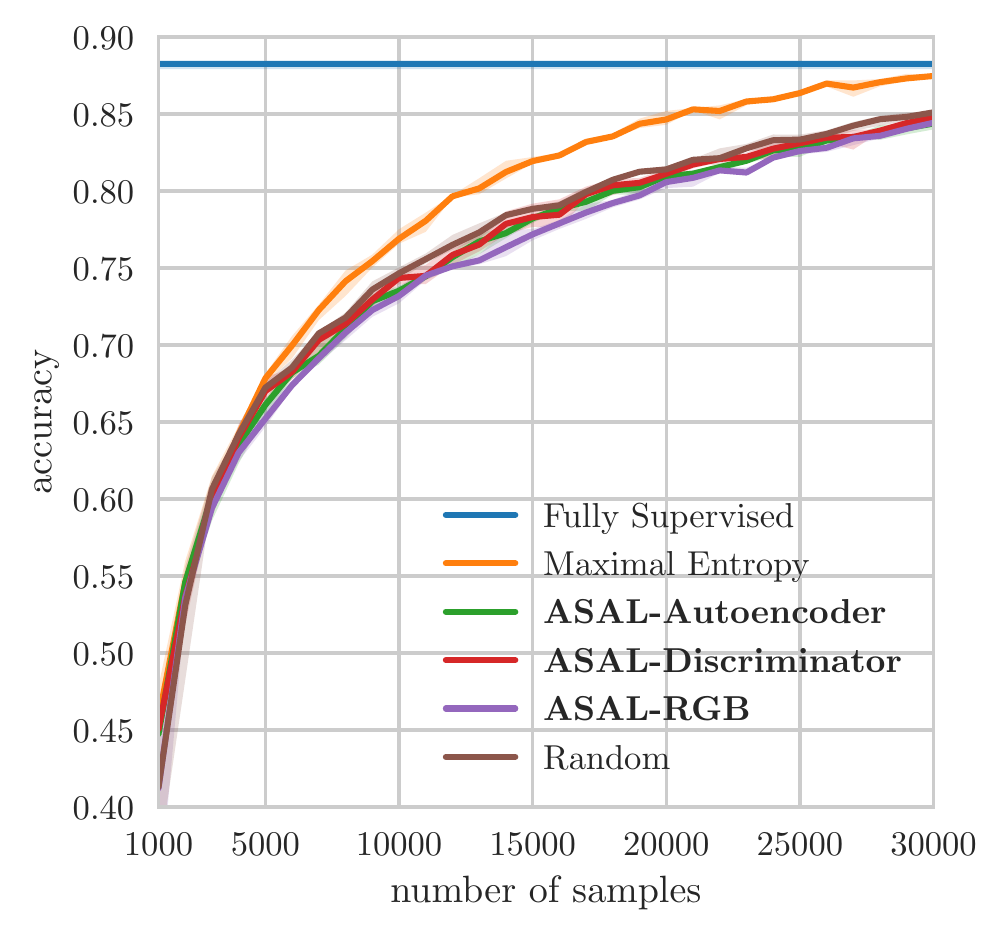}}
\subfloat[Resnet-WGAN-GP.]{\includegraphics[width=0.25\textwidth, keepaspectratio]{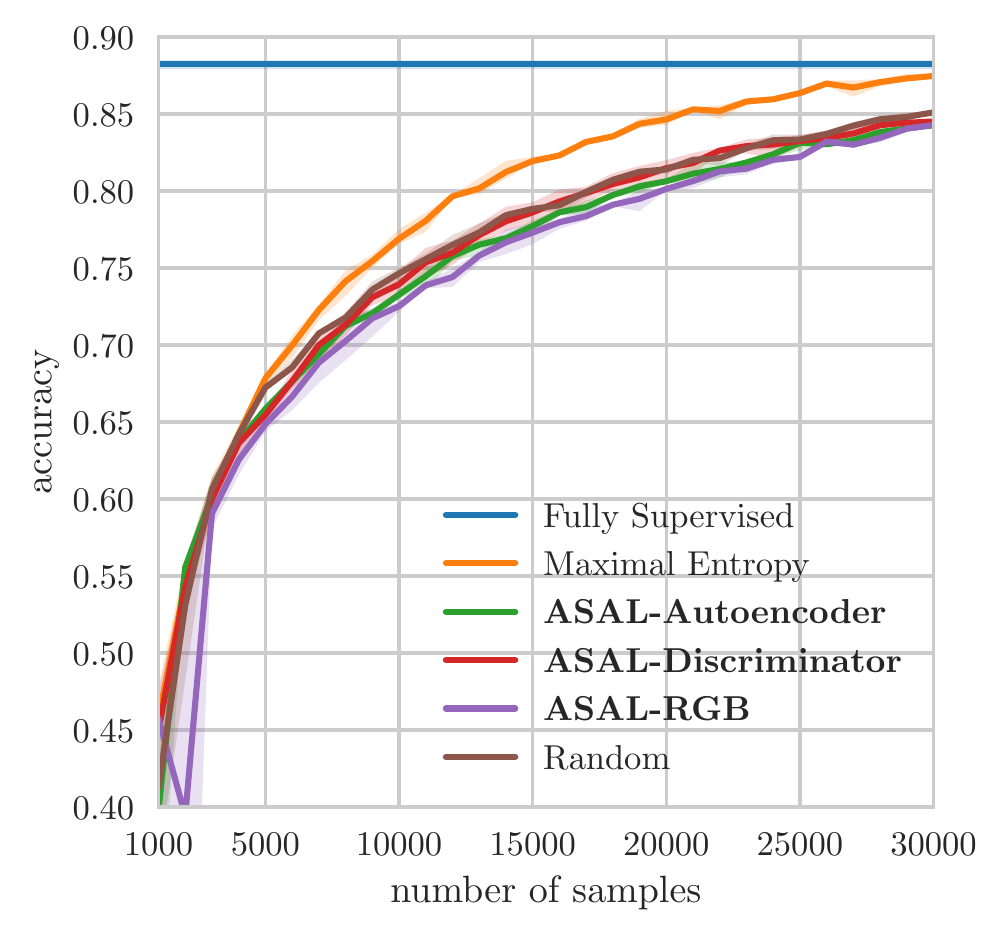}}
\subfloat[Resnet-WGAN-CT.]{\includegraphics[width=0.25\textwidth, keepaspectratio]{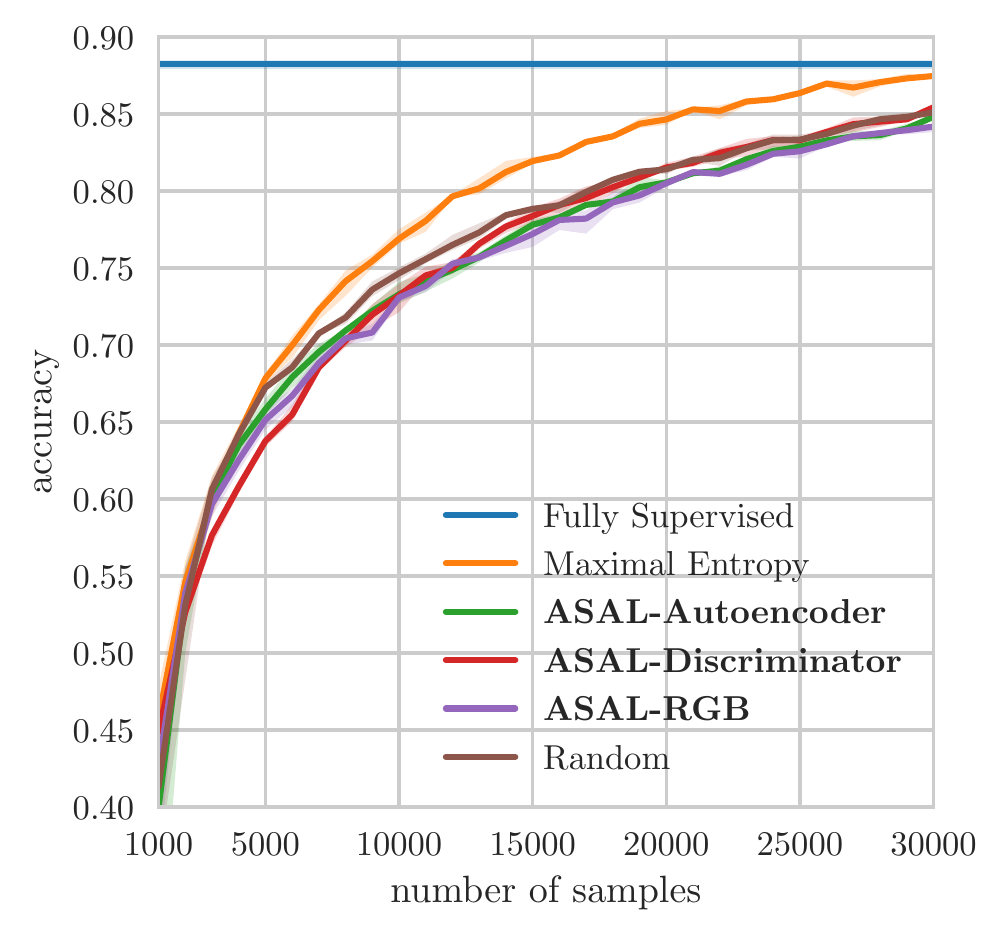}}

\caption{Validation accuracy on \emph{CIFAR-10 - ten classes} of a fully supervised model, for random sampling, uncertainty sampling and different \acrshortpl{asal} using different \acrshortpl{gan}. The proposed method performs slightly worse than random sampling independent of the sample matching of \acrshort{gan}.}
\label{app:fig:cifar-all-val-acc}
\end{figure*}

\begin{figure*}[t]
\centering
\subfloat[DCGAN.]{\includegraphics[width=0.25\textwidth, keepaspectratio]{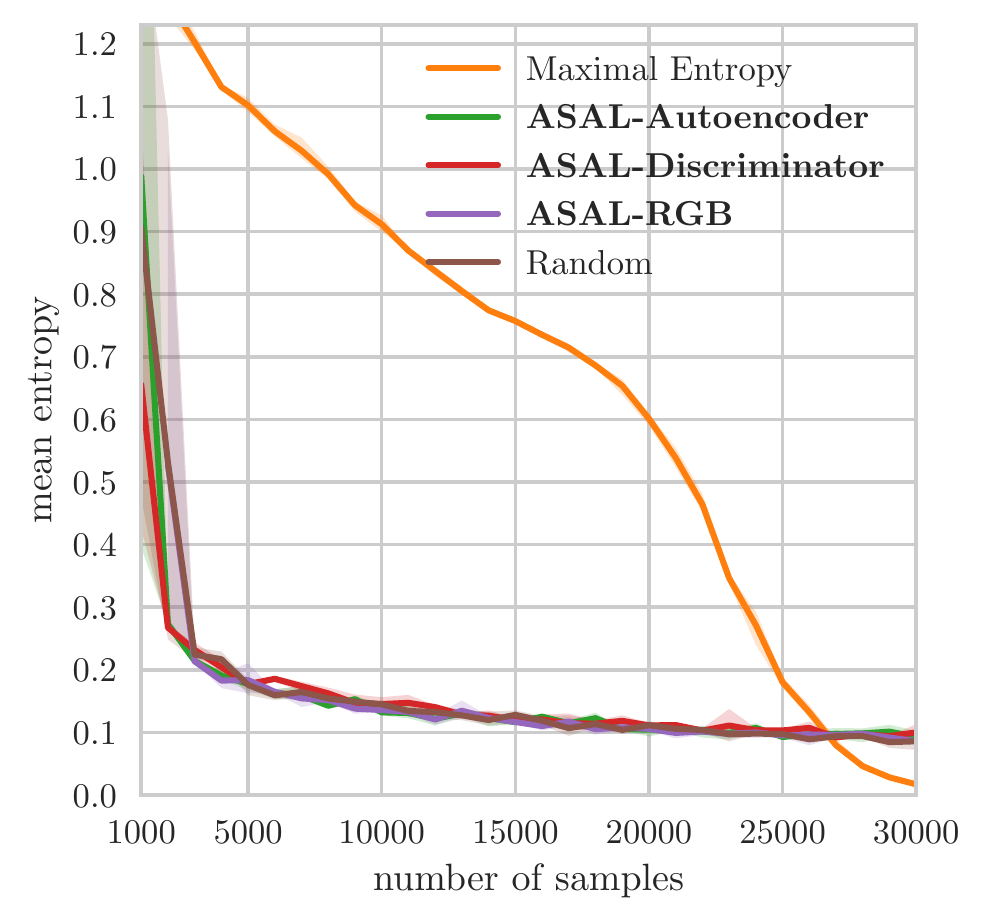}}
\subfloat[WGAN-GP.]{\includegraphics[width=0.25\textwidth, keepaspectratio]{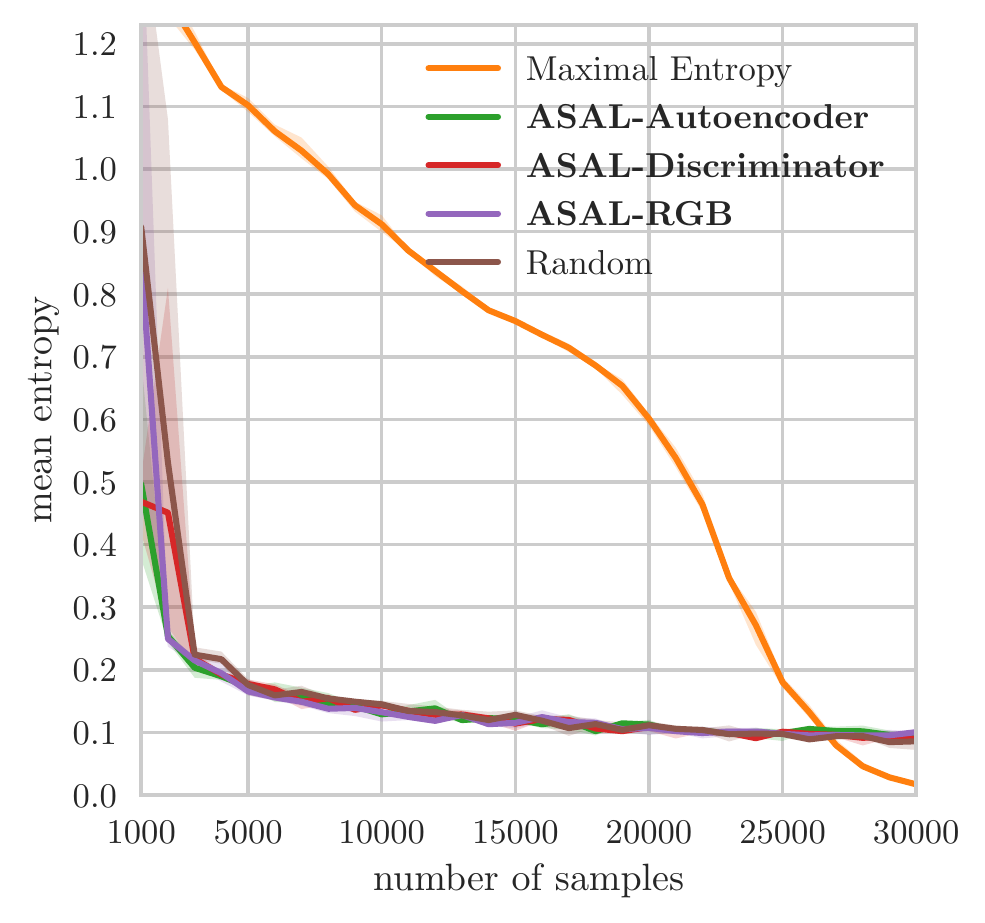}}
\subfloat[Resnet-WGAN-GP.]{\includegraphics[width=0.25\textwidth, keepaspectratio]{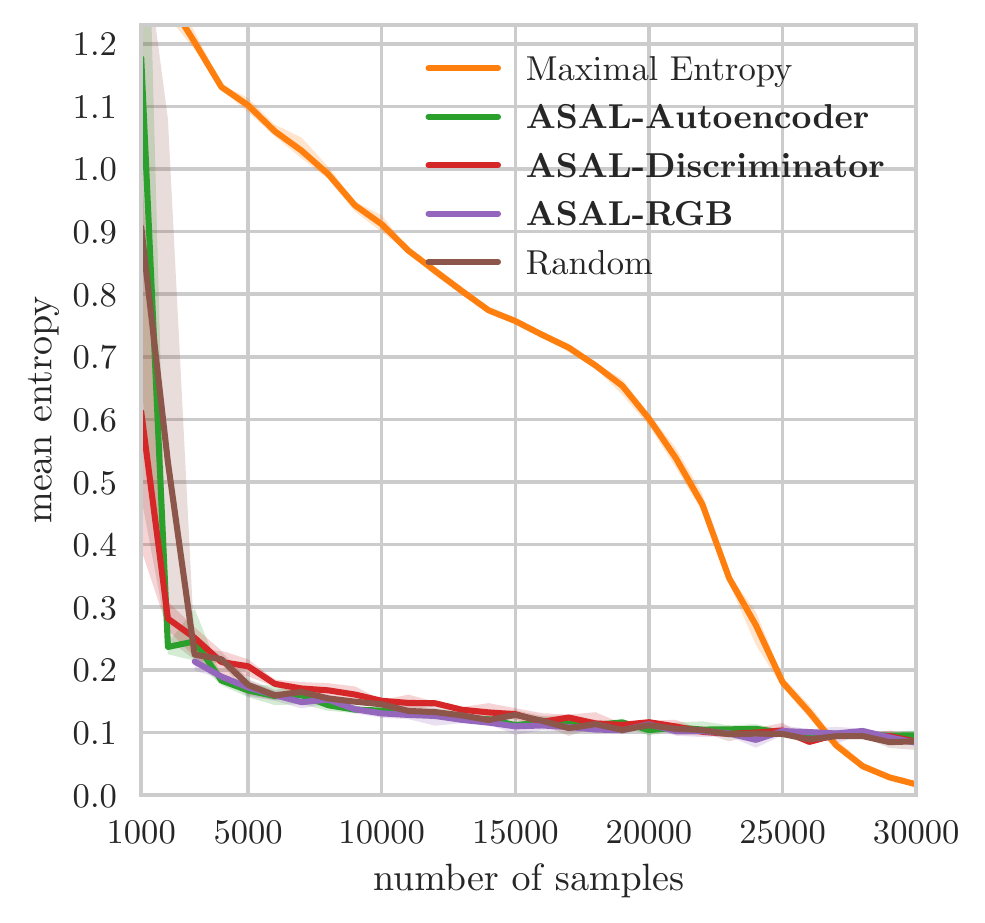}}
\subfloat[Resnet-WGAN-CT.]{\includegraphics[width=0.25\textwidth, keepaspectratio]{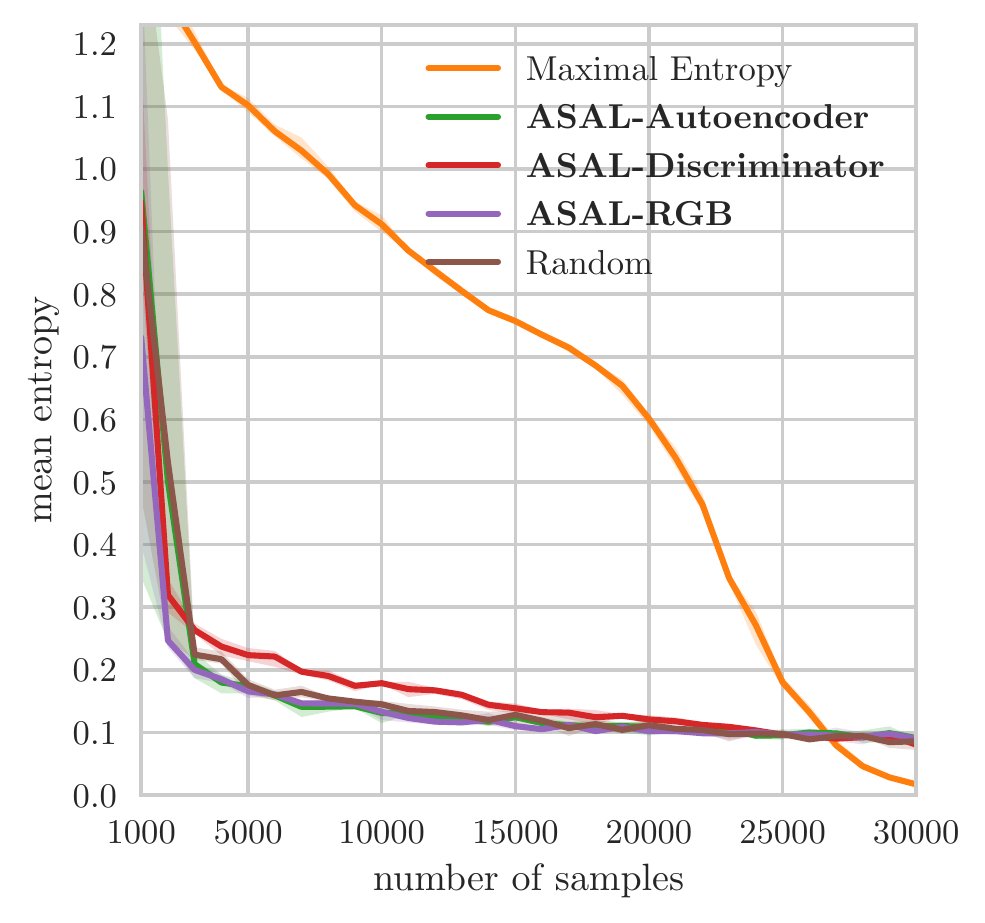}}

\caption{Average entropy of images that are selected and added to the training set for \emph{CIFAR-10 - ten classes} using different \acrshortpl{gan}. There is hardly any difference for random sampling and \acrshort{asal} in the entropy of newly added samples. Only at the beginning, random sampling retrieves samples with slightly higher entropy.}\label{app:fig:cifar-all-entropy}
\end{figure*}

\begin{figure*}[t]
\centering 
\subfloat[Random sampling.]{\includegraphics[width=0.25\textwidth, keepaspectratio]{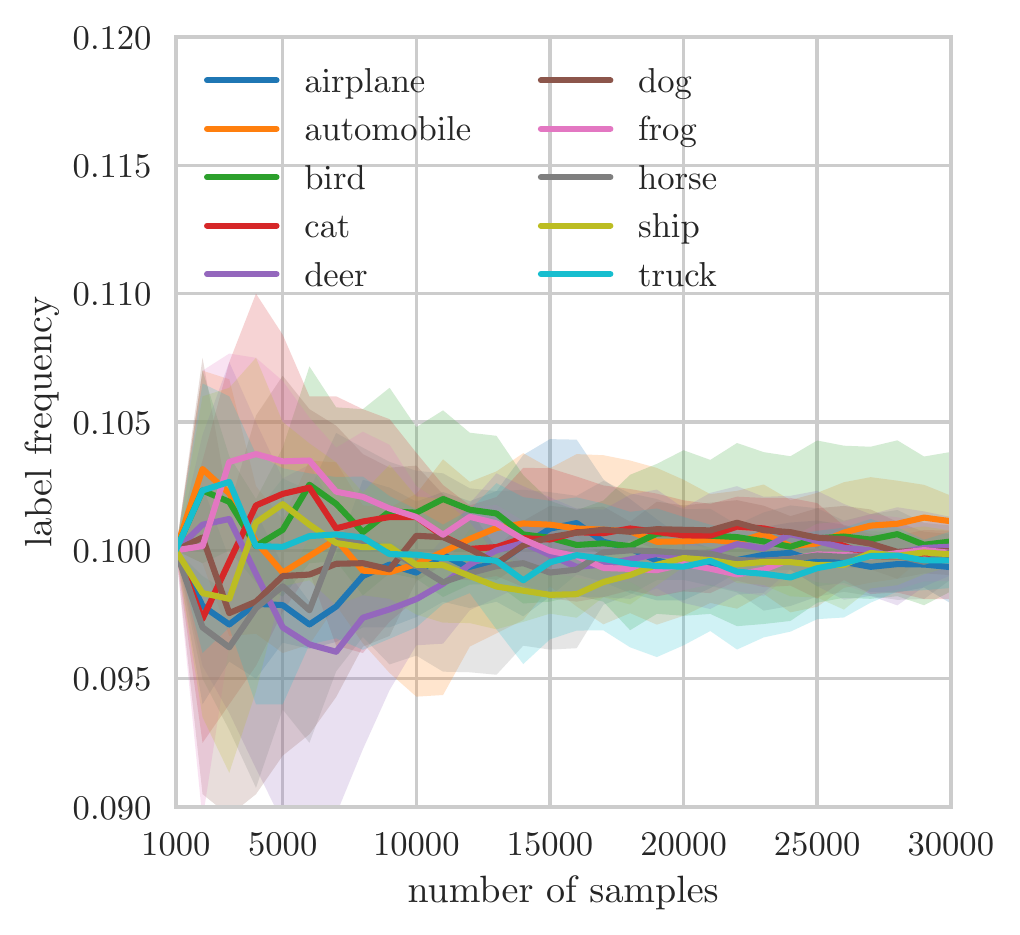}}
\subfloat[\emph{Maximum entropy} sampling.]{\includegraphics[width=0.25\textwidth, keepaspectratio]{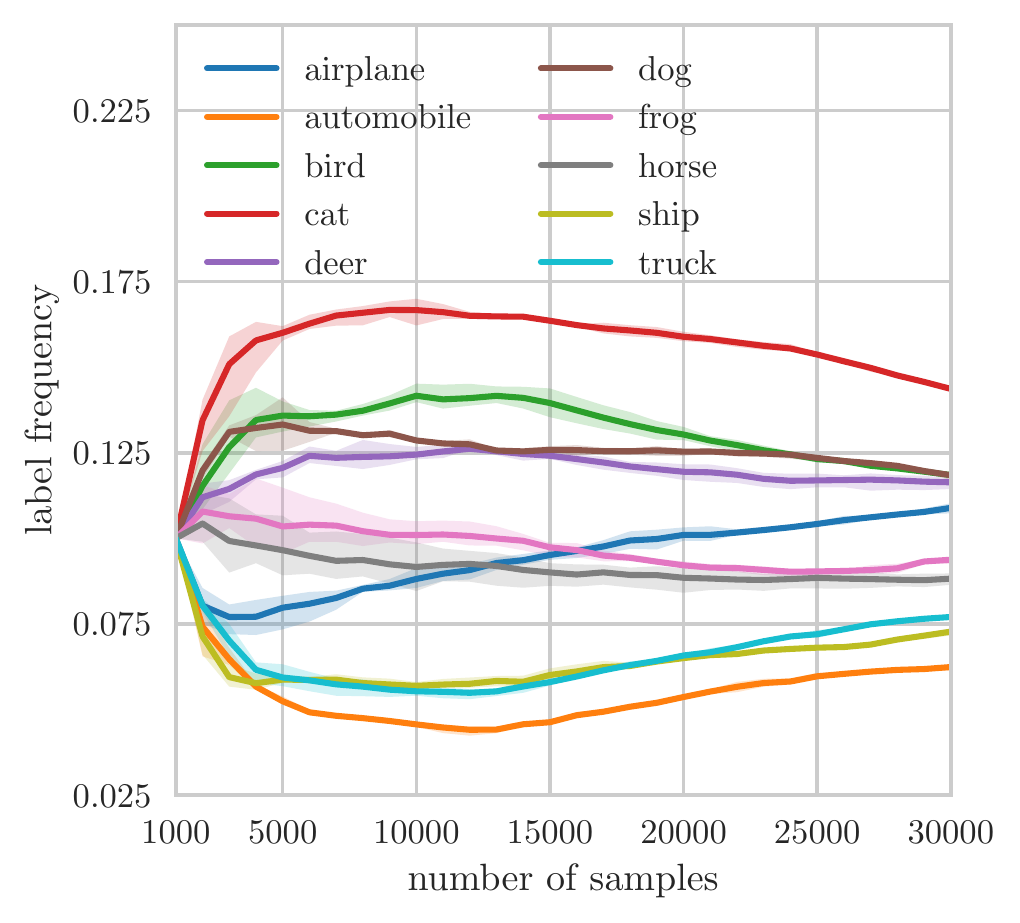}}
\subfloat[\emph{ASAL Classifer Features} sampling.]{\includegraphics[width=0.25\textwidth, keepaspectratio]{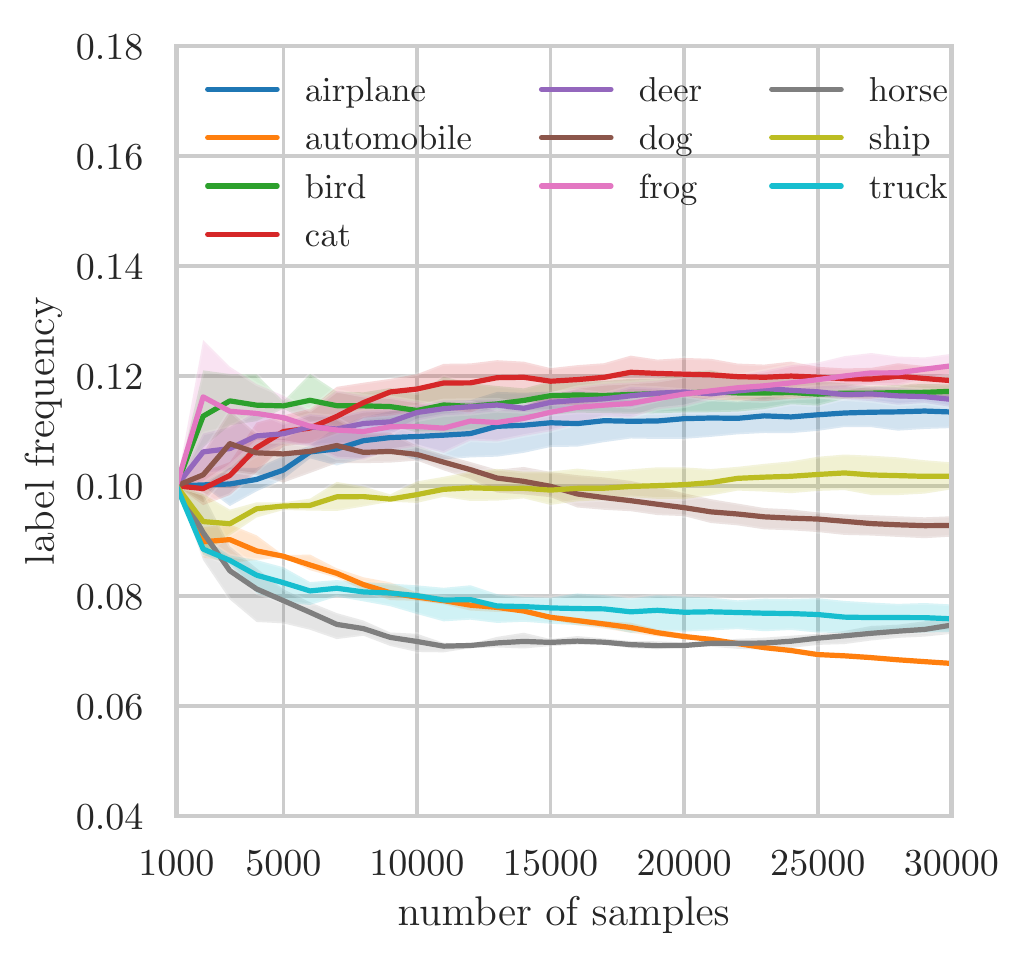}}

\caption{Label distribution for uncertainty sampling using maximum entropy and random sampling for \emph{CIFAR-10 - ten classes}. Random sampling converges to the true label distribution in the pool. Maximum entropy sampling selects most frequently \textsf{cat, dog, bird, deer} and least frequently \textsf{automobile, ship, truck} to exceed the classification quality of random sampling.}
\label{app:fig:cifar-all-label-dist}
\end{figure*}

Fig.~\ref{app:fig:cifar-all-val-acc} shows the test accuracy of \gls{asal} for different \glspl{gan}. Fig.~\ref{app:fig:cifar-all-label-dist} and~\ref{app:fig:cifar-all-label-dist-asal}  shows the label distribution during active learning for \gls{asal} and classical methods.
For CIFAR-10, we do not indicate the true label distribution by a tick because the validation set contains the same number of samples for each class.
Fig.~\ref{app:fig:cifar-all-entropy} shows the entropy of newly added samples.

\subsection{Discussion of results on CIFAR-10 - ten classes}

\begin{figure*}[t]
\centering
\includegraphics[width=0.9\textwidth, keepaspectratio]{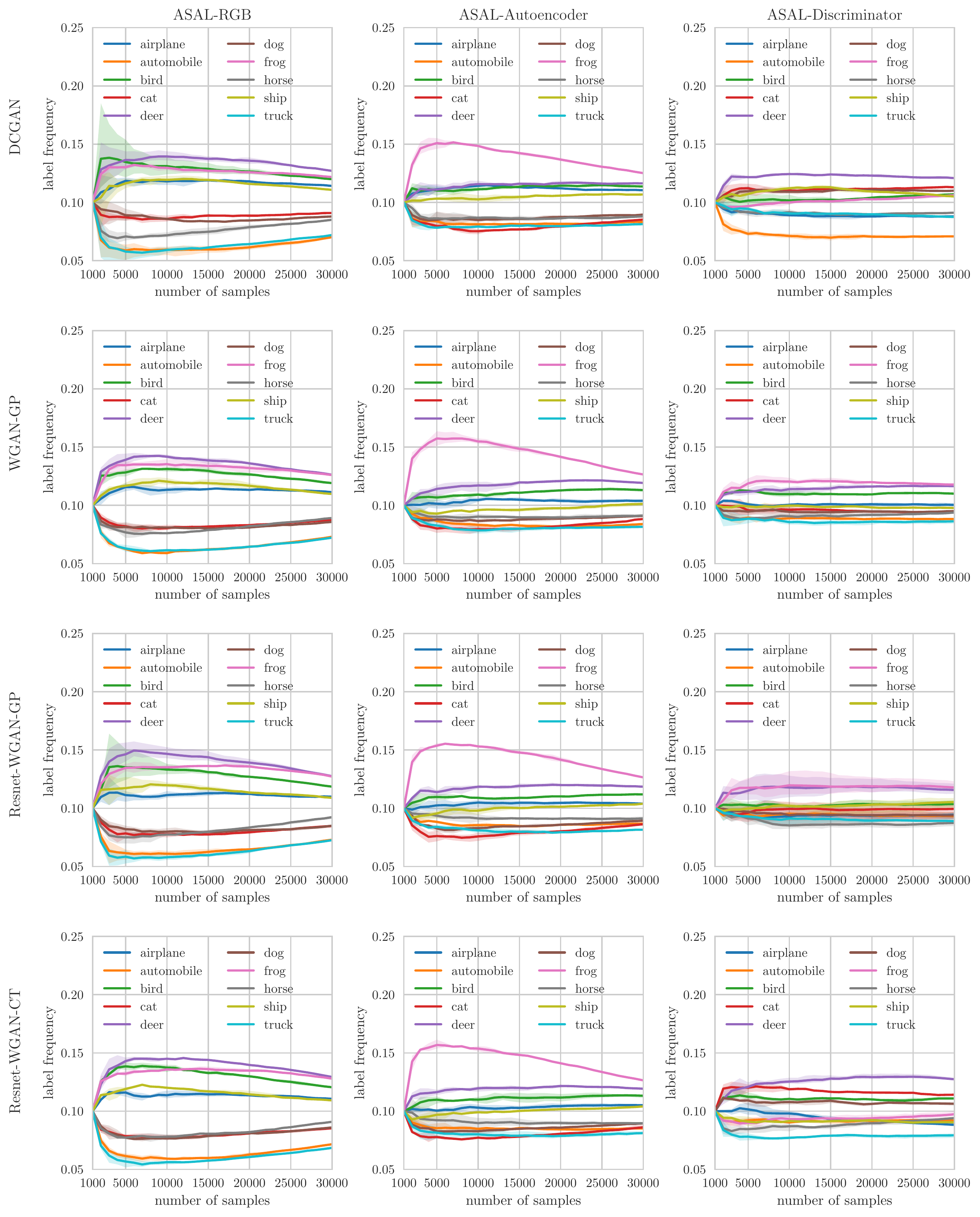} 
\caption{Label distribution for active learning using different matching strategies, uncertainty measures and \acrshortpl{gan} for \emph{CIFAR-10 - ten classes}. Exactly the classes \textsf{cat, dog} that are most common in the training set of uncertainty sampling are less common in the data sets of most setups. Conversely, \textsf{frog} is for many setups the most common class but is not particularly frequent in the uncertainty sampling data set.} 
\label{app:fig:cifar-all-label-dist-asal}
\end{figure*}

Fig.~\ref{app:fig:cifar-all-label-dist-asal} shows that maximum entropy sampling includes most frequently images showing cats, exactly one of the labels least frequent when using \gls{asal} with auto-encoder features. We observe that using $F_\textrm{CLS}$ leads to a similar distribution although less distinctive. Therefore, we conclude that instead of generating images of cat, the generator produces images of various classes and moves them close to the decision boundary to increase the entropy. However, note that truck and automobile are among the least frequent classes in any experiment and we conclude that the sample generating process is aware that these classes lead to small entropy and produces samples showing other classes.

\section{Matching Strategy Visualization}

\begin{figure*}[t]
\centering
\includegraphics[width=0.9\textwidth, keepaspectratio]{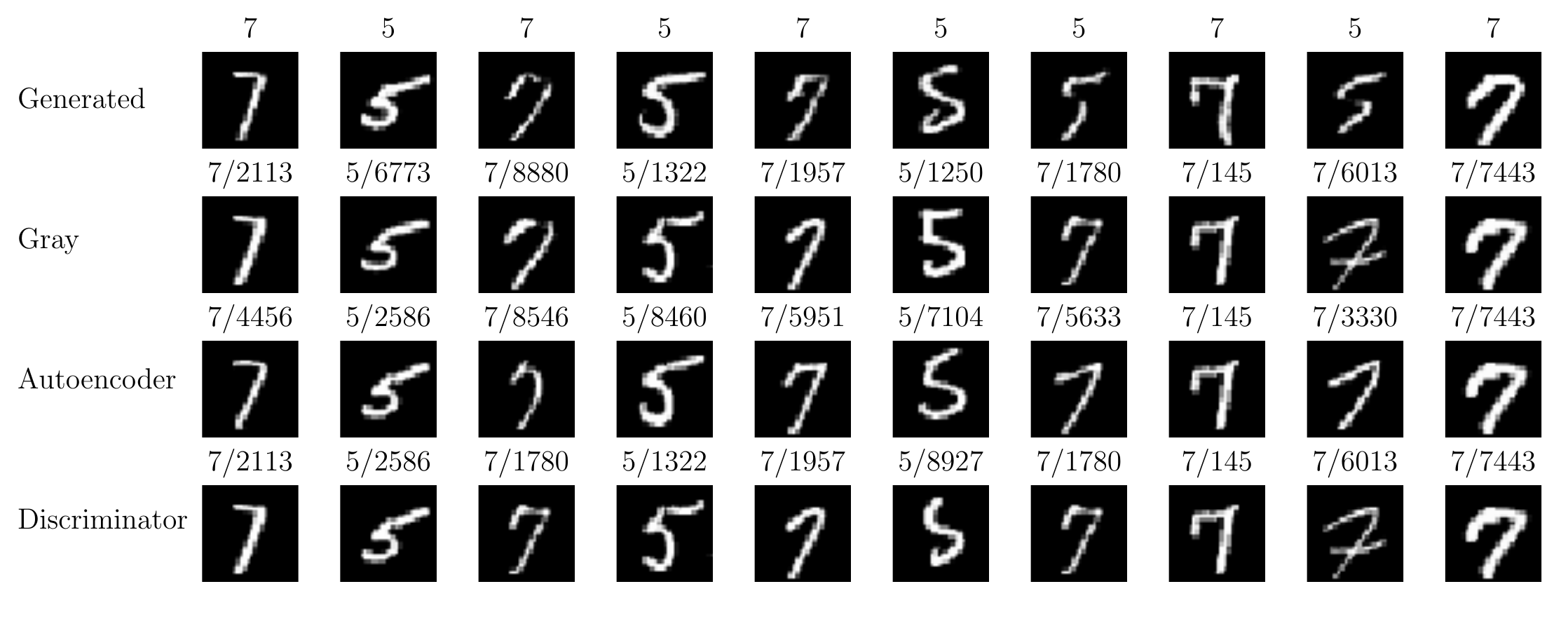}
\caption{The first row shows synthetic digits and the other the closest samples from the pool using different features for comparison. The numbers above the image denote the label and image id.}
\label{app:fig:mnist-binary-matching}
\end{figure*}

\begin{figure*}[t]
\centering 
\subfloat{\includegraphics[width=0.9\textwidth, keepaspectratio]{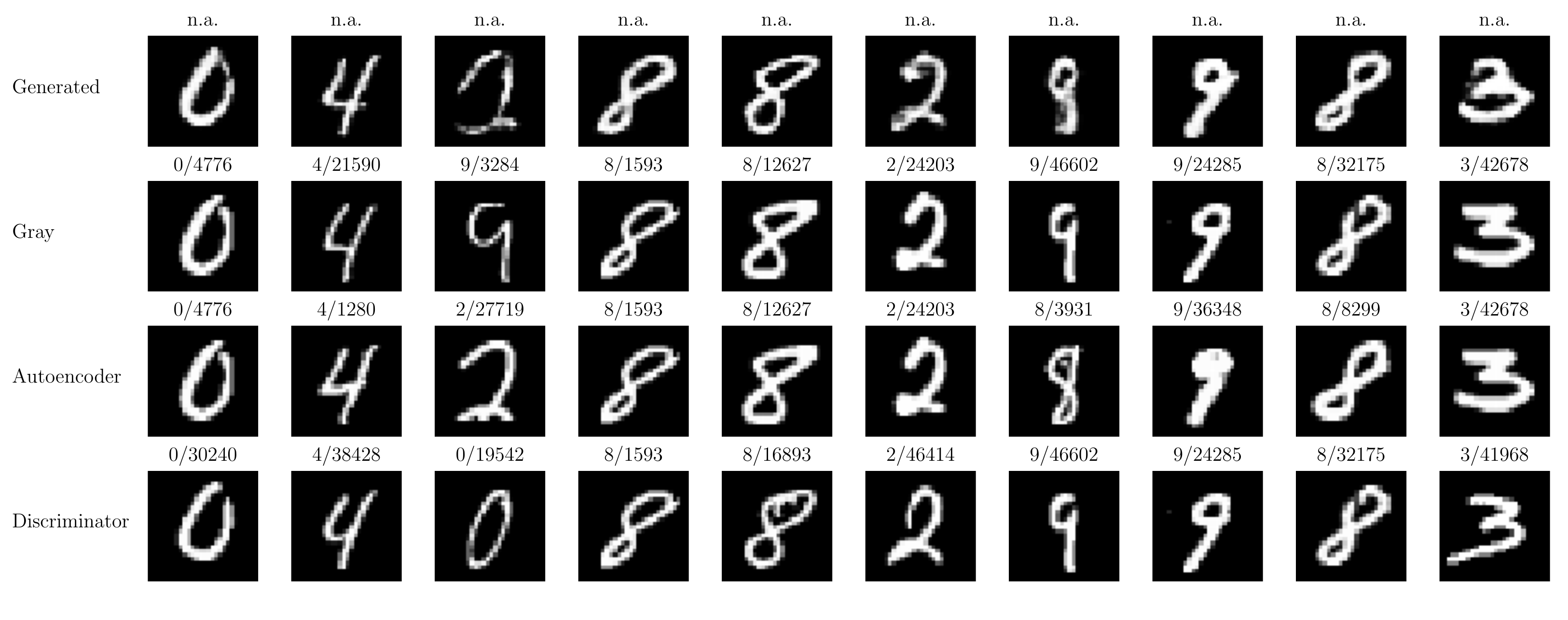}}

\caption{The rows show generated and matched images for \emph{MNIST - ten classes} using WGAN-GP.}
\label{app:fig:mnist-all-matching}
\end{figure*}

\begin{figure*}[t]
\centering 
\subfloat{\includegraphics[width=0.9\textwidth, keepaspectratio]{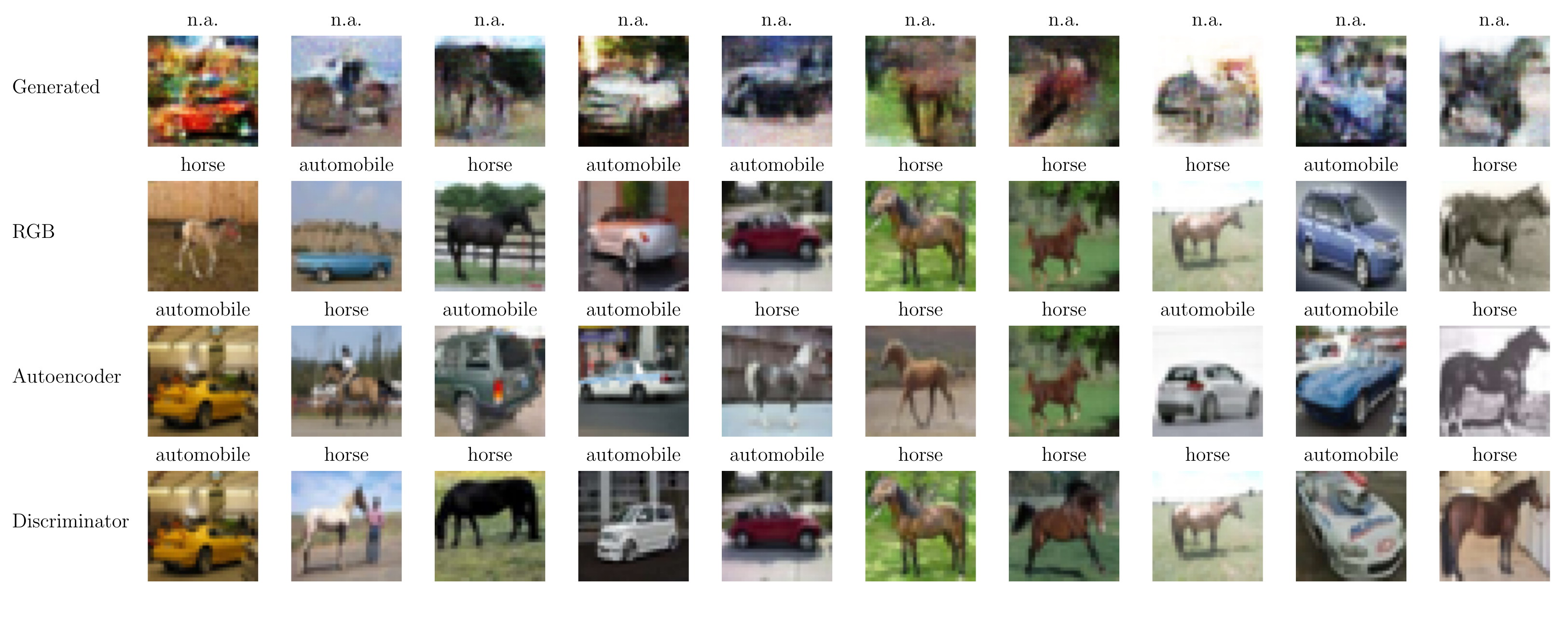}}

\caption{The rows show generated and matched images for \emph{CIFAR-10 - two classes} using WGAN-GP. The images have a reasonable quality and all matching strategies retrieve images that are visually close or show the same class.}
\label{app:fig:cifar-binary-matching}
\end{figure*}

\begin{figure*}[t]
\centering 
\subfloat{\includegraphics[width=0.9\textwidth, keepaspectratio]{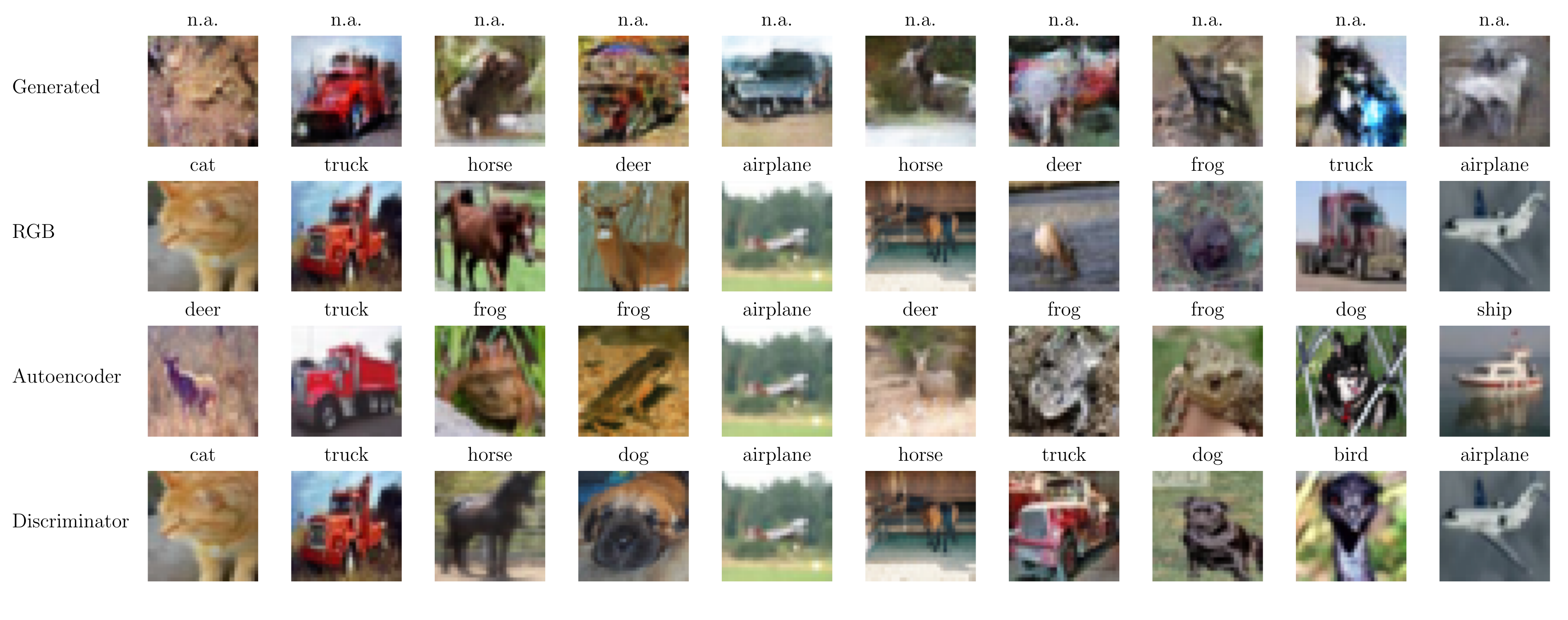}}
\caption{The rows show generated and matched images for \emph{CIFAR-10 - ten classes} using WGAN-GP. Most of the generated images achieve only a moderate quality and even the closest samples from the pool have a high perceptual visual distance or assign images that show non matching classes, see last column where the images have a similar appearance but an appropriate label for the generated images would be \textsf{horse} but the selected samples show \textsf{airplane} and \textsf{ship}.}
\label{app:fig:cifar-all-matching}
\end{figure*}

Figs.~\ref{app:fig:mnist-binary-matching}, \ref{app:fig:mnist-all-matching}, \ref{app:fig:cifar-binary-matching}, \ref{app:fig:cifar-all-matching} show examples of generated images of the same active learning cycle and the corresponding matches. All images are generated using WGAN-GP and the maximum entropy score. The generated images are not manually annotated. The moderate quality of the generated CIFAR-10 images prevents confidently annotating the images. Instead, \emph{n.a.} indicates that the manual annotation is missing.

\clearpage

\begin{table*}[t]
  \renewcommand{\arraystretch}{1.3}
  \newcommand{\head}[1]{\textnormal{\textbf{#1}}}

  \colorlet{tableheadcolor}{gray!25} 
  \newcommand{\headcol}{\rowcolor{tableheadcolor}} %
  \colorlet{tablerowcolor}{gray!10} 
  \newcommand{\rowcol}{\rowcolor{tablerowcolor}} %
  \setlength{\aboverulesep}{0pt}
  \setlength{\belowrulesep}{0pt}

  \newcommand*{\rulefiller}{
    \arrayrulecolor{tableheadcolor}
    \specialrule{\heavyrulewidth}{0pt}{-\heavyrulewidth}
    \arrayrulecolor{black}}

  \caption{Auto-encoder architecture for ASAL on MNIST.}
  \label{tab:auto-mnist}
  \centering
  \begin{tabular}{c*{2}{c}}
    \toprule
    \headcol \head{Encoder} & \head{Decoder} \\
    \midrule
    Input: $28\times28\times1$ & Input: $4\times4\times12$ \\
    \midrule
    $5\times5$ conv: 3, stride=2 & $5\times5$ deconv: 6 \\
    leakyReLU & ReLU \\
    \midrule
    $5\times5$ conv: 6, stride=2 & $5\times5$ deconv: 3 \\
    leakyReLU & ReLU \\
    \midrule
    $5\times5$ conv: 12, stride=2  & $5\times5$ deconv: 1 \\
    leakyReLU  & sigmoid \\
    \bottomrule
  \end{tabular}
\end{table*}
\begin{table*}[t]
  \renewcommand{\arraystretch}{1.3}
  \newcommand{\head}[1]{\textnormal{\textbf{#1}}}

  \colorlet{tableheadcolor}{gray!25} 
  \newcommand{\headcol}{\rowcolor{tableheadcolor}} %
  \colorlet{tablerowcolor}{gray!10} 
  \newcommand{\rowcol}{\rowcolor{tablerowcolor}} %
  \setlength{\aboverulesep}{0pt}
  \setlength{\belowrulesep}{0pt}

  \newcommand*{\rulefiller}{
    \arrayrulecolor{tableheadcolor}
    \specialrule{\heavyrulewidth}{0pt}{-\heavyrulewidth}
    \arrayrulecolor{black}}

  \caption{Model architectures for ASAL on MNIST.}
  \label{tab:models-mnist}
  \centering
  \begin{tabular}{c*{3}{c}}
    \toprule
    \headcol \head{Classifier} & \head{Generator} & \head{Discriminator} \\
    \midrule
    Input: $28\times28\times1$ & Input: $z\sim \mathcal{N}(0,1)$: 128 & Input: $28\times28\times1$\\
    \midrule
    $5\times5$ conv: 32 & linear: $128\times4096$ & $5\times5$ conv: 64, stride=2\\
    ReLU, Maxpool $2\times2$ & ReLU & leakyReLU \\
    \midrule
    $5\times5$ conv: 64 & $5\times5$ deconv: 128 & $5\times5$ conv: 128, stride=2\\
    ReLU, Maxpool $2\times2$ & ReLU & leakyReLU \\
    \midrule
    linear: $3136\times1024$ & $5\times5$ deconv: 64 & $5\times5$ conv: 256, stride=2\\
    ReLU, Dropout: 0.5 & ReLU & leakyReLU \\
    \midrule
    linear: $1024\times10$ & $5\times5$ deconv: 64& linear: $3136\times1$\\
    & sigmoid & \\
    \bottomrule
  \end{tabular}
\end{table*}

\begin{table*}[t]
  \renewcommand{\arraystretch}{1.3}
  \newcommand{\head}[1]{\textnormal{\textbf{#1}}}

  \colorlet{tableheadcolor}{gray!25} 
  \newcommand{\headcol}{\rowcolor{tableheadcolor}} %
  \colorlet{tablerowcolor}{gray!10} 
  \newcommand{\rowcol}{\rowcolor{tablerowcolor}} %
  \setlength{\aboverulesep}{0pt}
  \setlength{\belowrulesep}{0pt}

  \newcommand*{\rulefiller}{
    \arrayrulecolor{tableheadcolor}
    \specialrule{\heavyrulewidth}{0pt}{-\heavyrulewidth}
    \arrayrulecolor{black}}

  \caption{Auto-encoder architecture for ASAL on CIFAR-10 and SVHN.}
  \label{tab:auto-cifar}
  \centering
  \begin{tabular}{c*{2}{c}}
    \toprule
    \headcol \head{Encoder} & \head{Decoder} \\
    \midrule
    Input: $32\times32\times3$ & Input: $4\times4\times16$ \\
    \midrule
    $3\times3$ conv: 64, Batch norm & $3\times3$ deconv: 32 \\
    ReLU, Maxpool $2\times2$ & Batch norm, ReLU \\
    \midrule
    $3\times3$ conv: 32, Batch norm & $3\times3$ deconv: 64 \\
    ReLU, Maxpool $2\times2$ & Batch norm, ReLU \\
    \midrule
    $3\times3$ conv: 16, Batch norm  & $3\times3$ deconv: 3 \\
    ReLU, Maxpool $2\times2$  & Batch norm, sigmoid \\
    \bottomrule
  \end{tabular}
\end{table*}
\begin{table*}[t]
  \renewcommand{\arraystretch}{1.3}
  \newcommand{\head}[1]{\textnormal{\textbf{#1}}}

  \colorlet{tableheadcolor}{gray!25} 
  \newcommand{\headcol}{\rowcolor{tableheadcolor}} %
  \colorlet{tablerowcolor}{gray!10} 
  \newcommand{\rowcol}{\rowcolor{tablerowcolor}} %
  \setlength{\aboverulesep}{0pt}
  \setlength{\belowrulesep}{0pt}

  \newcommand*{\rulefiller}{
    \arrayrulecolor{tableheadcolor}
    \specialrule{\heavyrulewidth}{0pt}{-\heavyrulewidth}
    \arrayrulecolor{black}}

  \caption{Model architectures for ASAL on CIFAR-10 and SVHN (Generator and Discriminator).}
  \label{tab:models-cifar}
  \centering
  \begin{tabular}{c*{3}{c}}
    \toprule
    \headcol \head{Classifier} & \head{Generator} & \head{Discriminator} \\
    \midrule
    Input: $32\times32\times3$ & Input: $z\sim \mathcal{N}(0,1)$: 128 & Input: $32\times32\times3$\\
    \midrule
    $3\times3$ conv: 96 & linear: $128\times8192$ & $5\times5$ conv: 128, stride=2\\
    ReLU & Batch norm, ReLU & leakyReLU \\
    \midrule
    $3\times3$ conv: 96 & $5\times5$ deconv: 256 & $5\times5$ conv: 256, stride=2\\
    ReLU & Batch norm, ReLU & leakyReLU \\
    \midrule
    $3\times3$ conv: 96, stride=2& $5\times5$ deconv: 128 & $5\times5$ conv: 512, stride=2\\
    ReLU, dropout=0.5 & Batch norm, ReLU & leakyReLU \\
    \midrule
    $3\times3$ conv: 192 & $5\times5$ deconv: 3& linear: $8192\times1$\\
    ReLU & Tanh &  \\
    \midrule
    $3\times3$ conv: 192&  &  \\
    ReLU&  & \\
    \midrule
    $3\times3$ conv: 192, stride=2&  &  \\
    ReLU, dropout=0.5&  & \\
    \midrule
    $3\times3$ conv: 192&  &  \\
    ReLU&  & \\
    \midrule
    $1\times1$ conv: 192&  &  \\
    ReLU&  & \\
    \midrule
    $1\times1$ conv: 10&  &  \\
    global average pool&  & \\
    \bottomrule
  \end{tabular}
\end{table*}
\begin{table*}[t]
  \renewcommand{\arraystretch}{1.3}
  \newcommand{\head}[1]{\textnormal{\textbf{#1}}}

  \colorlet{tableheadcolor}{gray!25} 
  \newcommand{\headcol}{\rowcolor{tableheadcolor}} %
  \colorlet{tablerowcolor}{gray!10} 
  \newcommand{\rowcol}{\rowcolor{tablerowcolor}} %
  \setlength{\aboverulesep}{0pt}
  \setlength{\belowrulesep}{0pt}

  \newcommand*{\rulefiller}{
    \arrayrulecolor{tableheadcolor}
    \specialrule{\heavyrulewidth}{0pt}{-\heavyrulewidth}
    \arrayrulecolor{black}}

  \caption{Residual GAN architectures for ASAL on CIFAR-10.}
  \label{tab:resnet-cifar}
  \centering
  \begin{tabular}{c*{2}{c}}
    \toprule
    \headcol \head{Generator} & \head{Discriminator} \\
    \midrule
    Input: $z\sim \mathcal{N}(0,1)$: 128 & Input: $32\times32\times3$ &  \\
    \midrule
    linear: $128\times2048$ & $[3\times3]\times2$ ResidualBlock: 128 \\
    & Down=2 \\
    \midrule
    $[3\times3]\times2$ ResidualBlock: 128 & $[3\times3]\times2$ ResidualBlock: 128 \\
    Up=2 & Down=2 \\
    \midrule
    $[3\times3]\times2$ ResidualBlock: 128  & $[3\times3]\times2$ ResidualBlock: 128 \\
    Up=2  &  \\
    \midrule
    $[3\times3]\times2$ ResidualBlock: 128  & $[3\times3]\times2$ ResidualBlock: 128 \\
    Up=2  &  ReLU, MeanPool\\
    \midrule
    $3\times3$ conv: 3  & linear: $128\times1$ \\
    Tanh  &\\
    \bottomrule
  \end{tabular}
\end{table*}

\begin{table*}[t]
  \renewcommand{\arraystretch}{1.3}
  \newcommand{\head}[1]{\textnormal{\textbf{#1}}}

  \colorlet{tableheadcolor}{gray!25} 
  \newcommand{\headcol}{\rowcolor{tableheadcolor}} %
  \colorlet{tablerowcolor}{gray!10} 
  \newcommand{\rowcol}{\rowcolor{tablerowcolor}} %
  \setlength{\aboverulesep}{0pt}
  \setlength{\belowrulesep}{0pt}

  \newcommand*{\rulefiller}{
    \arrayrulecolor{tableheadcolor}
    \specialrule{\heavyrulewidth}{0pt}{-\heavyrulewidth}
    \arrayrulecolor{black}}

  \caption{Auto-encoder architecture for ASAL on CelebA and LSUN.}
  \label{tab:auto-celeba}
  \centering
  \begin{tabular}{c*{2}{c}}
    \toprule
    \headcol \head{Encoder} & \head{Decoder} \\
    \midrule
    Input: $64\times64\times3$ & Input: $4\times4\times16$ \\
    \midrule
    $5\times5$ conv: 128, stride=2 & $5\times5$ deconv: 32 \\
    Batch norm, ReLU & Batch norm, ReLU \\
    \midrule
    $5\times5$ conv: 64, stride=2 & $5\times5$ deconv: 64 \\
    Batch norm, ReLU & Batch norm, ReLU \\
    \midrule
    $5\times5$ conv: 32, stride=2 & $5\times5$ deconv: 128 \\
    Batch Norm, ReLU  & Batch norm, ReLU \\
    \midrule
    $5\times5$ conv: 16, stride=2 & $5\times5$ deconv: 3 \\
     & Tanh\\
    \bottomrule
  \end{tabular}
\end{table*}
\begin{table*}[t]
  \renewcommand{\arraystretch}{1.3}
  \newcommand{\head}[1]{\textnormal{\textbf{#1}}}

  \colorlet{tableheadcolor}{gray!25} 
  \newcommand{\headcol}{\rowcolor{tableheadcolor}} %
  \colorlet{tablerowcolor}{gray!10} 
  \newcommand{\rowcol}{\rowcolor{tablerowcolor}} %
  \setlength{\aboverulesep}{0pt}
  \setlength{\belowrulesep}{0pt}

  \newcommand*{\rulefiller}{
    \arrayrulecolor{tableheadcolor}
    \specialrule{\heavyrulewidth}{0pt}{-\heavyrulewidth}
    \arrayrulecolor{black}}

  \caption{Model architectures for ASAL on CelebA and LSUN (Generator and Discriminator).}
  \label{tab:models-celeba}
  \centering
  \begin{tabular}{c*{3}{c}}
    \toprule
    \headcol \head{Classifier} & \head{Generator} & \head{Discriminator} \\
    \midrule
    Input: $64\times64\times3$ & Input: $z\sim \mathcal{N}(0,1)$: 128 & Input: $64\times64\times3$\\
    \midrule
    $3\times3$ conv: 16 & linear: $128\times4096$ & $5\times5$ conv: 128, stride=2\\
    ReLU, Maxpool $2\times2$ & Batch norm, ReLU & leakyReLU \\
    \midrule
    $3\times3$ conv: 32 & $5\times5$ deconv: 256 & $5\times5$ conv: 256, stride=2\\
    ReLU, Maxpool $2\times2$ & Batch norm, ReLU & leakyReLU \\
    \midrule
    $3\times3$ conv: 64 & $5\times5$ deconv: 128 & $5\times5$ conv: 512, stride=2\\
    ReLU, Maxpool $2\times2$ & Batch norm, ReLU & leakyReLU \\
    \midrule
    linear: $4096\times1024$ & $5\times5$ deconv: 64& $5\times5$ conv: 512, stride=2\\
    ReLU, Dropout 0.5& Batch norm, ReLU & leakyReLU \\
    \midrule
    linear: $1024\times1$ & $5\times5$ deconv: 3 & linear: $8192\times1$ \\
     & Tanh & \\
    \bottomrule
  \end{tabular}
\end{table*}






\end{document}